\documentclass[journal]{IEEEtran}
\usepackage{amssymb}
\usepackage{graphicx}

\usepackage{url}

\usepackage{xcolor}
\usepackage{multirow}

\usepackage{subfigure}
\usepackage{algorithm}
\usepackage{algorithmicx}
\usepackage{algpseudocode}
\usepackage{amsmath}
\usepackage{ifthen}
\usepackage[figuresright]{rotating}

\usepackage{booktabs}
\usepackage{url}
\usepackage{cite}
\usepackage{graphics}
\usepackage{amssymb}
\usepackage{booktabs,multirow,longtable}

\usepackage{longtable}
\usepackage{float}
\usepackage{makecell}

\usepackage{xcolor}
\usepackage{colortbl,booktabs}
\definecolor{mygray}{gray}{0.75}

\newcommand{\tabincell}[2]{\begin{tabular}{@{}#1@{}}#2\end{tabular}}
\ifCLASSINFOpdf
\else
\fi
\begin{document}
%
\title{A Lagrangian Dual-based Theory-guided Deep Neural Network}

\author{Miao~Rong,
        Dongxiao~Zhang,
        Nanzhe~Wang
\thanks{M. Rong is with the Intelligent Energy Laboratory, PengCheng Laboratory, Shenzhen, 518000, P. R. China.}
\thanks{D. Zhang is with the School of Environmental Science and Engineering, Southern University of Science and Technology, Shenzhen, 518055, P. R. China.}
\thanks{N. Wang is with the College of Engineering, Peking University, Beijing, 100871, P. R. China.}
}

\markboth{A Lagrangian Dual-based Theory-guided Deep Neural Network}%
{Rong \MakeLowercase{\textit{et al.}}: Bare Demo}
\maketitle

\begin{abstract}
The theory-guided neural network (TgNN) is a kind of method which improves the effectiveness and efficiency of neural network architectures by incorporating scientific knowledge or physical information. Despite its great success, the theory-guided (deep) neural network possesses certain limits when maintaining a tradeoff between training data and domain knowledge during the training process. In this paper, the Lagrangian dual-based TgNN (TgNN-LD) is proposed to improve the effectiveness of TgNN. We convert the original loss function into a constrained form with fewer items, in which partial differential equations (PDEs), engineering controls (ECs), and expert knowledge (EK) are regarded as constraints, with one Lagrangian variable per constraint. These Lagrangian variables are incorporated to achieve an equitable tradeoff between observation data and corresponding constraints, in order to improve prediction accuracy, and conserve time and computational resources adjusted by an \emph{ad-hoc} procedure. To investigate the performance of the proposed method, the original TgNN model with a set of optimized weight values adjusted by \emph{ad-hoc} procedures is compared on a subsurface flow problem, with their L2 error, R square (R2), and computational time being analyzed. Experimental results demonstrate the superiority of the Lagrangian dual-based TgNN.
\end{abstract}

\begin{IEEEkeywords}
theory-guided neural network; Lagrangian dual; weights adjustment; tradeoff.
\end{IEEEkeywords}

%
\IEEEpeerreviewmaketitle

\section{Introduction}
\IEEEPARstart
{T}he deep neural network (DNN) has achieved significant breakthroughs in various scientific and industrial fields \cite{4aISI:000368673800028, chen2019communication,chen2019network,8aISI:000529484002100}. Like most data-driven models in artificial intelligence \cite{15aISI:000520407400011}, they are also dependent on a large amount of training data. However, the cost and the difficulty of collecting data in some areas, especially in energy-related fields, hinders the development of (deep) neural networks. To further increase their generalization, theory-guided data science models, which bridge scientific problems and complex physical phenomena, have gained increased popularity in recent years \cite{1a7959606,5a1710.11431,6aISI:000453776000028}.

As a successful representative, the theory-guided neural network framework, also called a physical-informed neural network framework or an informed deep learning framework, which incorporates the theory (e.g., governing equations, other physical constraints, engineering controls, and expert knowledge) into (deep) neural network training, has been applied to construct the prediction model, especially in industries with limited training data \cite{1a7959606,5a1710.11431}. Herein, the theory may refer to scientific laws and engineering theories \cite{1a7959606}, which can be regarded as a kind of prior knowledge. Such knowledge is combined with training data to improve training performance during the learning process. In the loss function, they are usually transformed into regularization terms and added up with the training term \cite{2aISI:000527390200029}. Owing to the existence of theory, predictions obtained by TgNN take physical feasibility and knowledge beyond the regimes covered with the training data into account. As a result, TgNN can obtain a training model with better generalization, and can achieve higher accuracy than the traditional DNN \cite{4aISI:000368673800028,5a1710.11431}.

Although the introduction of theory expands the application of data-driven models, the tradeoff between observation data and theory should be equitable. Herein, we first provide its theory-incorporated mathematical formulation, as shown in Eq.\ref{neqf1}:
\begin{equation}
\begin{aligned}
L\left( \theta  \right)=&\lambda_{DATA}MSE_{DATA}+\lambda_{IC}MSE_{IC}\\
&+\lambda_{BC}MSE_{BC}+\lambda_{PDE}MSE_{PDE}\\
&+\lambda_{EC}MSE_{EC}+\lambda_{EK}MSE_{EK}\\
\end{aligned}
\label{neqf1}
\end{equation}
where $\lambda _i$ and $MSE_i$ denote the weight and mean square error for $i$-th term, respectively; and $\lambda  = \left[ {\lambda _{DATA} ,\lambda _{IC} ,\lambda _{BC} ,\lambda _{PDE} ,\lambda _{EC} ,\lambda _{EK} } \right]$, where the term $DATA$ refers to the observation data or training data. The remaining terms are the added theory in the (D)NN model. The governing equations consist of terms $PDE$, $IC$, $BC$, $EC$, and $EK$, referring to partial differential equations, initial conditions, boundary conditions, engineering control, and engineering knowledge, respectively. Each weight term represents the importance of the corresponding term in the loss function. In addition, only the might the values of these terms be at different scales, but their physical meanings and dimensional units can also be distinct. Therefore, balancing the tradeoff among these terms is critical.

If viewing these weight variables as neural architecture parameters, the gradient of $\lambda _i$ is calculated as $MSE_i$ from Eq.\ref{neqf1}, i.e., a constant nonnegative value, making $\lambda _i$ continuously decrease until negative infinity with the increase of iterations at the stage of back-propagation \cite{7aISI:000457843608090,chen2019federated}. Therefore, due to the existence of theory, i.e., regularization terms in the loss function, it is difficult to determine the weights of each term in comparison with the training data term. If set inappropriately, it is highly possible to increase the training time, or even impede the convergence of the optimizer, contributing to an inaccurate training model. Consequently, the adjustment of these weight values is essential. In most existing literature, researchers often adjust these values by experience \cite{1a7959606,2aISI:000527390200029,6aISI:000453776000028}. However, if these weights are not at the same scale, this will inevitably create a heavy burden on researchers and place a constraint on the ability to conserve human time.

Recently, babysitting or evolutionary computation based techniques \cite{addaISI:000522097600020}, such as grid search \cite{18aISI:000470661600015,19aISI:000424854300002} and genetic algorithm \cite{16aISI:000518864800019,17aISI:000467910600004}, have achieved rising popularity for hyper-parameter optimization \cite{11aISI:000537570300005,12ayao2018cost,14aarticle}. If TgNN first generates an initial set of weights, followed by comparing and repeating this searching process until the most suitable set of weight values is found or the stopping criterion is met, the training time will be absolutely extended. In contrast, if the search for optimized weight values can be incorporated into the training process, the training time may be shortened.

In recent years, Lagrangian dual approaches have been widely combined with the (deep) neural network to improve the latter's ability when dealing with problems with constraints \cite{3a2001.09394,21a2005.10674,22a2005.10691,23aISI:000502589700005}. Ferdinando et al. pointed out that Lagrangian duality can bring significant benefits for applications in which the learning task must enforce constraints on the predictor itself, such as in energy systems, gas networks, transprecision computing, among others \cite{3a2001.09394}. Walker et al. incorporated the Lagrangian dual approach into laboratory and prospective observational studies \cite{24aISI:000460696900032}. Gan et al. developed a Lagrangian dual-based DNN framework for trajectory simulation \cite{23aISI:000502589700005}. Pundir and Raman proposed a dual deep learning method for image-based smoke detection \cite{25aISI:000509346100025}. The above contributions can improve the performance of the original deep neural network framework and provide more accurate predictive training models.

Having realized this, we propose the Lagrangian dual-based TgNN (TgNN-LD) to provide theoretical guidance for the adjustment of weight values in the loss function of the theory-guided neural network framework. In our method, the Lagrangian dual framework is incorporated into the TgNN training model, and controls the update of weights with the purpose of automatically changing the weight values and producing accurate predictive results within limited training time. Moreover, to better set forth our approach, we select a subsurface flow problem as a test case in the experiment.

The reminder of this paper proceeds as follows. Section \ref{sec2} briefly describes the mathematical formulation of the TgNN model, followed by details of the proposed method. The experimental settings with the investigation of corresponding results are provided in Section \ref{sec3} and \ref{sec4}, respectively. Finally, Section \ref{sec5} concludes the paper and suggests directions for future research.
\section{Proposed Method}\label{sec2}
In this paper, we consider the following optimization problem:
\begin{equation}
\begin{aligned}
L\left( \theta  \right)=&MSE_{DATA}+MSE_{IC}+MSE_{BC}\\
&+\lambda _{PDE}MSE_{PDE}+\lambda _{EC} MSE_{EC}\\
&+\lambda _{EK}MSE_{EK}\\
\end{aligned}
\label{neqf2}
\end{equation}
since the initial condition (IC) and boundary condition (BC), which impose restrictions for the decision space, can be regarded as a part of the data term. The values of $\lambda _{PDE}$ , ${\rm{ }}\lambda _{EC}$, and $\lambda _{EK}$ have a great impact on the optimization results. As discussed previously, not only might the values of these terms be at different scales, but their physical meanings and dimensional units can also be dissimilar. As a consequence, by introducing the formation of Eq.\ref{neqf2}, $\lambda _i$ is expected to achieve a normalized balance between training data and the other terms. If it is inappropriately assigned, however, the prediction accuracy will be diminished, and both training time and computational cost will markedly increase. To theoretically determine the weight values and maintain the balance of each term in TgNN, we propose the Lagrangian dual-based TgNN framework.
\subsection{Problem description}
We first introduce the following mathematical descriptions of governing equations of the underlying physical problem:
\begin{equation}
\begin{array}{l}
 L_p u\left( {x,t} \right) = l\left( {x,t} \right),x \in \Omega ,t \in \left( {0,T} \right]
 \end{array}
\label{neqaddgen1}
\end{equation}
\begin{equation}
\begin{array}{l}
 Iu\left( {x,0} \right) = q\left( {x,t} \right),x \in \Omega
 \end{array}
\label{neqaddgen2}
\end{equation}
\begin{equation}
\begin{array}{l}
 Bu\left( {x,t} \right) = p\left( x \right),x \in \partial \Omega ,t \in \left( {0,T} \right]
 \end{array}
\label{neqaddgen3}
\end{equation}
where $\Omega  \subset R^d$ and $t \in \left( {0,T} \right)$ denote the spatial and temporal domain, respectively, with $\partial \Omega$ as the spatial boundaries; $L_p$ and $L_p u$ represent a differential operator and its spatial derivatives, respectively, of $u$; $l$ is a forcing term; and $I$ and $B$ are two other operators which define the initial and boundary conditions, respectively.

As discussed in Eq.\ref{neqf1}, TgNN incorporates theory into DNN by the summation of corresponding terms. It is constructed based on the following six general parts shown in Eq.\ref{neqaddgen4}:
\begin{equation}
\left\{ {\begin{array}{*{20}l}
   {MSE_{DATA}  = \frac{1}{{n_d }}\sum\limits_{i=1}^{n_d} {\left| {N_{u} \left( {x_{}^i ,y_{}^i } \right) - u_i \left( {x_{}^i ,y_{}^i } \right)} \right|^2 } } \hfill  \\
   {MSE_{PDE} {\rm{ = }}\frac{1}{{n_f }}\sum\limits_{i=1}^{n_f } {\left| {f\left( {t_f^i ,x_f^i ,y_f^i } \right)} \right|^2 } } \hfill  \\
   {MSE_{IC} {\rm{ = }}\frac{1}{{n_{IC} }}\sum\limits_{i=1}^{n_{IC} } {\left| {N_{u} \left( {x_i ,0} \right) - u_{I} \left( {x_i ,0} \right)} \right|^2 } } \hfill  \\
   {MSE_{BC} {\rm{ = }}\frac{1}{{n_{BC} }}\sum\limits_{i=1}^{n_{BC} } {\left| {N_{u} \left( {\partial \Omega _i ,t_i } \right) - u_{B} \left( {\partial \Omega _i ,t_i } \right)} \right|^2 } } \hfill  \\
   \begin{array}{l}
 MSE_{EC} {\rm{ = }}\frac{1}{{n_{EC} }}\sum\limits_{i=1}^{n_{EC} } {\left| {N_{u} \left( {x_{}^i ,y_{}^i } \right) - u_{EC} \left( {x_{}^i ,y_{}^i } \right)} \right|^2 }  \\
 MSE_{EK} {\rm{ = }}\frac{1}{{n_{EK} }}\sum\limits_{i=1}^{n_{EK} } {\left| {N_{u} \left( {x_{}^i ,y_{}^i } \right) - u_{EK} \left( {x_{}^i ,y_{}^i } \right)} \right|^2 }  \\
 \end{array} \hfill  \\
\end{array}} \right.
\label{neqaddgen4}
\end{equation}
where $f: = L_p N_{u} \left( {x,t} \right) - l\left( {x,t,y} \right)$ needs to approach to 0, representing the residual of the partial differential Eq.\ref{neqaddgen1}; $\left\{ {t_f^i ,x_f^i ,y_f^i } \right\}_{i = 1}^{n_f }$ denotes the collocation points of the residual function with the size of $n_f$, which can be randomly chosen because no labels are needed for these points; $N_{u}$ is the approximation of a solution $u$ obtained by the (deep) neural network; $n_{d}$ represents the numbers of training data; $n_{IC}$ and $n_{BC}$ denote the collocation points for the evaluation of initial and boundary conditions, respectively; and $n_{EC}$ and $n_{EK}$ denote the collocation points of engineering control and knowledge, respectively.
\subsection{Problem transformation}
Having realized the mathematical description of each term, we then convert the original loss function of the TgNN model, which incorporates scientific knowledge and engineering controls, to be re-written as Eq.\ref{neqf4}:
\begin{equation}
\begin{array}{l}
 \min {\rm{ }}L\left( \theta  \right){\rm{ = }}MSE_{DATA}  + MSE_{IC}  + MSE_{BC}  \\
 s.t.{\rm{   }}\left\{ {\begin{array}{*{20}l}
   {f = 0}  \\
   {EC \le 0}  \\
   {EK \le 0}  \\
\end{array}} \right. \\
 \end{array}
\label{neqf4}
\end{equation}

Following this, a Lagrangian duality framework \cite{3a2001.09394}, which incorporates a Lagrangian dual approach into the learning task, is employed to learn this constrained optimization problem and approximate minimizer  $L\left( \theta  \right)$. Given three multipliers, $\lambda _1$, $\lambda _2$, and $\lambda _3$, corresponding to per constraint, consider the Lagrangian loss function
\begin{equation}
L_\lambda  \left( \theta  \right){\rm{ = }}L\left( \theta  \right) + \lambda _1 \nu \left( {f = 0} \right) + \lambda _2 \nu \left( {EC \le 0} \right) + \lambda _3 \nu \left( {EK \le 0} \right)
\label{neqf5}
\end{equation}
where $\nu$ can be written as:
\begin{equation}
\nu \left( {g\left( x \right)} \right) = \left\{ {\begin{array}{*{20}l}
   {g\left( x \right),{\rm{if }}g\left( x \right){\rm{  =  0}}} \hfill  \\
   {{\mathop{\rm ReLU}\nolimits} \left( {g\left( x \right)} \right),{\rm{ if }}g\left( x \right){\rm{ }} \le {\rm{ 0}}} \hfill  \\
\end{array}} \right.
\label{neqf6}
\end{equation}

According to \cite{3a2001.09394}, the previous Lagrangian loss function with respect to multipliers $\lambda _i \left( {i = 1,...,3} \right)$, can then be transformed into a function with the purpose of finding the $\omega$, which can minimize the Lagrangian loss function, solving the optimization problem shown as follows:
\begin{equation}
\omega ^* \left( {\lambda _i } \right) = \mathop {\arg \min }\limits_\omega  L_{\lambda _i } \left( {{\rm M}\left[ \omega  \right]} \right)
\label{neqf7}
\end{equation}

Herein, we denote an approximation of the approximated optimizer ${\rm O}$ as $\tilde {\rm O}_\lambda   = {\rm M}\left[ {\omega ^* \left( \lambda  \right)} \right]\left( {\lambda  = \left\{ {\lambda _i \left| {i = 1,...,3} \right.} \right\}} \right)$, which can be produced by an optimizer (in our paper, we use Adam) during the training process.

Next, the above problem is transformed by the Lagrangian dual approach into searching the required optimal multipliers for a max-min problem shown as follows:
\begin{equation}
\lambda ^*  = \arg \mathop {\max }\limits_{\lambda _i \left( {i = 1,...3} \right)} \mathop {\min }\limits_\omega  \sum\limits_{j = 1}^n {L_\lambda  \left( {{\rm M}\left[ {\omega ^* \left( \lambda  \right)} \right]} \right)}
\label{neqf8}
\end{equation}

The same as \cite{3a2001.09394}, we denote this approximation as $\tilde {\rm O}_\lambda   = {\rm M}\left[ {\omega ^* \left( {\lambda _{}^* } \right)} \right]$.

To summarize \cite{3a2001.09394}, the repeated calculation and search process by the Lagrangian dual framework adhere to the following steps:\\
(a)	Learn $\tilde {\rm O}^k _\lambda$;\\
(b)	Let $y_j^k  = \tilde {\rm O}_\lambda  \left( {d_j } \right)$, where $y_j^k$ denotes the $j$-th training sample, and $d_j^{}$ refers to the label;\\
(c)	$\lambda _i^{k + 1}  = \lambda _i^k  + s_k \sum\nolimits_{j = 1}^n {v\left( {g\left( {y_j^k ,d_j } \right) \le 0} \right)} ,{\rm{ }}i = 1,...,3$.\\
where $s_k$ and $k$ refer to the update step size and the current iteration, respectively. In our problem, we recommend $s_k  \in \left[ {1.1,1.4} \right]$, and we utilize $s_k {\rm{ = }}1.25$ in our experiment.

\section{Experimental Settings}\label{sec3}
In this section, we take a 2-D unsteady-state single phase subsurface flow problem \cite{2aISI:000527390200029} as the test case to investigate the performance of the proposed Lagrangian dual-based TgNN.
\subsection{Parameter settings and scenario description}
The governing equation of the subsurface flow problem in our experiment can be written as Eq.\ref{neqfaddss}:
\begin{equation}
\begin{aligned}
S_s \frac{{\partial h\left( {t,x,y} \right)}}{{\partial t}} = &\frac{\partial }{{\partial x}}\left( {K\left( {x,y} \right) \cdot \frac{{\partial h\left( {t,x,y} \right)}}{{\partial x}}} \right) \\
&- \frac{\partial }{{\partial y}}\left( {K\left( {x,y} \right) \cdot \frac{{\partial h\left( {t,x,y} \right)}}{{\partial y}}} \right)
\end{aligned}
\label{neqfaddss}
\end{equation}
where $h$ is the hydraulic head that needs to be predicted; and $S_s  = 0.0001$ and $K$ are the specific storage and the hydraulic conductivity field, respectively. When $h$ is approximate with the neural network, ${N_h \left( {t,x,y;\theta } \right)}$, the residual of the governing equation of flow can be written as:
\begin{equation}
\begin{aligned}
f: = &S_s \frac{{\partial N_h \left( {t,x,y;\theta } \right)}}{{\partial t}} \\
&- \frac{\partial }{{\partial x}}\left( {K\left( {x,y} \right) \cdot \frac{{\partial N_h \left( {t,x,y;\theta } \right)}}{{\partial x}}} \right) \\
&- \frac{\partial }{{\partial y}}\left( {K\left( {x,y} \right) \cdot \frac{{\partial N_h \left( {t,x,y;\theta } \right)}}{{\partial y}}} \right)
\end{aligned}
\label{neqfadd1}
\end{equation}

In Eq.\ref{neqfadd1}, the partial derivatives of ${\partial N_h \left( {t,x,y;\theta } \right)}$ can be calculated in the network while the partial derivatives of ${K\left( {x,y} \right)}$ are, in general, required to be computed by numerical difference. Herein, we view the hydraulic conductivity field as a heterogeneous parameter field, i.e., a random field following a specific distribution with corresponding covariance \cite{2aISI:000527390200029}. Since its covariance is known, the Karhunen-Loeve expansion (KLE) is utilized to parameterize this kind of heterogeneous model. As a result, the residual of the governing equation can be re-written as Eq.\ref{neqfadd2}:
\begin{equation}
\begin{aligned}
 f: = &S_s \frac{{\partial N_h \left( {t,x,y;\theta } \right)}}{{\partial t}} \\
 &- \frac{\partial }{{\partial x}}\left( {e^{\bar Z\left( {x,y} \right) + \sum\limits_{i = 1}^n {\sqrt {\lambda _i } f_i \left( {x,y} \right)\xi _i \left( \tau  \right)} }  \cdot \frac{{\partial N_h \left( {t,x,y;\theta } \right)}}{{\partial x}}} \right) \\
  &- \frac{\partial }{{\partial y}}\left( {e^{\bar Z\left( {x,y} \right) + \sum\limits_{i = 1}^n {\sqrt {\lambda _i } f_i \left( {x,y} \right)\xi _i \left( \tau  \right)} }  \cdot \frac{{\partial N_h \left( {t,x,y;\theta } \right)}}{{\partial y}}} \right) \\
 \end{aligned}
\label{neqfadd2}
\end{equation}
where ${\bar Z\left( {x,y} \right) + \sum\limits_{i = 1}^n {\sqrt {\lambda _i } f_i \left( {x,y} \right)\xi _i \left( \tau  \right)} }$ represents the hydraulic conductivity field $Z\left( {x,y} \right) = \ln K\left( {x,\tau } \right)$ with ${\xi _i \left( \tau  \right)}$ as the $i$-th independent random variable of the field $Z\left( {x,y} \right)$. For the sake of fairness, the settings of our experiments remain the same as those in \cite{2aISI:000527390200029}, listed in Tab.\ref{tab:settings}. MODFLOW software is adopted to perform the simulations to obtain the required training dataset.
\begin{table*}[htbp]
  \centering
  \caption{Scenario settings}
    \begin{tabular}{cccccccccccccccc}
    \toprule
      a square domain    &  evenly divided into $51\times51$ grid blocks \\
      \hline
      length in both directions of domain & \tabincell{c}{1020$[Len]$,\\$Len$ denotes any consistent length unit} \\
      \hline
      specific storage $S_s$ & $0.0001\left[ {Len^{ - 1} } \right]$\\
      \hline
      the total simulation time	& $10\left[ T \right]$ ($\left[ T \right]$ denotes any consistent time unit)\\
      \hline
      each time step&$0.2\left[ T \right]$ (50 time steps in total)\\
      \hline
      the correlation length of the field & $\eta  = 408\left[ {Len} \right]$\\
      \hline
      hydraulic conductivity field settings &	\tabincell{c}{parameterized through KLE \\with 20 terms retained in the expansion, \\i.e., 20 random variables represent this field}\\
      \hline
      initial conditions&\tabincell{c}{$H_{t = 0,x = 0}  = 1\left[ {Len} \right]$ \\ $H_{t = 0,x \ne 0}  = 0\left[ {Len} \right]$}\\
      \hline
      prescribed heads&\tabincell{c}{the left boundary: $H_{x = 0}  = 1\left[ {Len} \right]$ \\ the right boundary: $H_{x = 1020}  = 0\left[ {Len} \right]$\\      two lateral boundaries: no-flow boundaries}\\
      \hline
      the log hydraulic conductivity&\tabincell{c}{mean: $\left\langle {\ln K} \right\rangle  = 0$\\
      variance:$\sigma _K^2  = 1.0$}\\
      \bottomrule
      \label{tab:settings}
      \end{tabular}
\end{table*}
\subsection{Compared methods}
Wang et al. determined a set of $\lambda$ values for TgNN via an \emph{ad-hoc} procedure, which is $\lambda=\left[ {\lambda _{DATA} ,\lambda _{IC} ,\lambda _{BC} ,\lambda _{PDE} ,\lambda _{EC} ,\lambda _{EK} } \right]=[1,1,1,100,1,1]$ \cite{2aISI:000527390200029}, for this particular problem. To investigate whether there are improvements in predictive accuracy, we utilize this weight setting as one of the compared methods and denote this case as TgNN. In addition, for comparison, we also take a naive approach by setting the weights equally as $\lambda=[1,1,1,1,1,1]$ and denote this case as TgNN-1.
\section{Comparisons and Results}\label{sec4}
This section first evaluates the predictive accuracy of the proposed Lagrangian dual-based TgNN framework on a subsurface flow problem, in comparison with TgNN and TgNN-1. Subsequently, we reduce the training epochs to observe whether the efficiency can be improved with less training time. Changes of Lagrangian multipliers are then recorded with their final values being assigned into the loss function to compare predictive performances obtained by dynamic adjustment and fixed values. Furthermore, different levels of noise are added into the training data to observe the effect on the predictive results caused by noise. Finally, the stopping criterion is substituted with a dynamic epoch, which has a relationship with changes of loss values, to control the training process.

\begin{figure*}[htb]
\begin{center}
\subfigure[$\ln K$]{\includegraphics[width=0.30\textwidth]{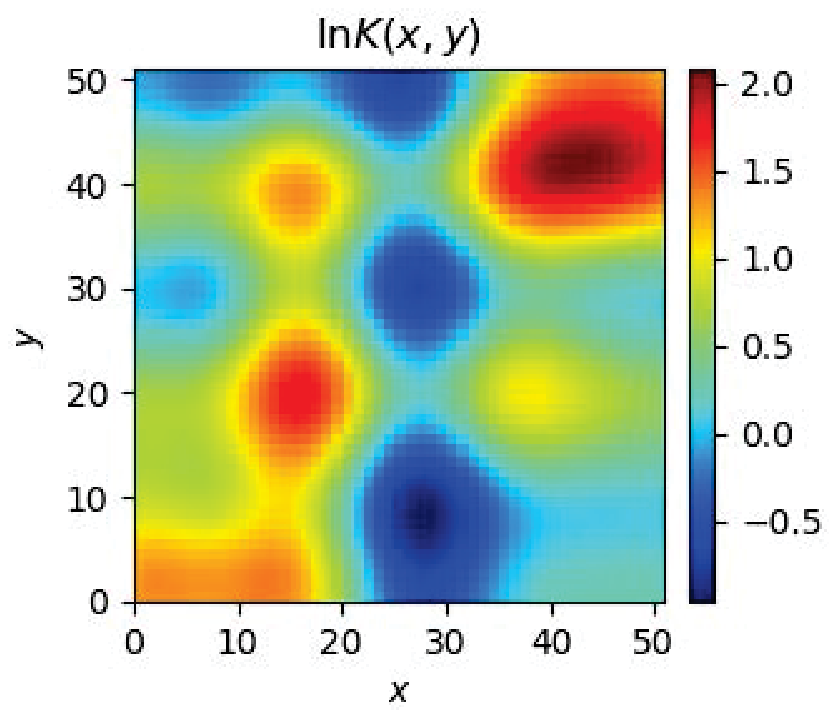}}
\subfigure[Reference $h$ at $t=50$]{\includegraphics[width=0.34\textwidth]{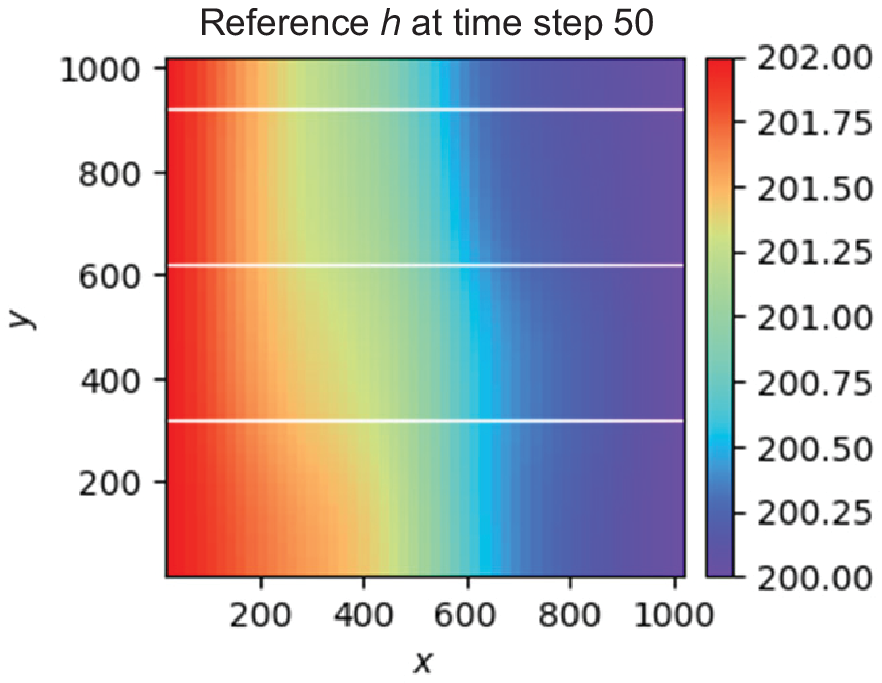}}
\caption{The distribution of the hydraulic conductivity field and the reference hydraulic head at $t=50$.}
\label{fig3add}
\end{center}
\end{figure*}

First, the distribution of the hydraulic conductivity field is provided in Fig.\ref{fig3add}.(a) with the reference hydraulic head at $t=50$ in Fig.\ref{fig3add}.(b).
\subsection{Predictive accuracy}
We first compare the above three methods with the number of iterations of 2000. Tab.\ref{tab:table2} provides the results of error L2 and R2, as well as the training time, with the best result per measure metric being marked in bold.
\begin{table*}[htbp]
  \centering
  \caption{Results of error L2, R2, and training time obtained by TgNN-LD, TgNN, and TgNN-1}
    \begin{tabular}{cccccccccccccccc}
    \toprule
          & Error L2 & R2 &	Training time/s\\
    \midrule
TgNN-LD&\textbf{2.0833E-04}&\textbf{9.9532E-01}&	204.3223\\
    \midrule
    TgNN	&3.5648E-04&	9.8735E-01	&195.3986\\
    \midrule
    TgNN-1&	4.5596E-04	&9.7931E-01&	\textbf{186.2980}\\
      \bottomrule
      \label{tab:table2}
      \end{tabular}
\end{table*}

As shown in Tab.\ref{tab:table2}, TgNN-LD can obtain the best error L2 and R2 results, which are obviously smaller and larger than the other two, respectively. It is worth noting that its training time seems lightly inferior among all three methods, probably due to the extra computation brought by the introduction of three Lagrangian multipliers. However, in comparison with TgNN-1 without any human adjustment, our proposed TgNN-LD achieves much better error L2 and R2 results. Moreover, compared to TgNN, in which the weights are adjusted by expertise, our method can not only save babysitting time in the preliminary stage of determining a better set of weights, but produce superior results, as well.

\begin{figure*}[htb]
\begin{center}
\subfigure[LD-TgNN: $loss$]{\includegraphics[width=0.3\textwidth]{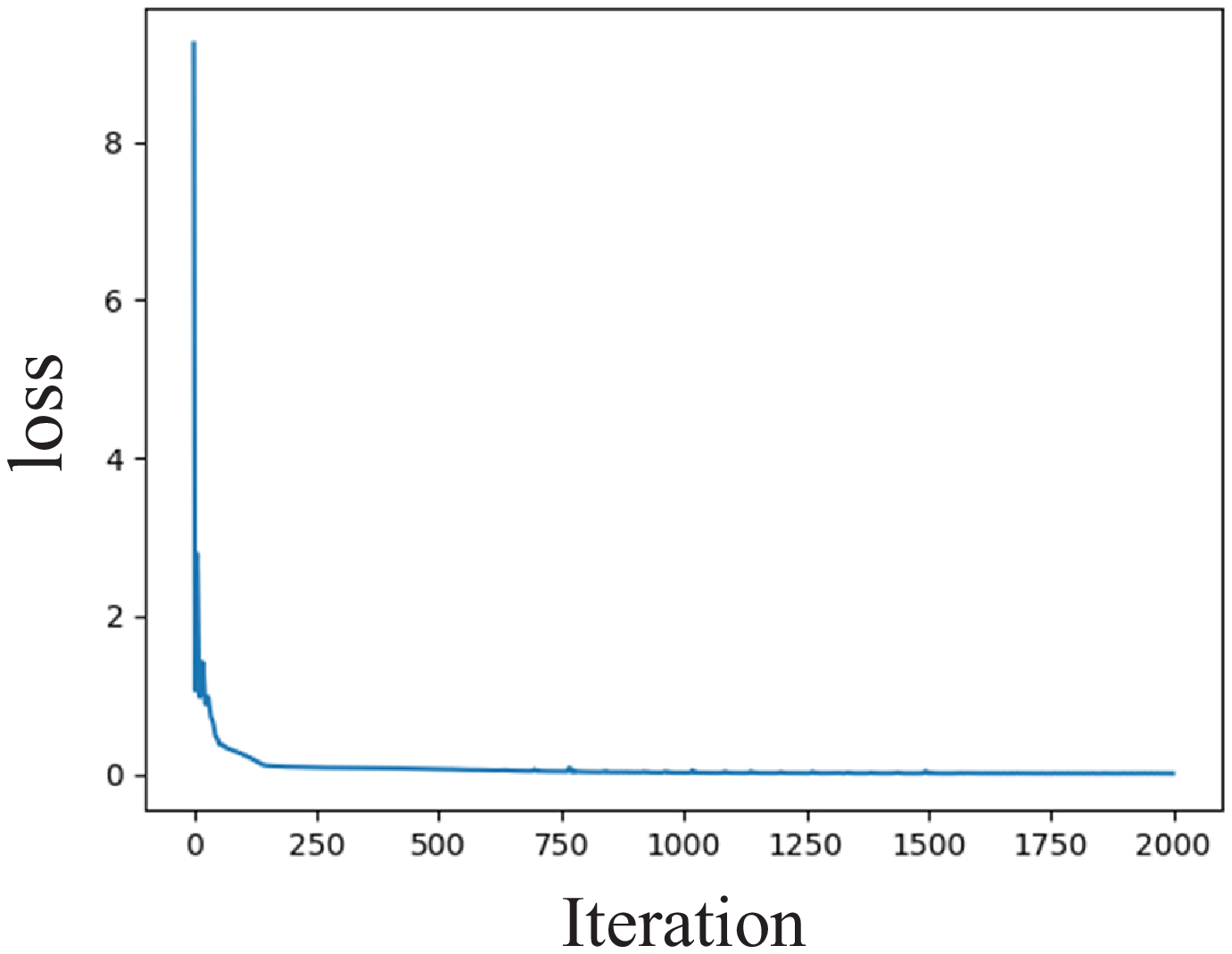}}
\subfigure[LD-TgNN: $f1\_loss$]{\includegraphics[width=0.30\textwidth]{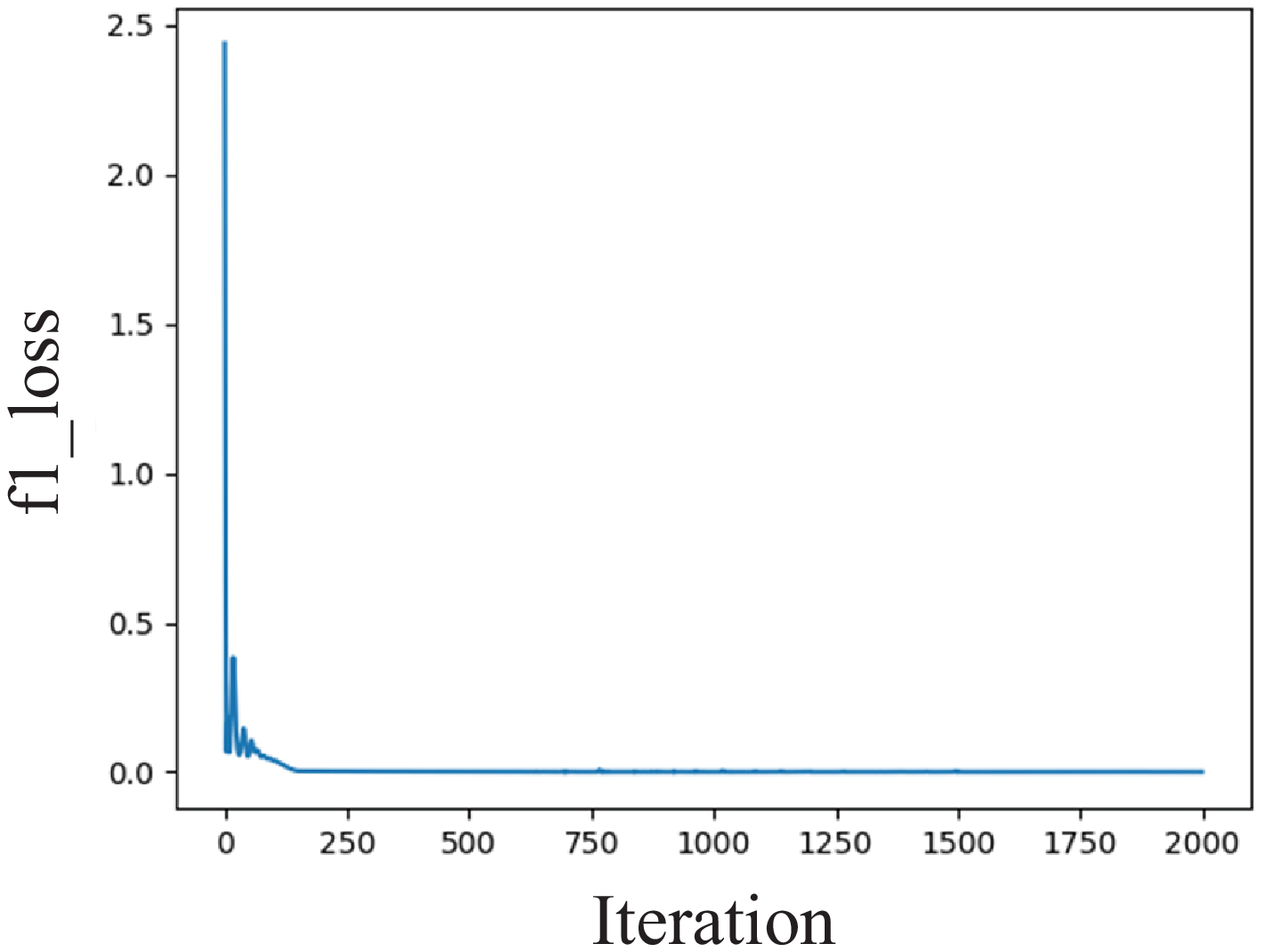}}
\subfigure[LD-TgNN: $f2\_loss$]{\includegraphics[width=0.30\textwidth]{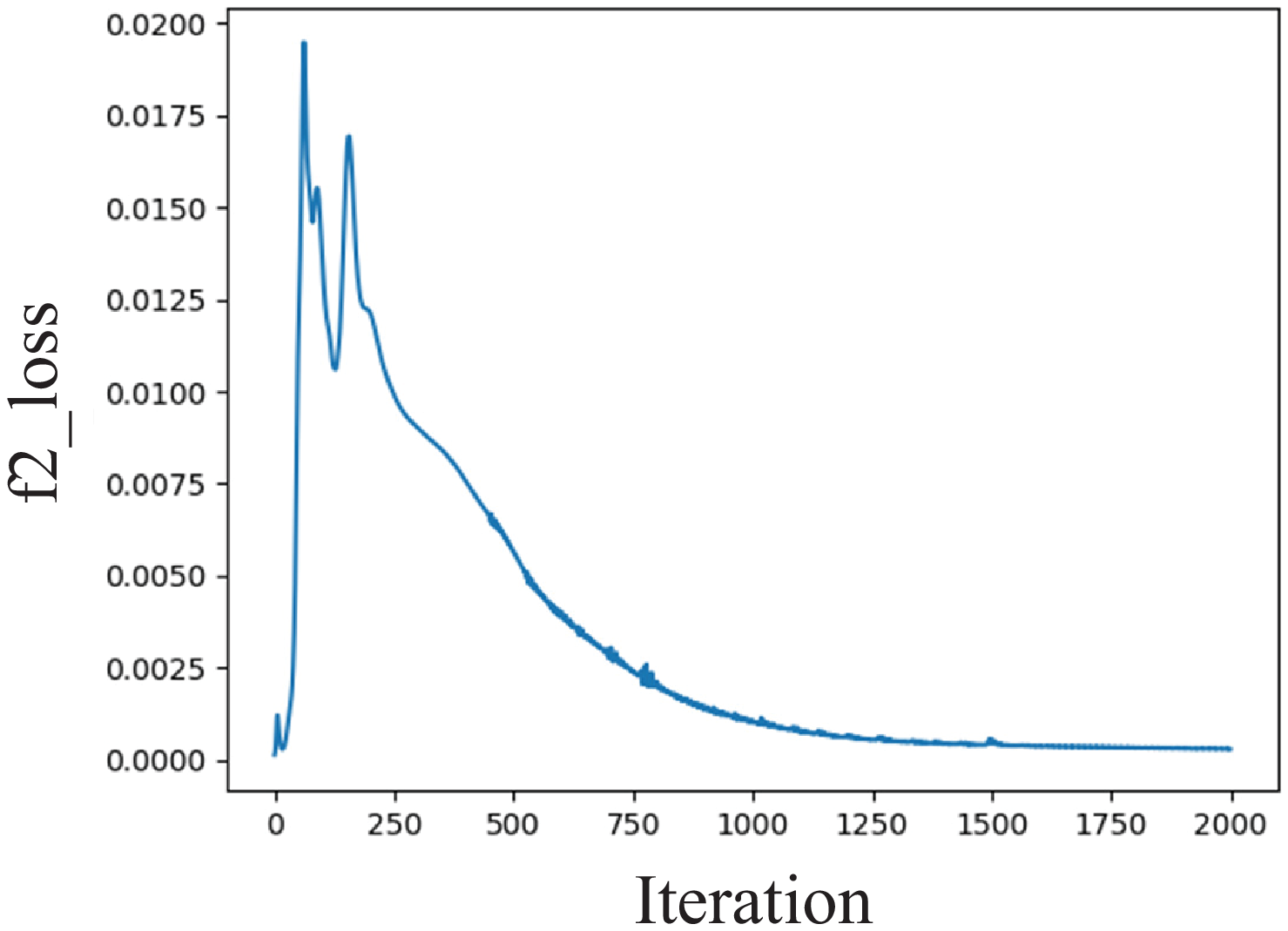}}\\
\subfigure[TgNN: $loss$]{\includegraphics[width=0.3\textwidth]{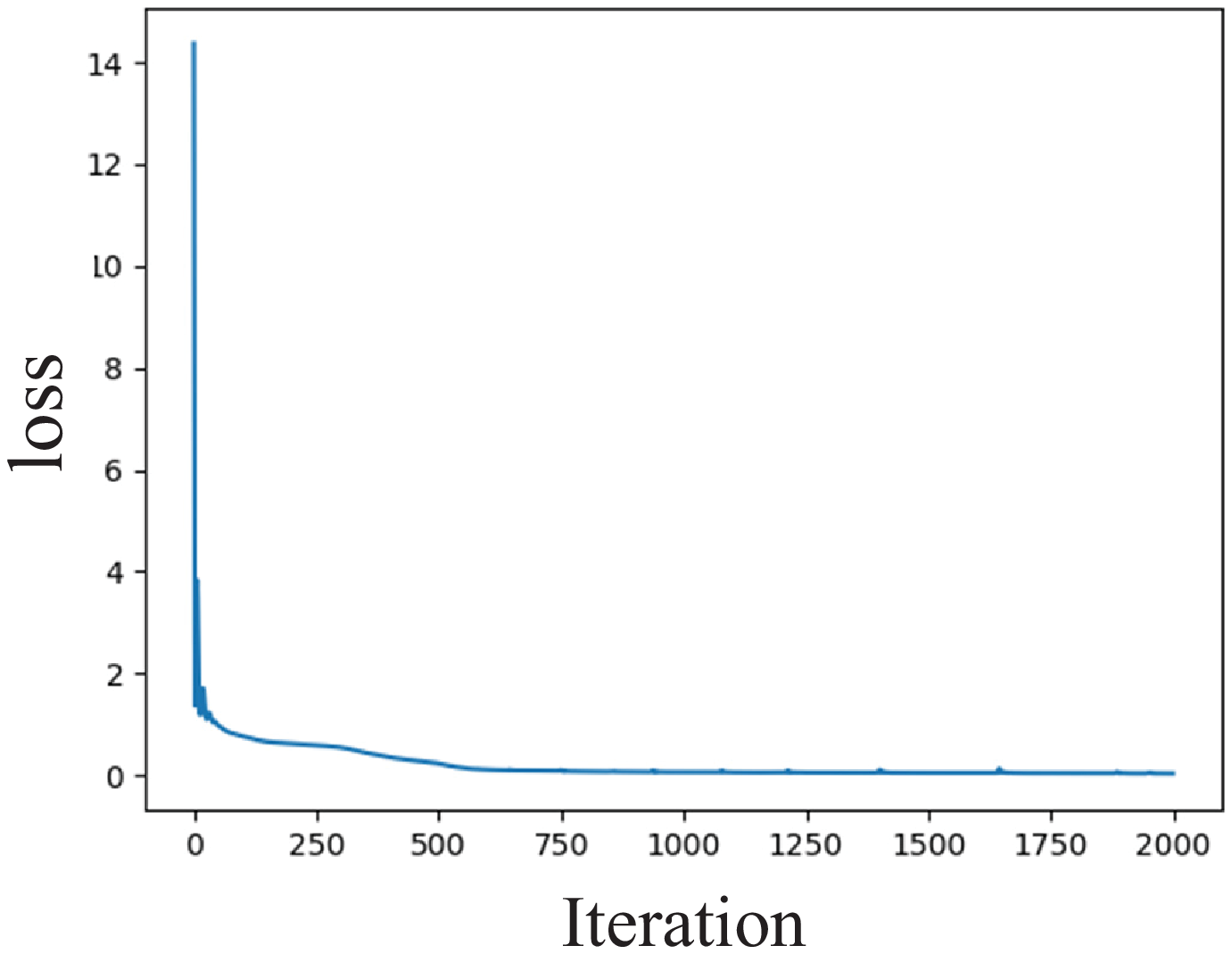}}
\subfigure[TgNN: $f1\_loss$]{\includegraphics[width=0.30\textwidth]{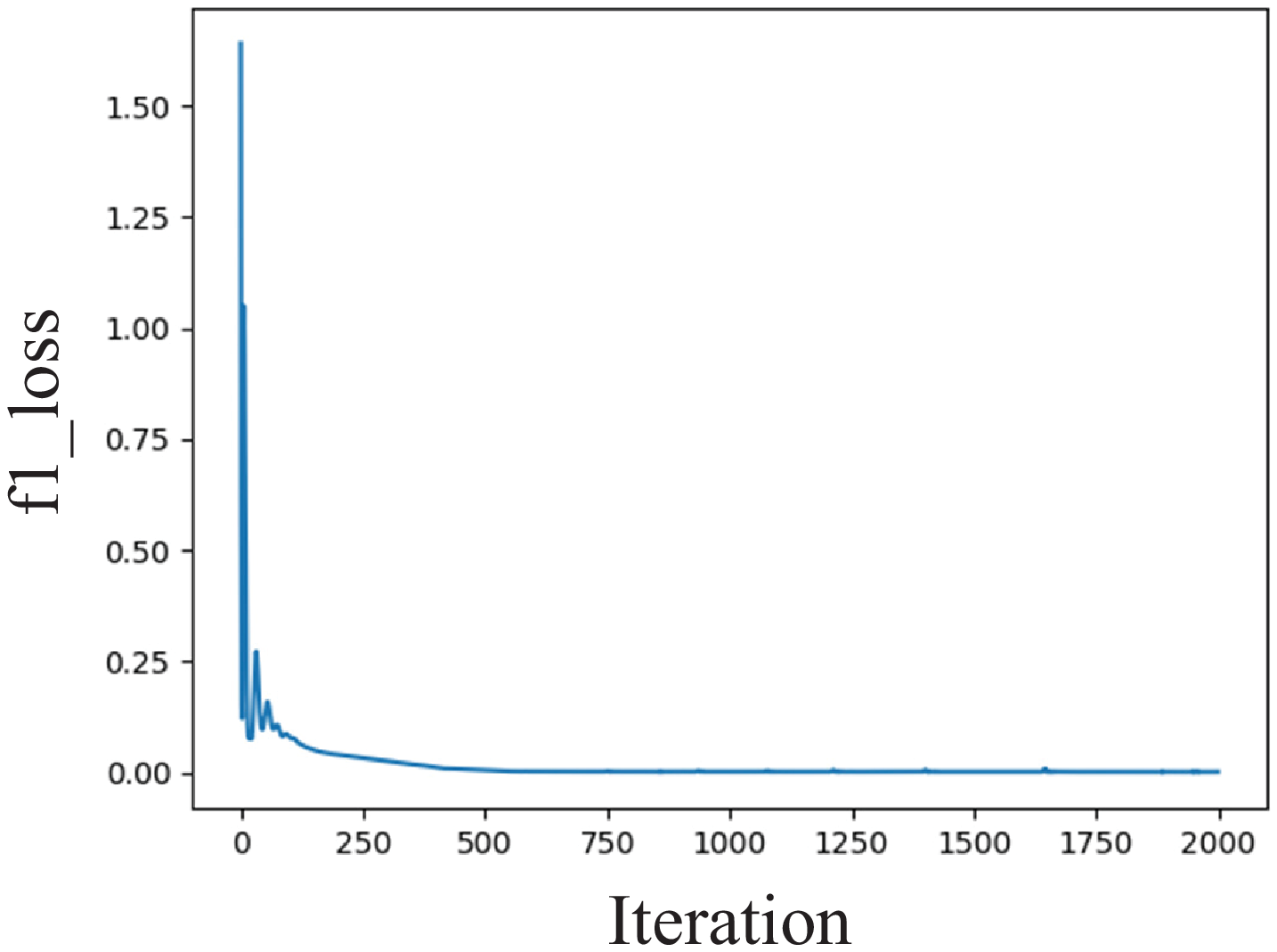}}
\subfigure[TgNN: $f2\_loss$]{\includegraphics[width=0.30\textwidth]{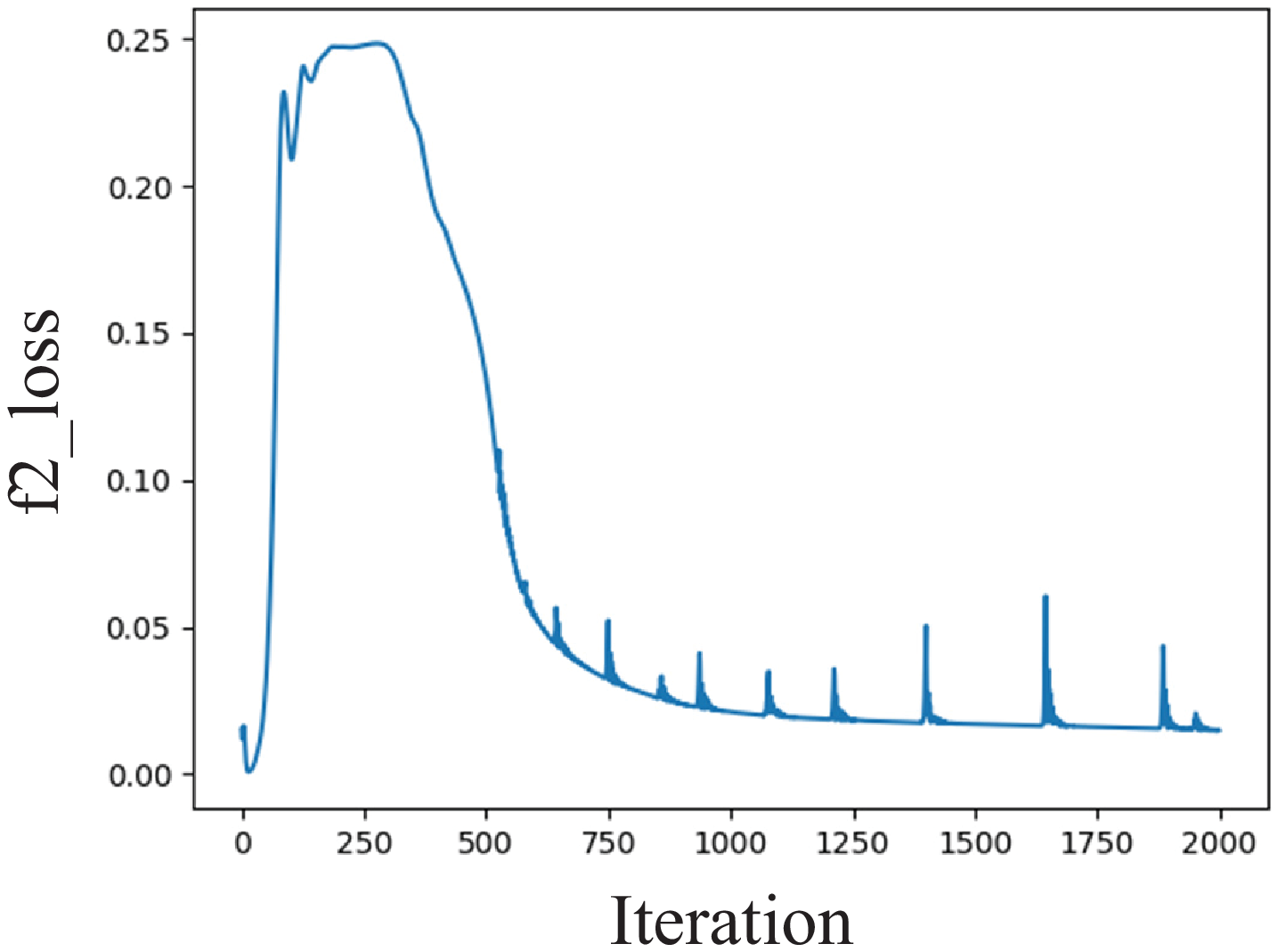}}\\
\subfigure[TgNN-1: $loss$]{\includegraphics[width=0.30\textwidth]{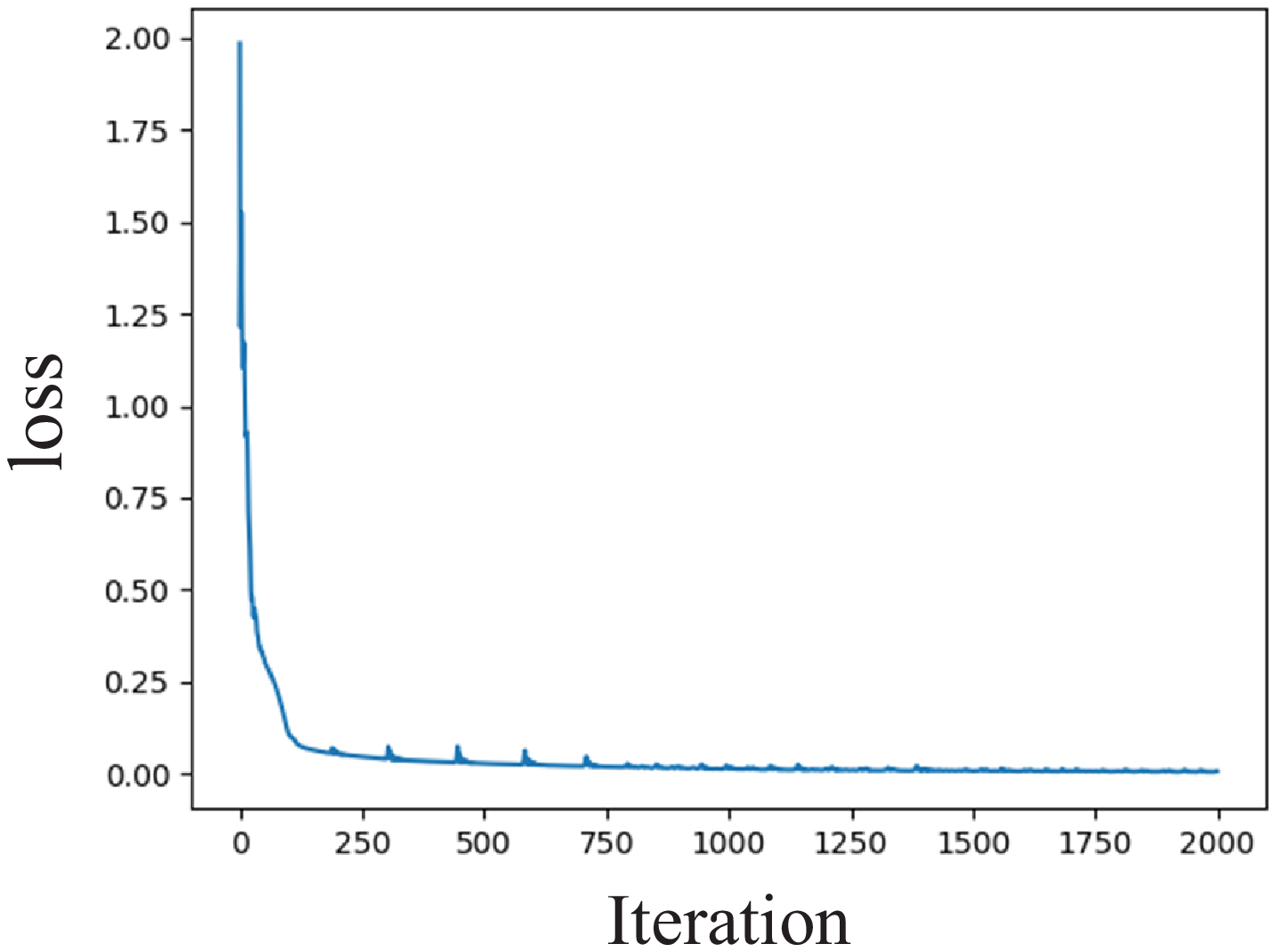}}
\subfigure[TgNN-1: $f1\_loss$]{\includegraphics[width=0.30\textwidth]{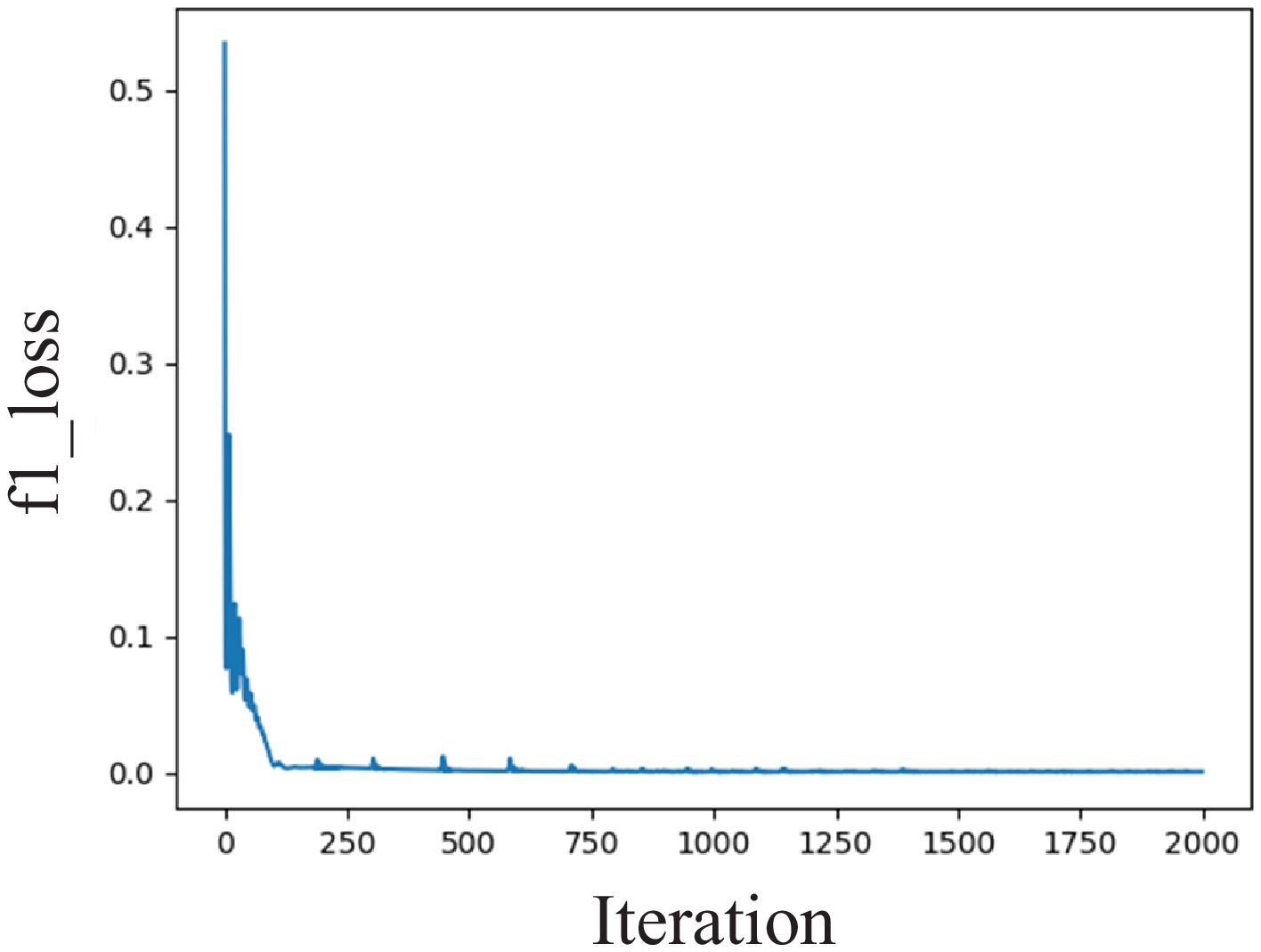}}
\subfigure[TgNN-1: $f2\_loss$]{\includegraphics[width=0.30\textwidth]{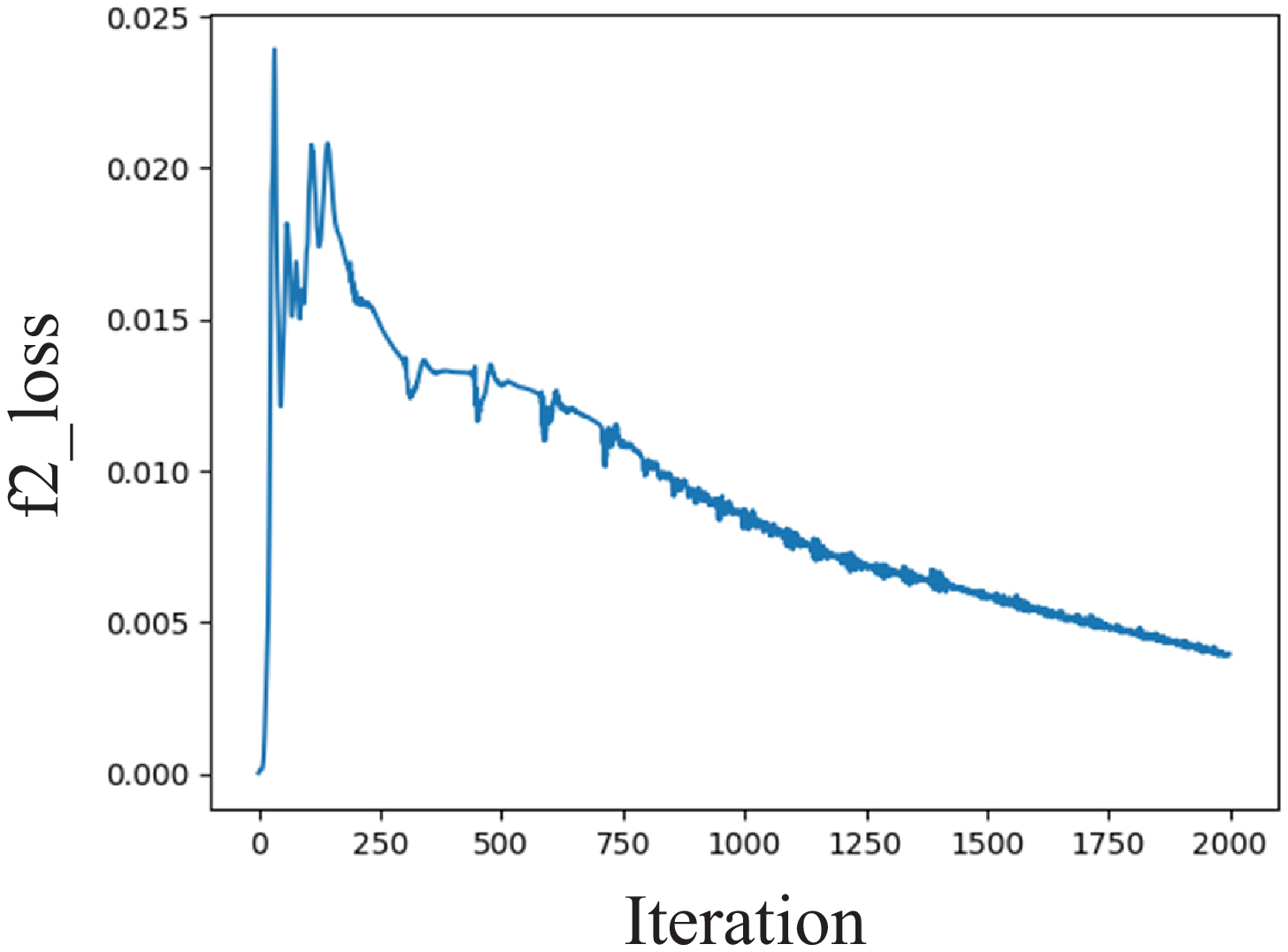}}
\caption{Changes of per loss term versus iteration with the total iterations number of 2000.}
\label{fig1}
\end{center}
\end{figure*}

In order to observe the changing trend of each loss, we plot the loss values versus iterations for each method, as shown in Fig.\ref{fig1}. Since TgNN places more emphasis on the PDE term, herein, we only provide changes of the total loss (denoted as $loss$), the data term (denoted as $f1\_loss$), and the PDE term (denoted as $f2\_loss$).
\begin{figure*}[htb]
\begin{center}
\subfigure[TgNN-LD]{\includegraphics[width=0.3\textwidth]{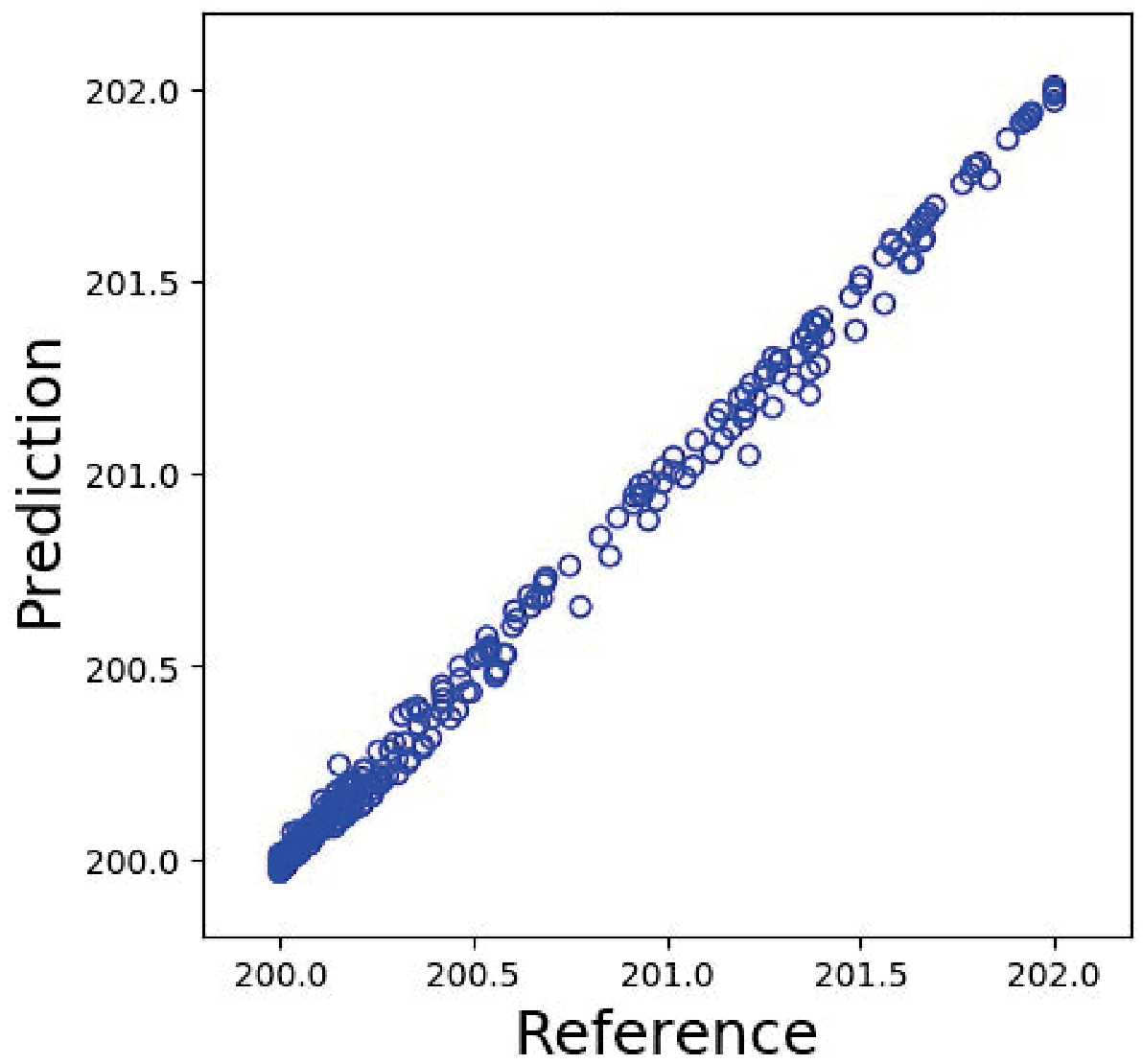}}
\subfigure[TgNN]{\includegraphics[width=0.30\textwidth]{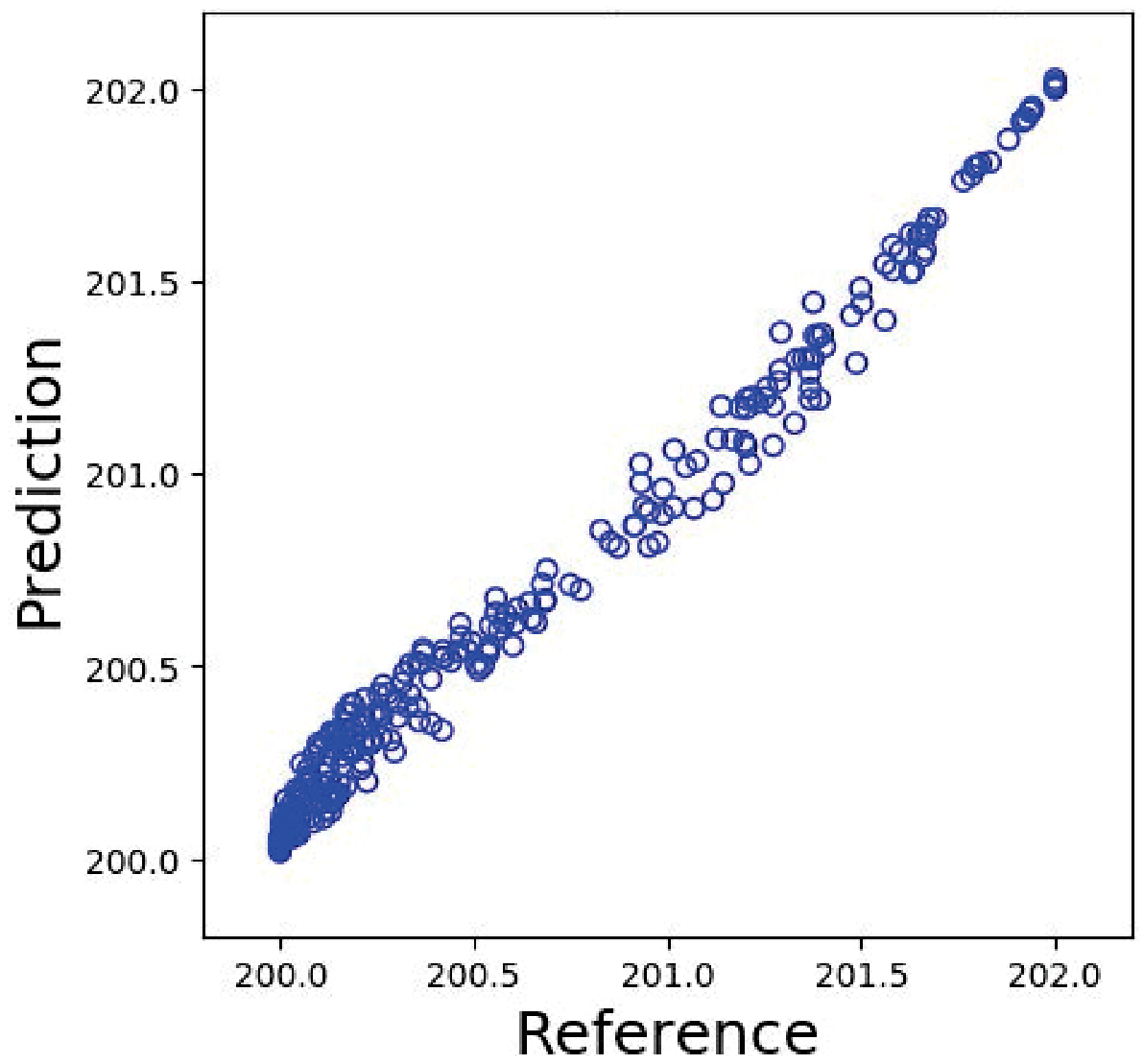}}
\subfigure[TgNN-1]{\includegraphics[width=0.30\textwidth]{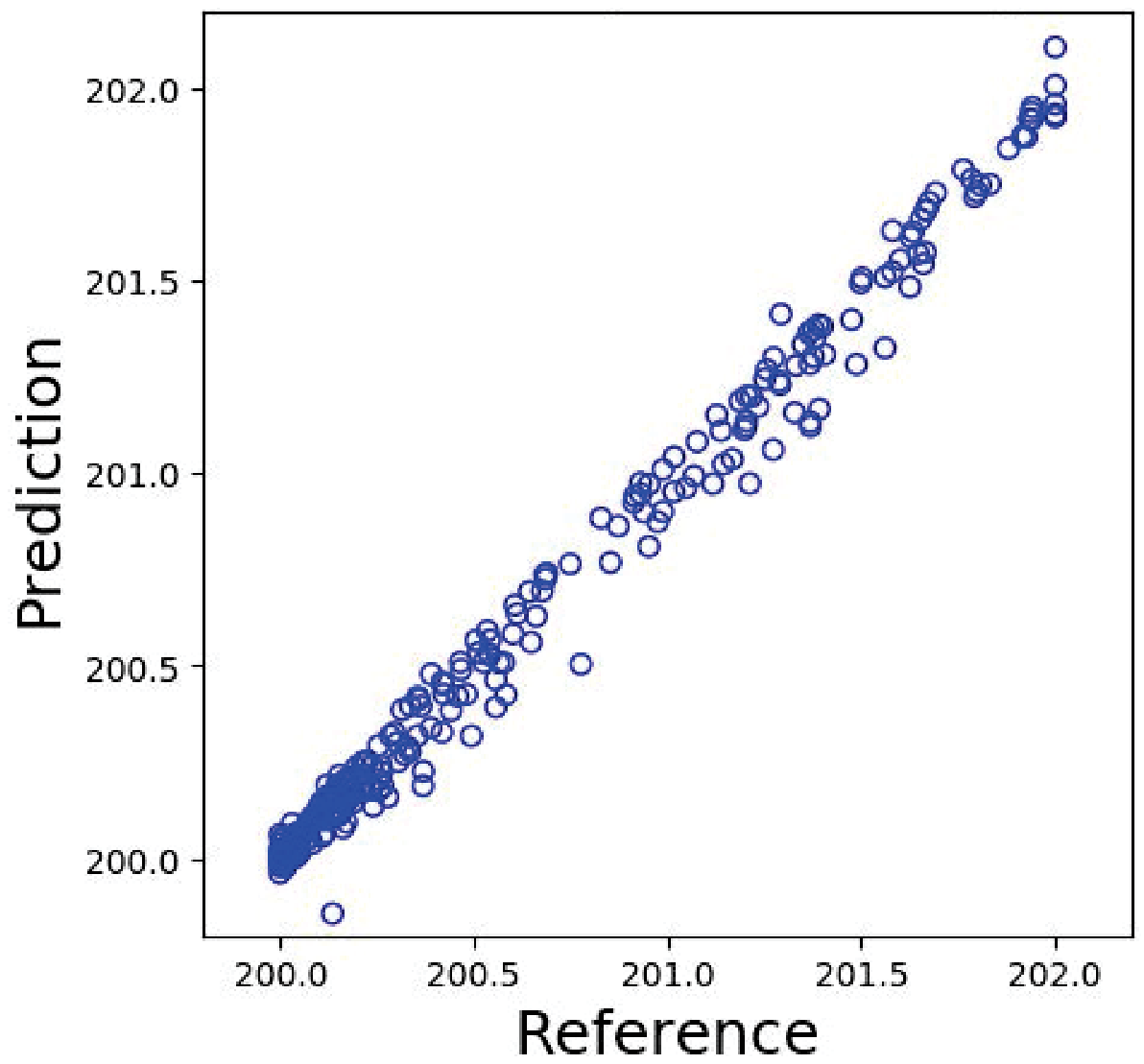}}\\
\caption{Correlation between the reference and predicted hydraulic head with the iteration number of 2000.}
\label{fig2}
\end{center}
\end{figure*}

It can be seen from Fig.\ref{fig1} that TgNN-LD can obtain losses with less fluctuation, and TgNN takes second place, and TgNN-1 achieves the worst results. The most remarkable difference lies in changes of $f2\_loss$, which corresponds to the PDE term. The proposed TgNN-LD exhibits much more stable states whereas there are many shocks in both TgNN and TgNN-1. Indeed, in terms of both smoothness and the number of iterations, TgNN-LD achieves superior performance.

\begin{figure*}[htb]
\begin{center}
\subfigure[TgNN-LD: Prediction vs. reference]{\includegraphics[width=0.30\textwidth]{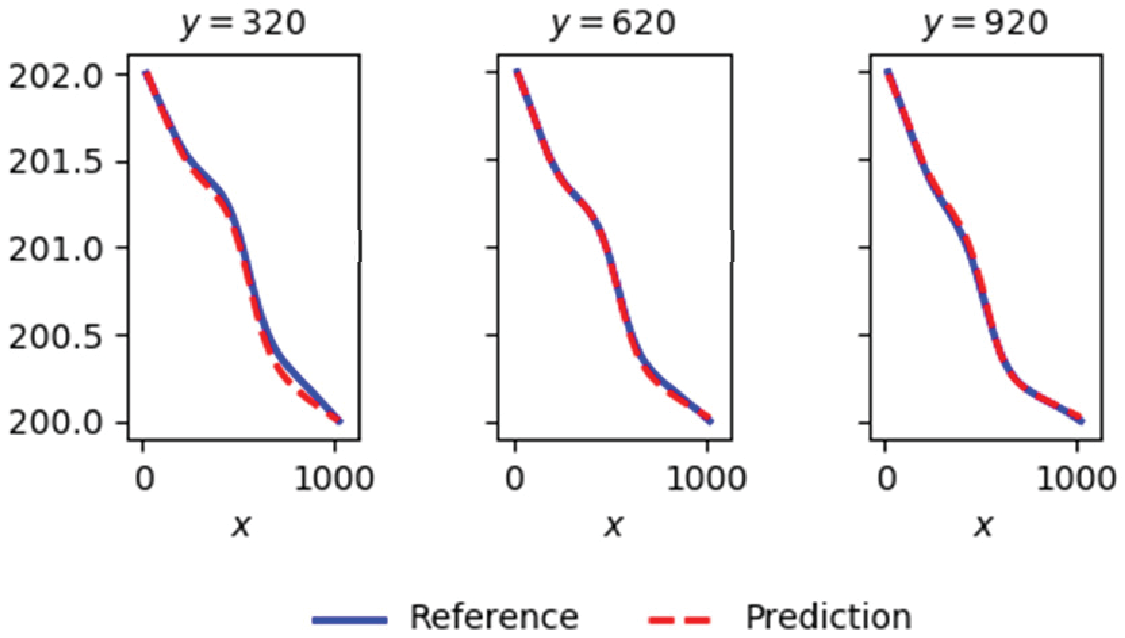}}
\subfigure[TgNN: Prediction vs. reference]{\includegraphics[width=0.30\textwidth]{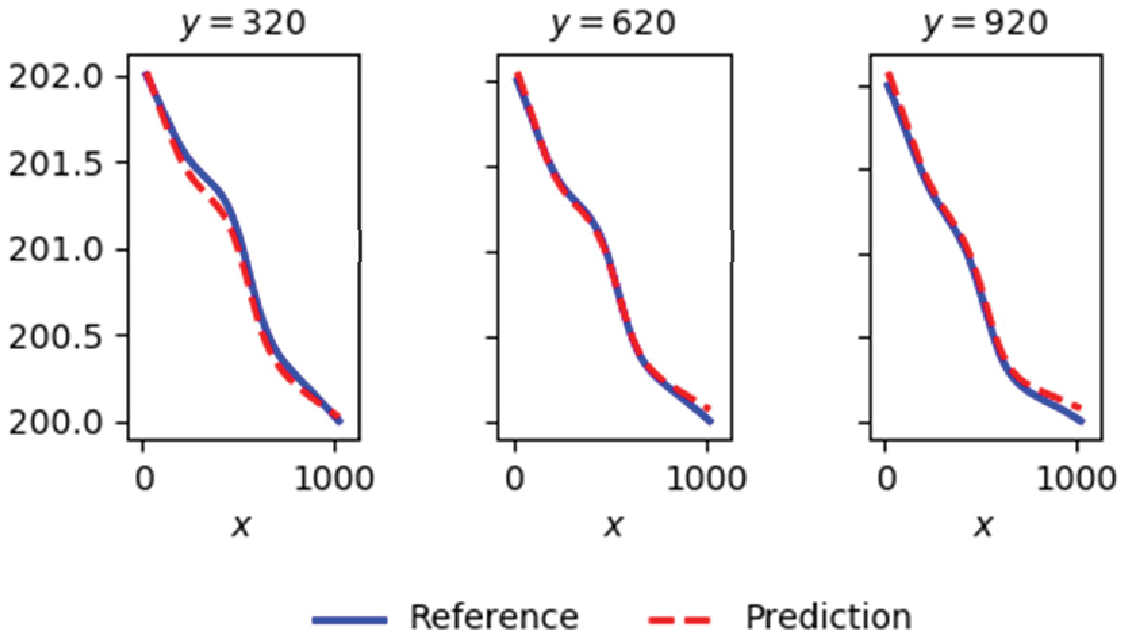}}
\subfigure[TgNN-1: Prediction vs. reference]{\includegraphics[width=0.30\textwidth]{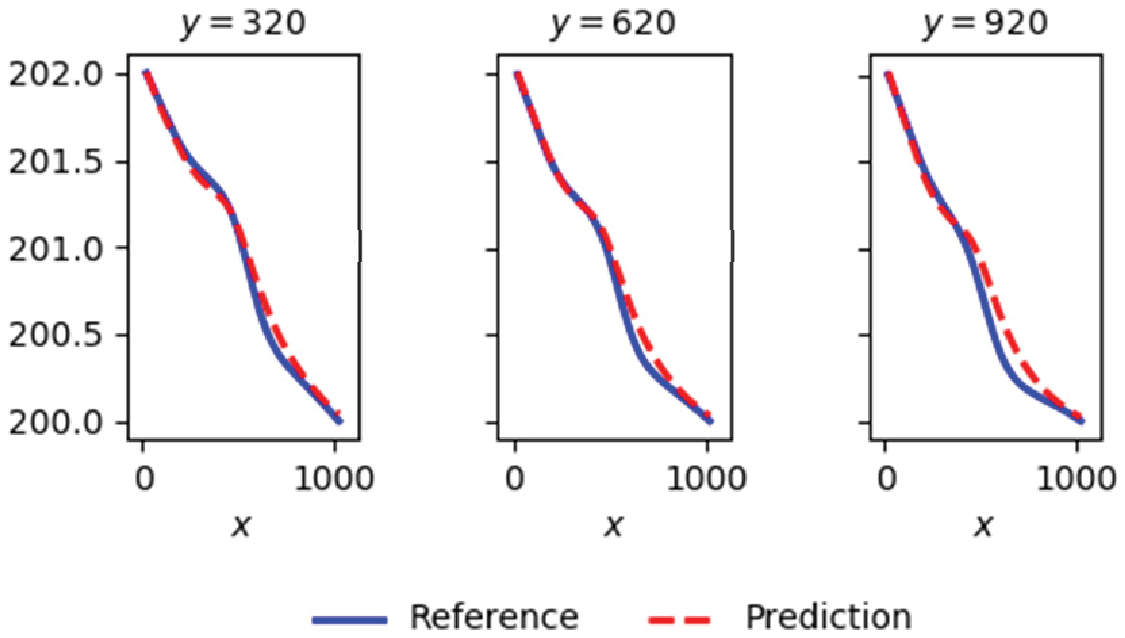}}\\
\subfigure[TgNN-LD: Predicted $h$ at $t=50$]{\includegraphics[width=0.30\textwidth]{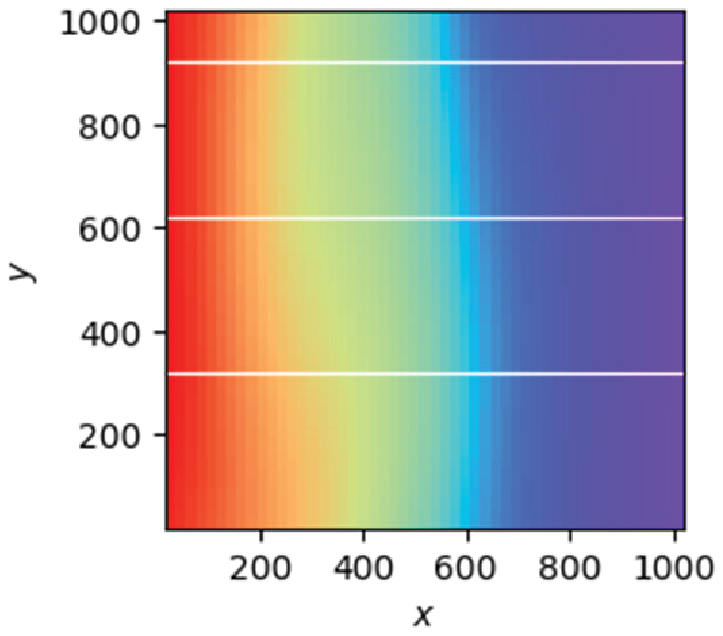}}
\subfigure[TgNN: Predicted $h$ at $t=50$]{\includegraphics[width=0.30\textwidth]{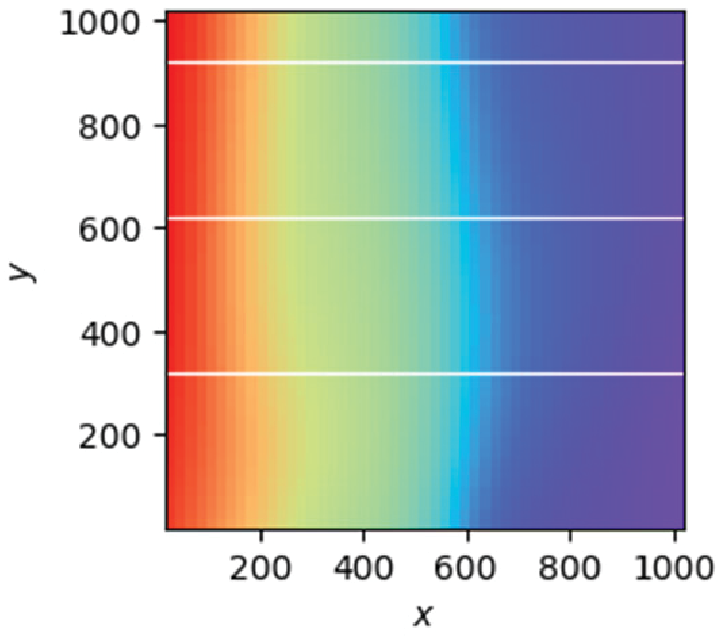}}
\subfigure[TgNN-1: Predicted $h$ at $t=50$]{\includegraphics[width=0.36\textwidth]{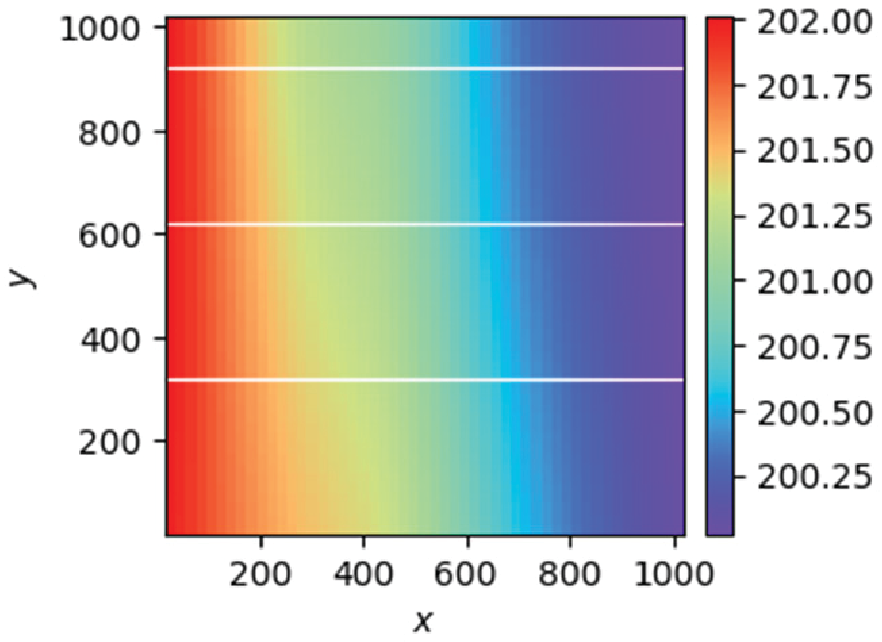}}\\
\caption{Prediction results of TgNN-LD, TgNN, and TgNN-1.}
\label{fig3}
\end{center}
\end{figure*}

Fig.\ref{fig2} compares the correlation between the reference and predicted hydraulic head with the iteration number of 2000. Fig.\ref{fig3} provides prediction results of TgNN-LD, TgNN, and TgNN-1. From these figures, it can be clearly seen that the prediction of TgNN-LD matches the reference values well and are superior to the predictions of the other two.
\subsection{Reduced training epochs}
From Figs.\ref{fig1}.(j)-(l), it can be observed that when the number of iterations is approximately 1750, three losses are approaching to converge, suggesting that the number of iterations could be reduced to shorten the training time. To verify its effect, we set different numbers of iterations to test our method. The related results are presented in Tab.\ref{tab:table3results_with_diff_iter}.
\begin{table*}[htbp]
  \centering
  \caption{Results versus different numbers of iterations.}
    \begin{tabular}{cccccccccccccccccccccccccccc}
    \toprule
    Number of iterations&Error L2&	R2	& Training time/s\\
    \midrule
    1500	& 3.5398E-04	&9.8753E-01	&\textbf{149.6769}\\
    \midrule
    1700&	2.4923E-04	&9.9382E-01&	159.2913\\
    \midrule
     1750	&\textbf{1.9887E-04}	&\textbf{9.9606E-01}&	164.3937\\
    \midrule
    1800	&2.3961E-04	&9.9429E-01&	168.9544\\
    \midrule
    2000	&2.0833E-04	&9.9532E-01&	204.3223\\
    \bottomrule
    \end{tabular}%
  \label{tab:table3results_with_diff_iter}%
\end{table*}%

As shown in Tab.\ref{tab:table3results_with_diff_iter}, when the number of iteration is 1750, even though the training time is 15s longer than that obtained with the iteration number of 1500, the error L2 and R2 results exhibit the best performance among the compared sets. Especially, in comparison with a result iteration number of 2000, the training time almost decreases by 20\%, while the error L2 and R2 results are superior.
\subsection{Changes of Lagrangian multipliers}
\begin{figure*}[htb]
\begin{center}
\subfigure[Total epoch=1700: $\lambda _{PDE}$]{\includegraphics[width=0.30\textwidth]{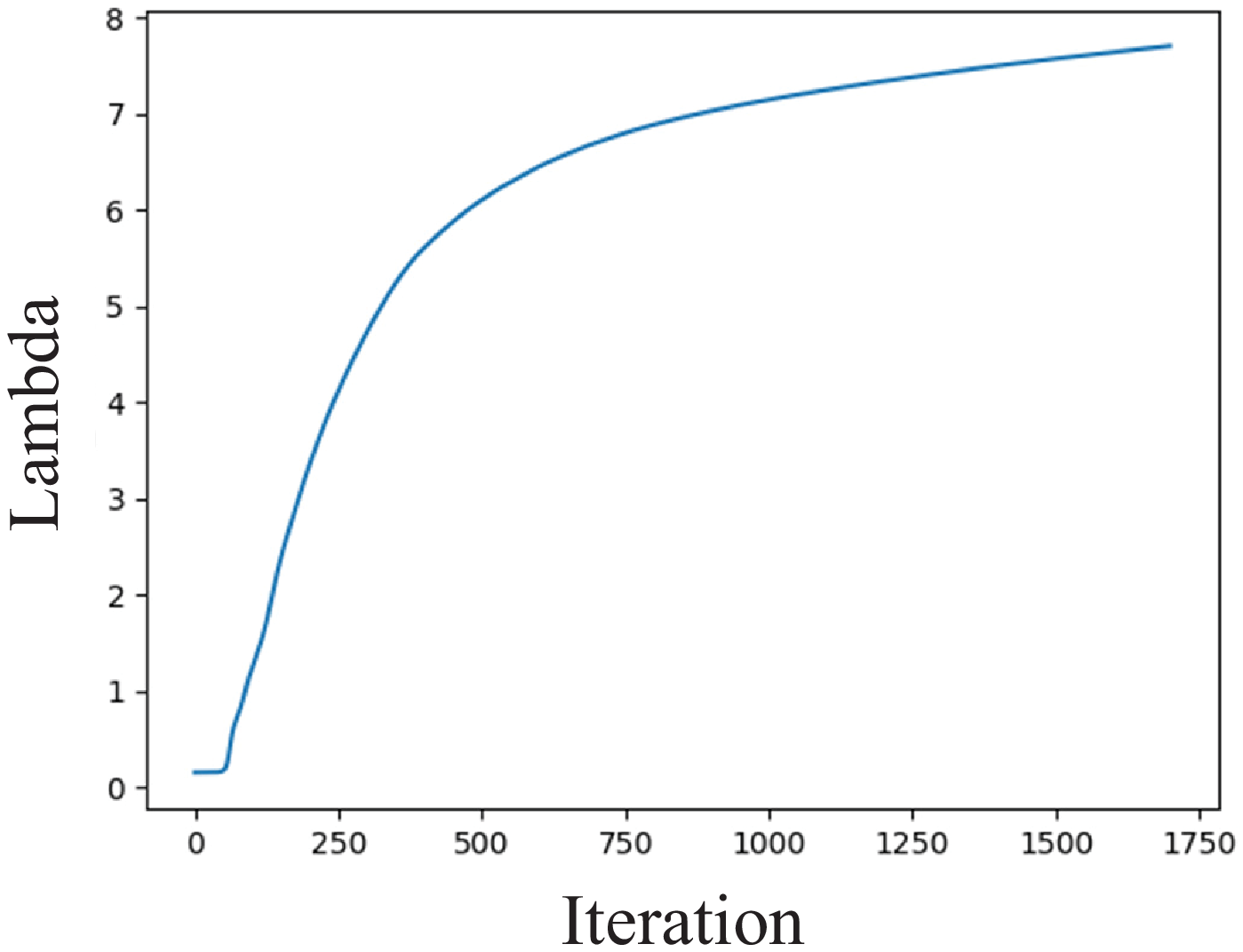}}
\subfigure[Total epoch=1700: $\lambda _{EC}$]{\includegraphics[width=0.30\textwidth]{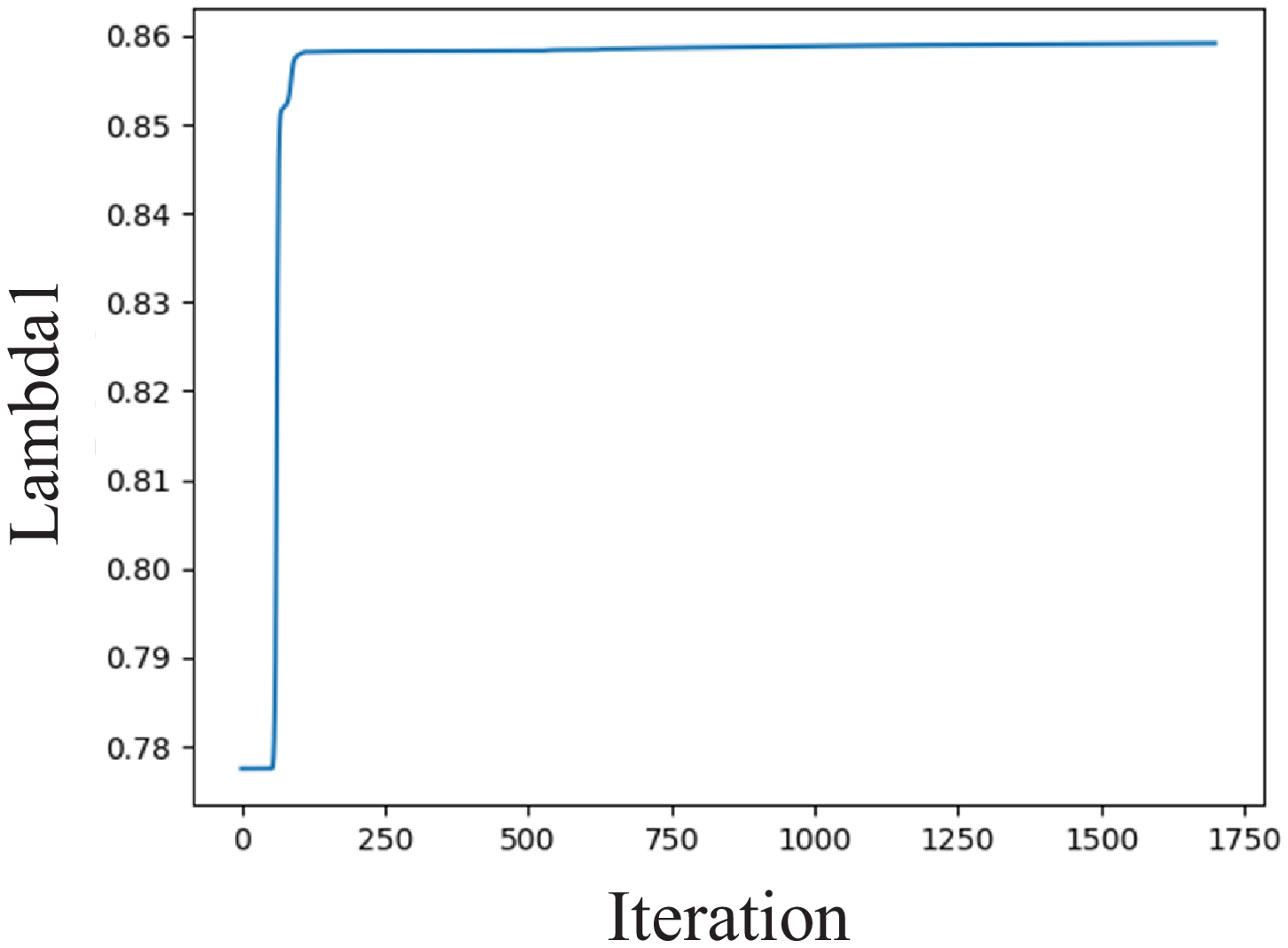}}
\subfigure[Total epoch=1700: $\lambda _{EK}$]{\includegraphics[width=0.30\textwidth]{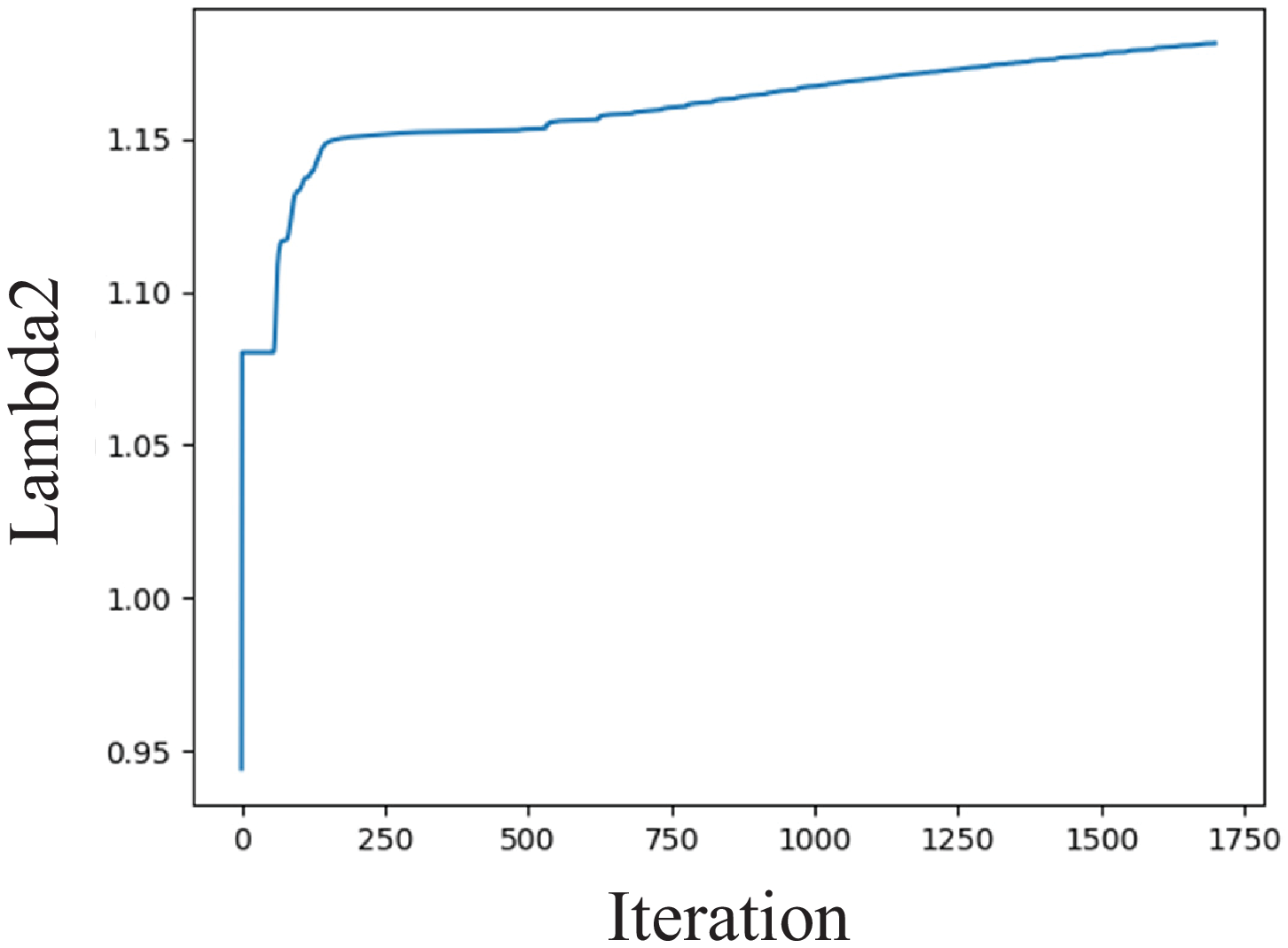}}\\
\subfigure[Total epoch=1750: $\lambda _{PDE}$]{\includegraphics[width=0.30\textwidth]{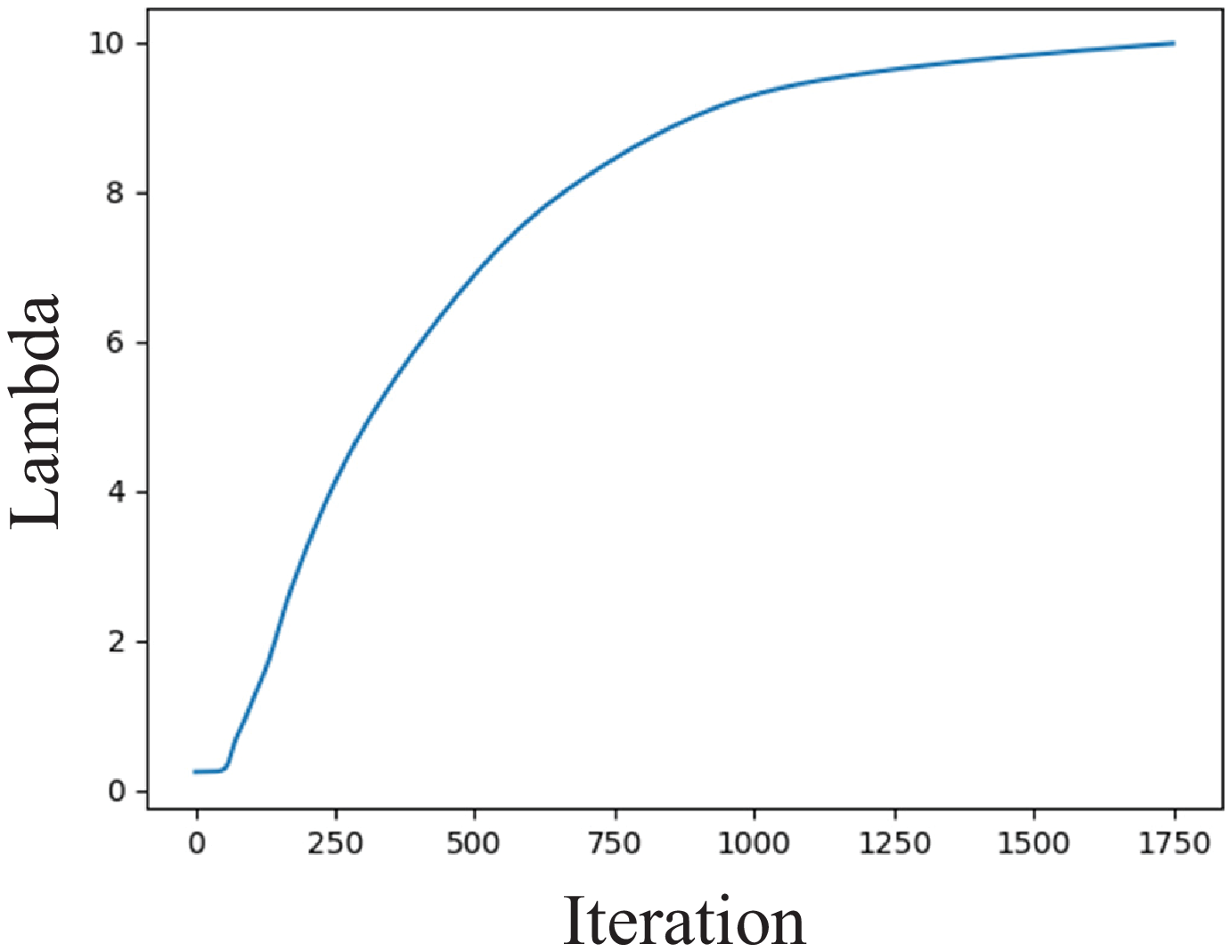}}
\subfigure[Total epoch=1750: $\lambda _{EC}$]{\includegraphics[width=0.30\textwidth]{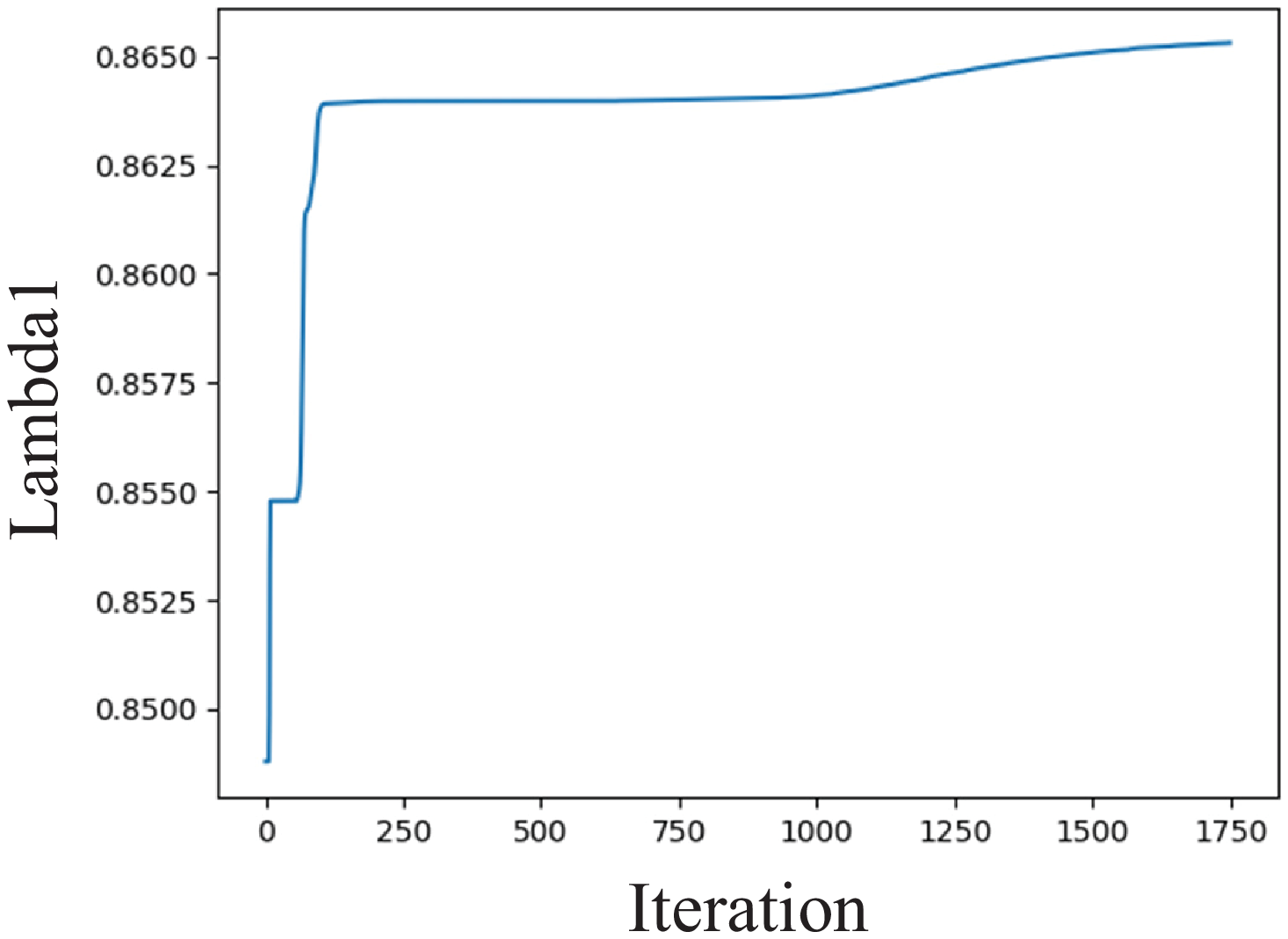}}
\subfigure[Total epoch=1750: $\lambda _{EK}$]{\includegraphics[width=0.30\textwidth]{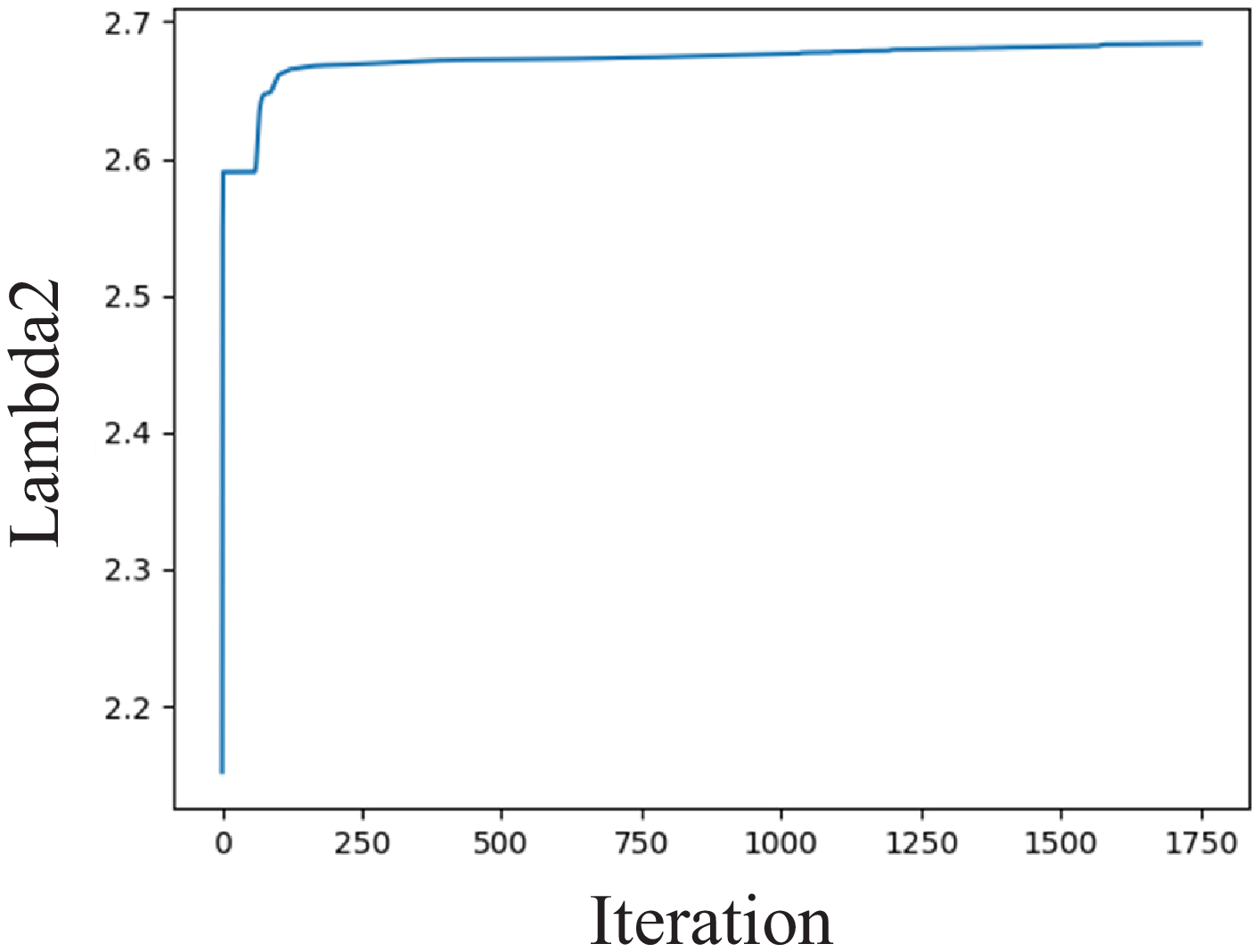}}\\
\subfigure[Total epoch=1800: $\lambda _{PDE}$]{\includegraphics[width=0.30\textwidth]{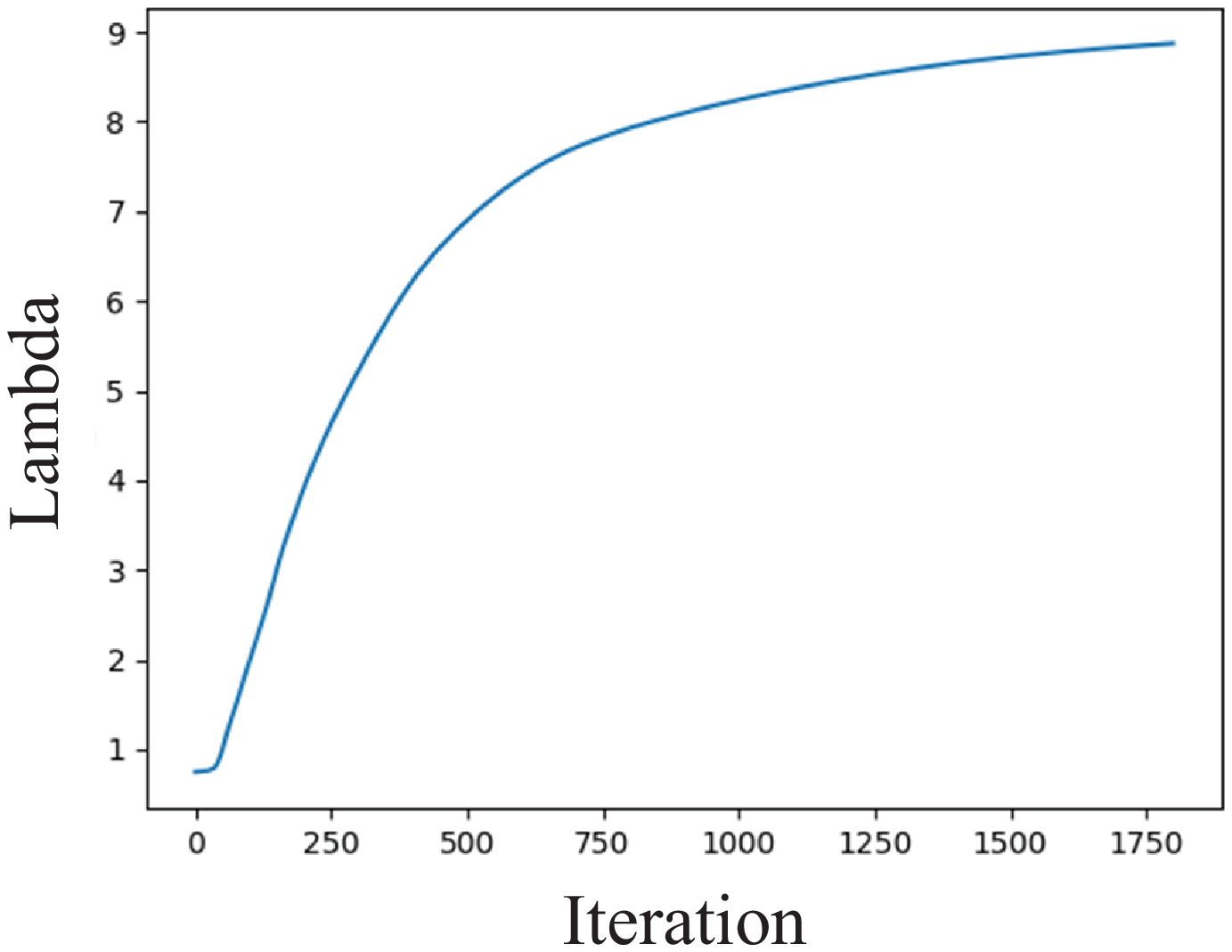}}
\subfigure[Total epoch=1800: $\lambda _{EC}$]{\includegraphics[width=0.30\textwidth]{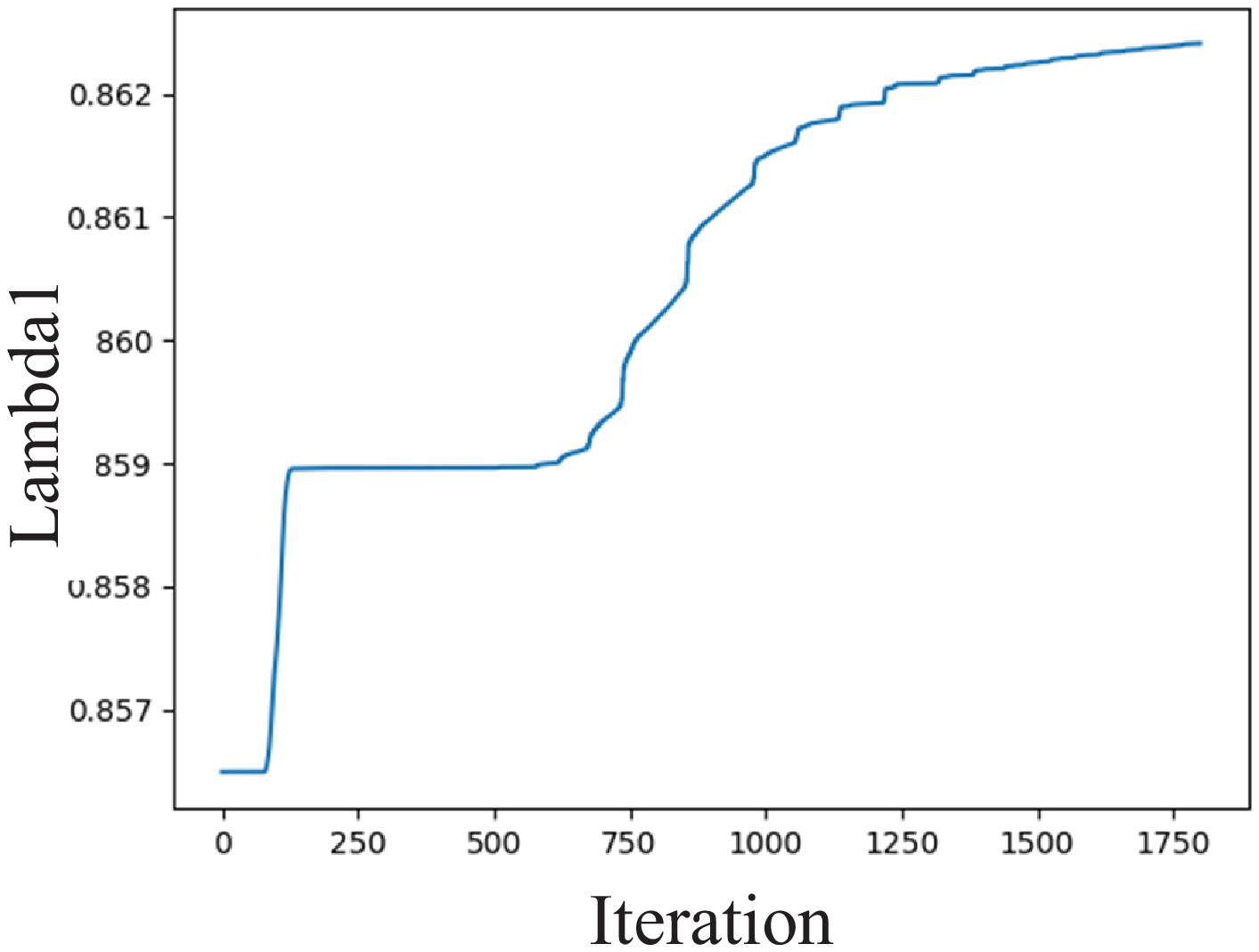}}
\subfigure[Total epoch=1800: $\lambda _{EK}$]{\includegraphics[width=0.30\textwidth]{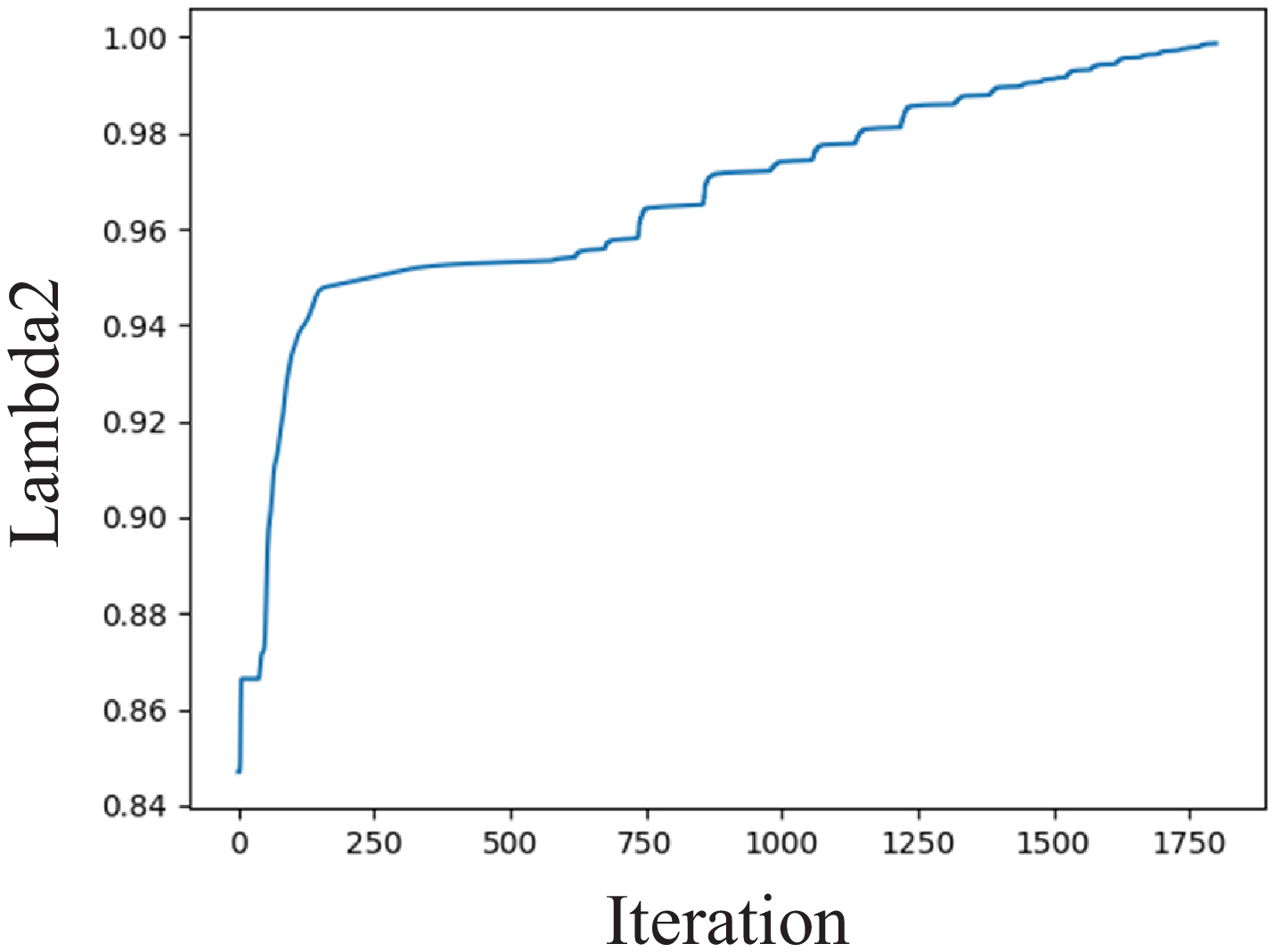}}\\
\subfigure[Total epoch=2000: $\lambda _{PDE}$]{\includegraphics[width=0.3\textwidth]{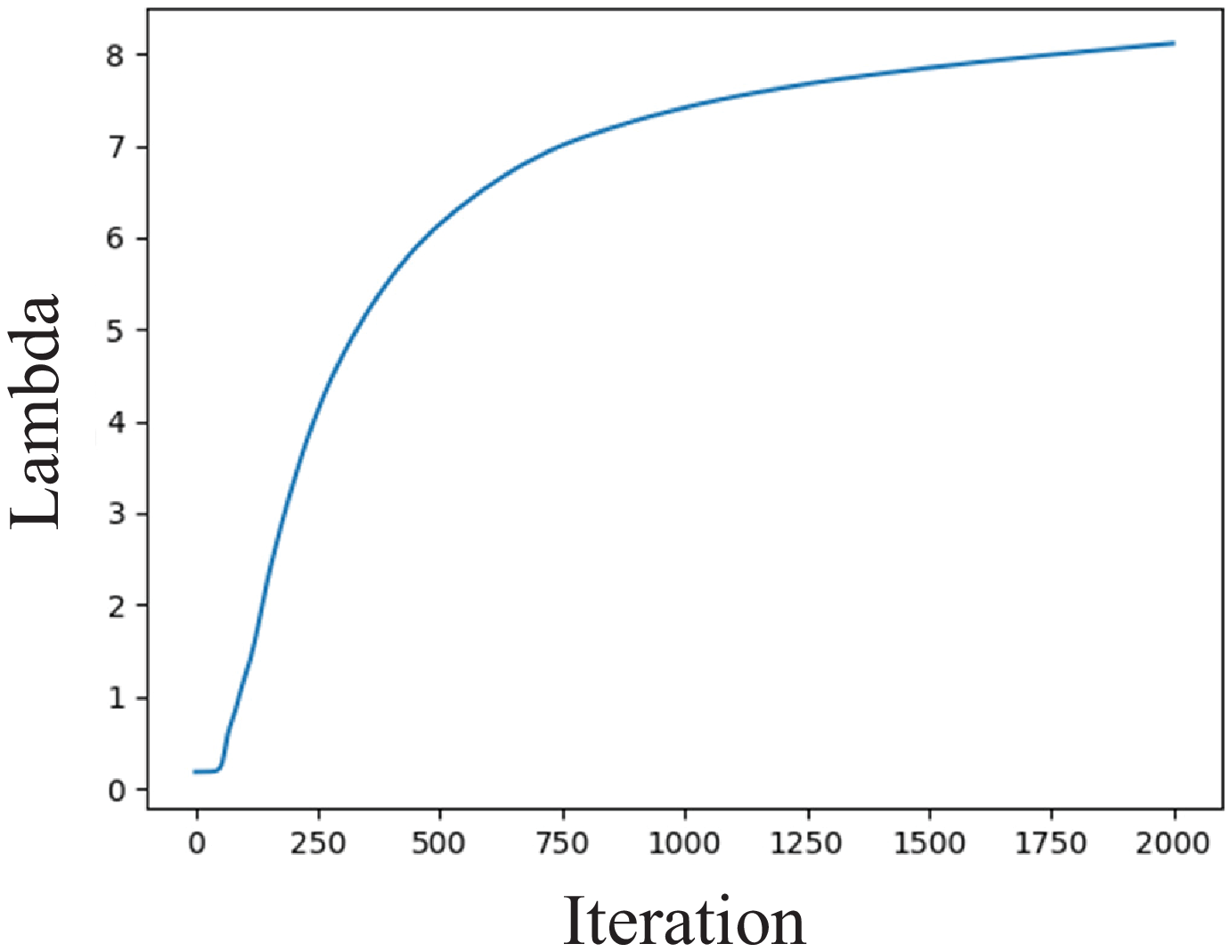}}
\subfigure[Total epoch=2000: $\lambda _{EC}$]{\includegraphics[width=0.30\textwidth]{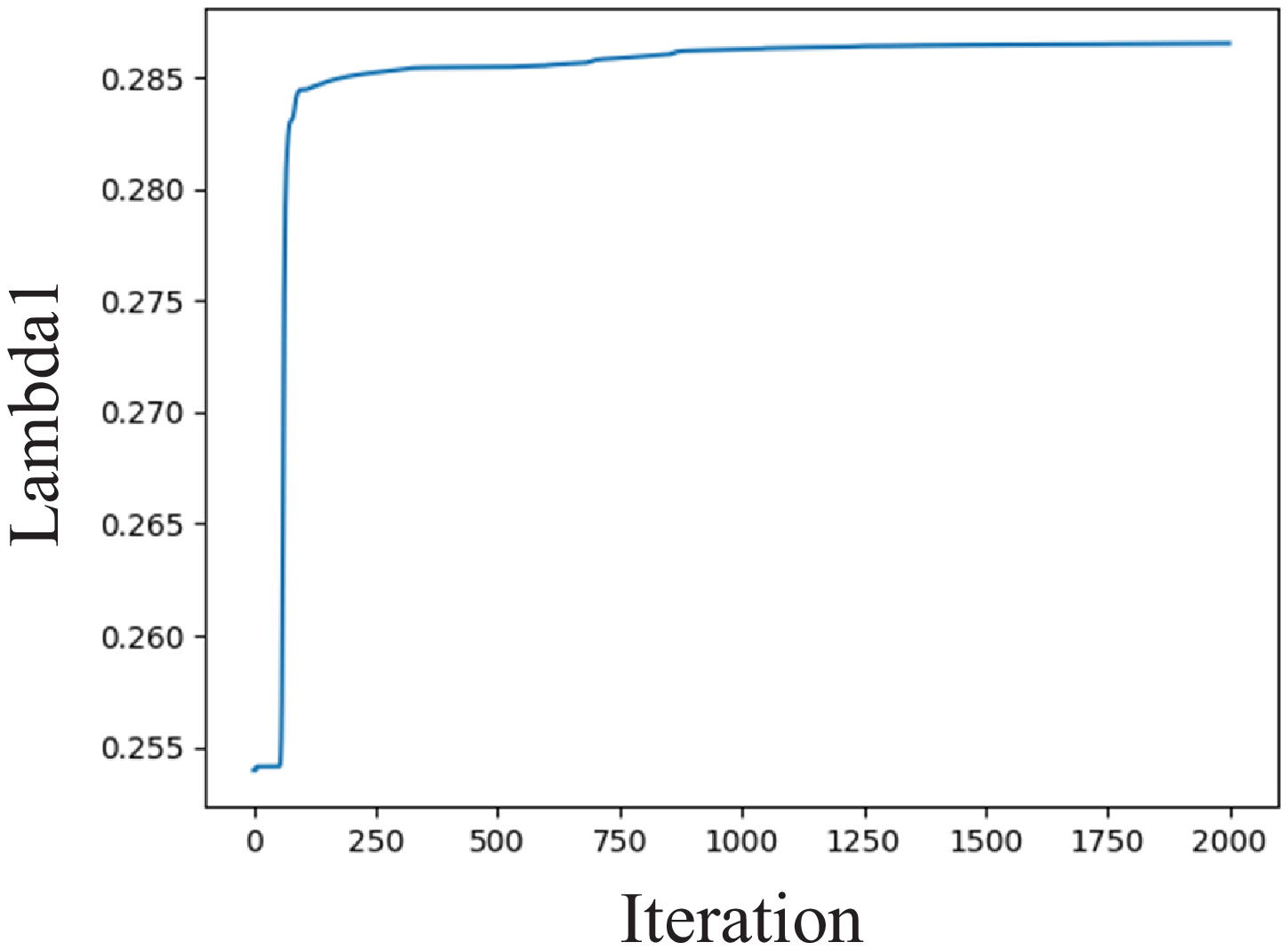}}
\subfigure[Total epoch=2000: $\lambda _{EK}$]{\includegraphics[width=0.30\textwidth]{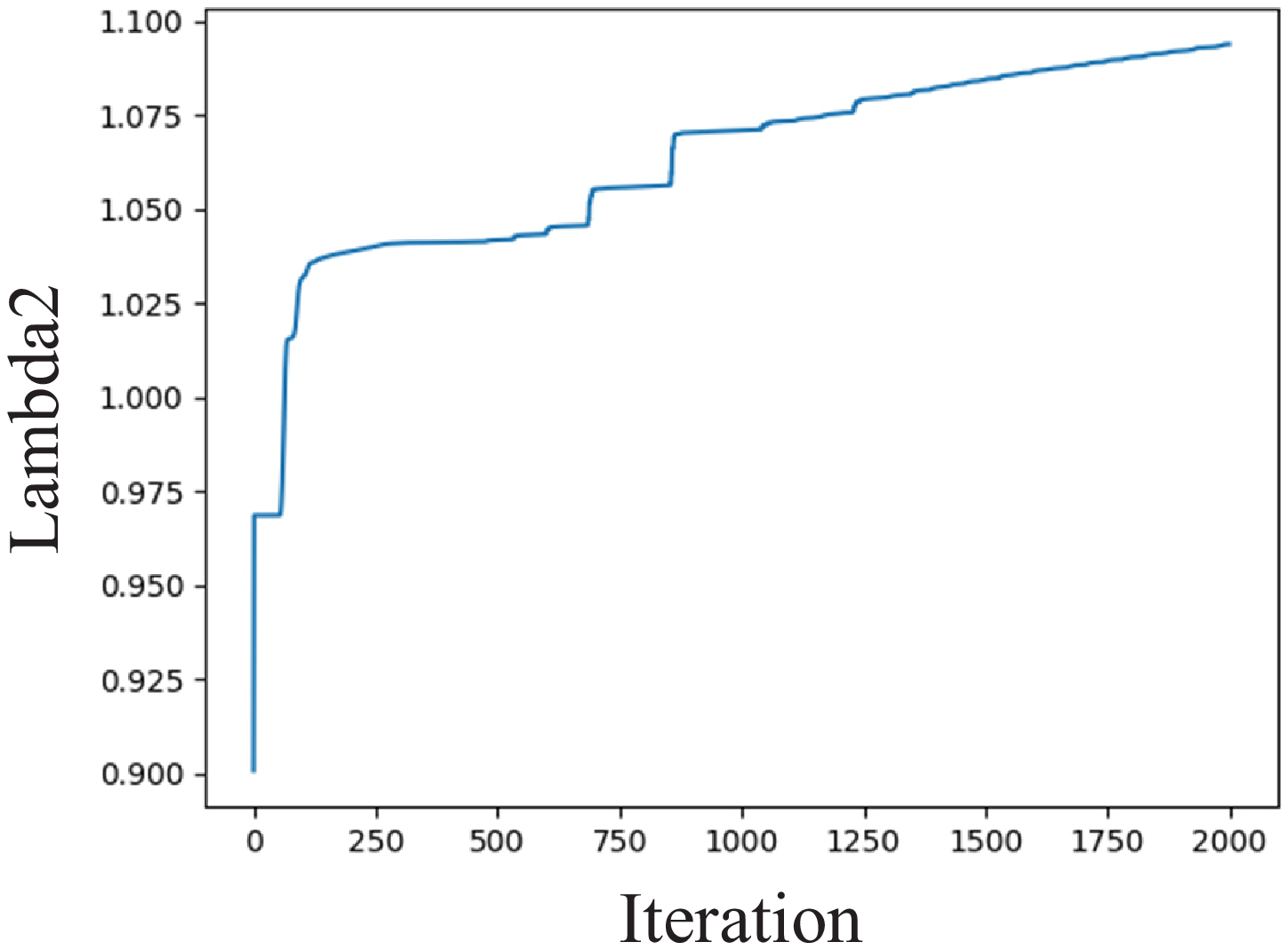}}\\
\caption{Changes of Lagrangian multipliers versus iteration.}
\label{fig4}
\end{center}
\end{figure*}

To further investigate the effect of Lagrangian multipliers under different iteration values, we plot their changes with iterations, as shown in Fig.\ref{fig4}, where Lambda, Lambda1, and Lambda2 refer to $\lambda _{PDE}$, $\lambda _{EC}$, and $\lambda _{EK}$, respectively.

It can be seen from Fig.\ref{fig4} that these Lagrangian multipliers cannot converge to a fixed value, irrespective of the number of iterations, which is rooted in randomly selected seeds leading to various initial values. Tab.\ref{tab:table4Values_of_Lagrangian_multipliers} lists their final values with respect to different settings of iterations.
\begin{table*}[htbp]
  \centering
  \caption{Values of Lagrangian multipliers at the final iteration versus different iteration number.}
    \begin{tabular}{cccccccccccccccccccccccccccc}
    \toprule
Number of iterations&$\lambda _{PDE}$& $\lambda _{EC}$& $\lambda _{EK}$\\
\midrule
1700	&7.7076E+00&	8.5914E-01	&1.1813E+00\\
\midrule
1750	&9.9854E+00	&8.6531E-01&	2.6842E+00\\
\midrule
1800	&8.8720E+00	&8.6241E-01	&9.9864E-01\\
\midrule
2000&	8.1127E+00&	2.8652E-01	&1.0940E+00\\
    \bottomrule
    \end{tabular}%
  \label{tab:table4Values_of_Lagrangian_multipliers}%
\end{table*}%

\begin{figure*}[htb]
\begin{center}
\subfigure[$loss$]{\includegraphics[width=0.3\textwidth]{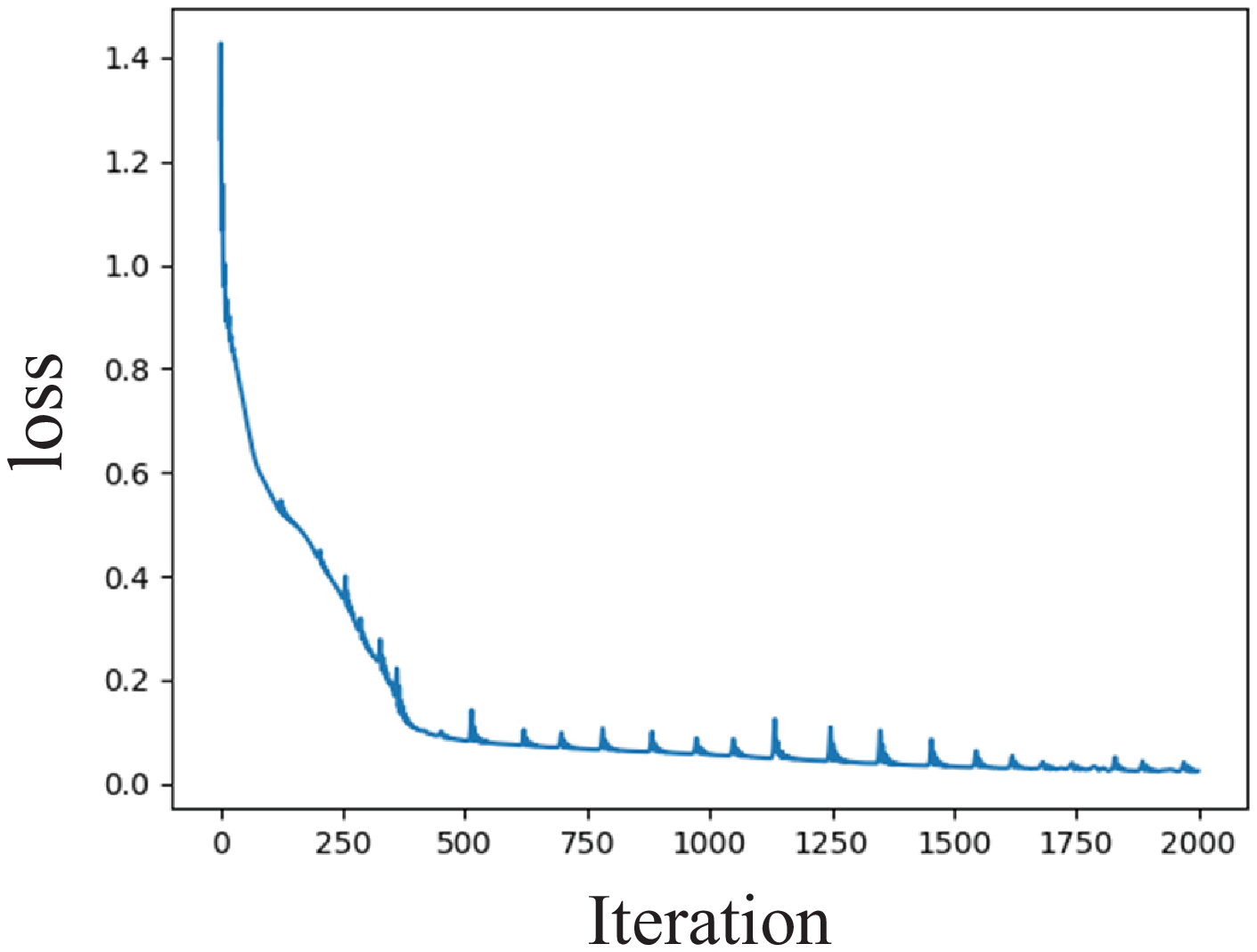}}
\subfigure[$f1\_loss$]{\includegraphics[width=0.30\textwidth]{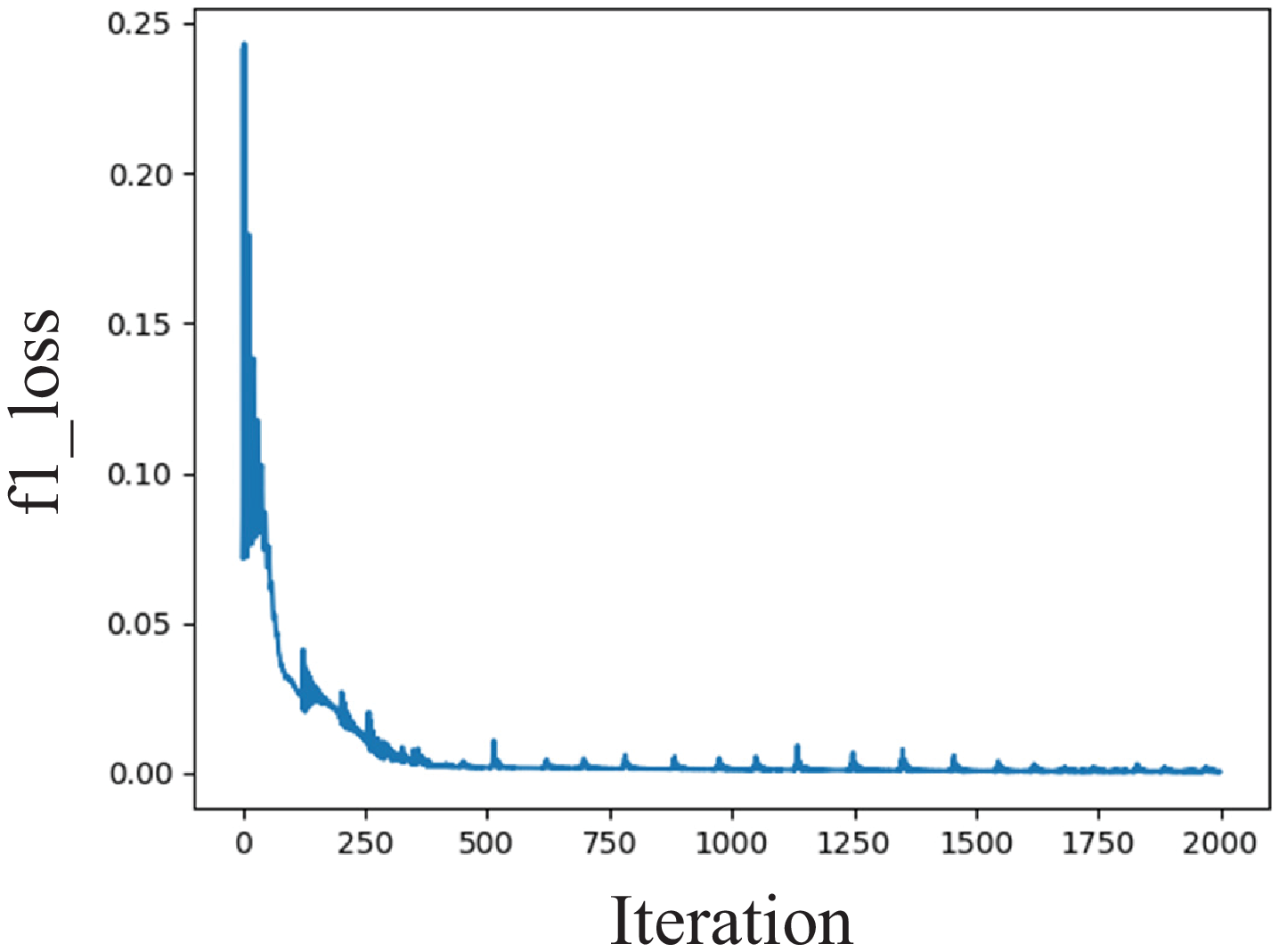}}
\subfigure[$f2\_loss$]{\includegraphics[width=0.30\textwidth]{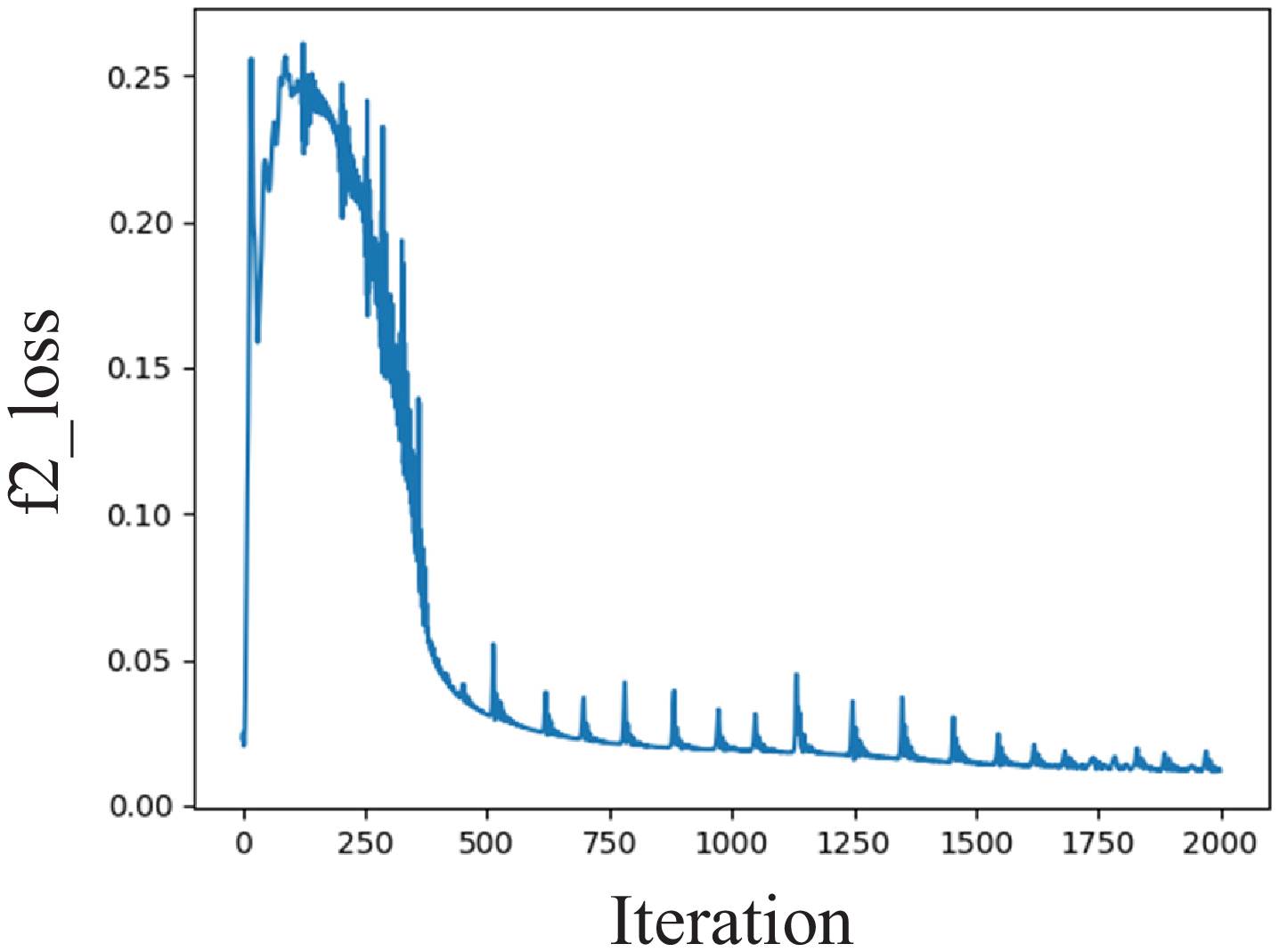}}\\
\subfigure[Correlation between reference and prediction]{\includegraphics[width=0.30\textwidth]{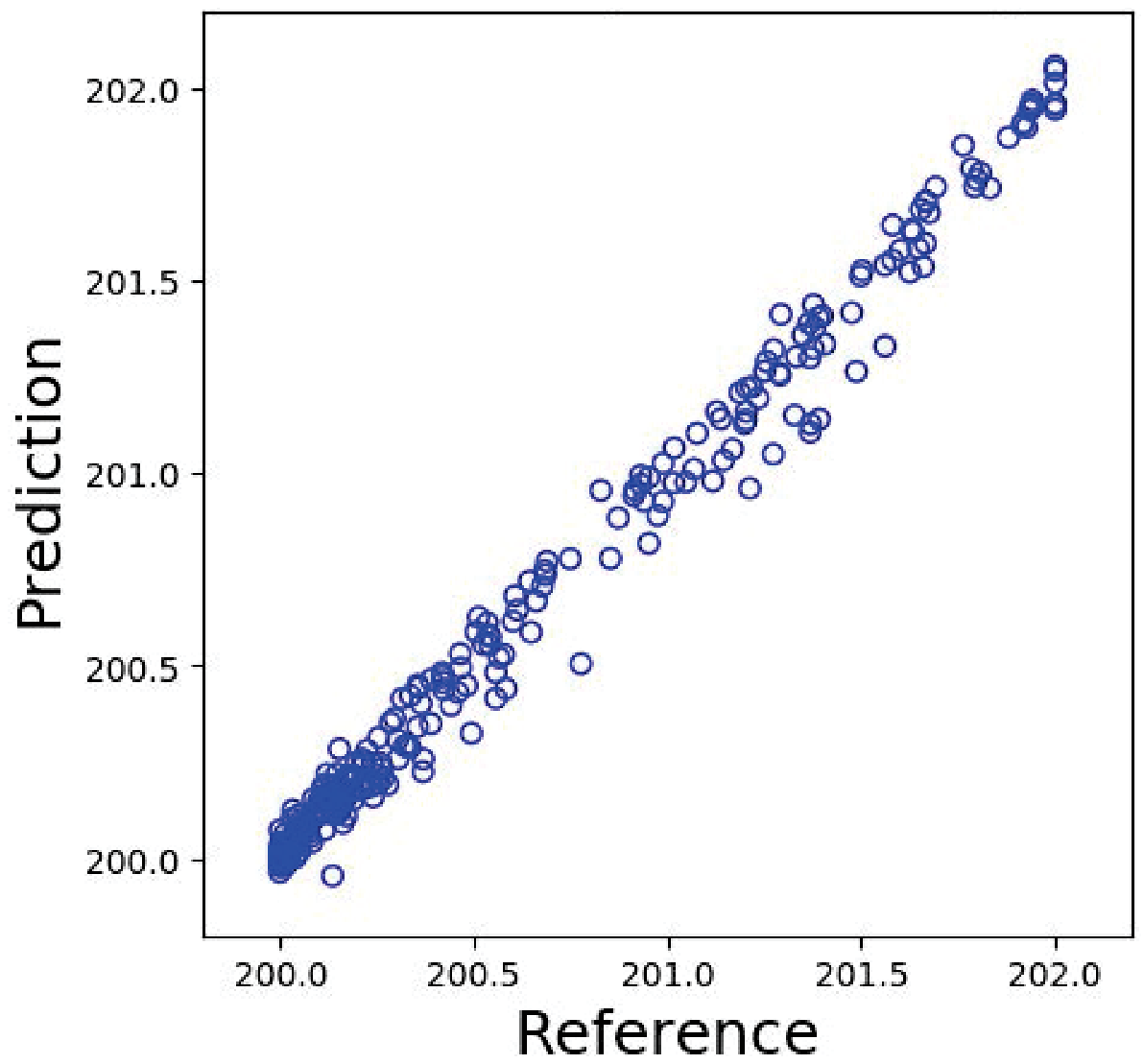}}
\subfigure[Predctition vs reference]{\includegraphics[width=0.50\textwidth]{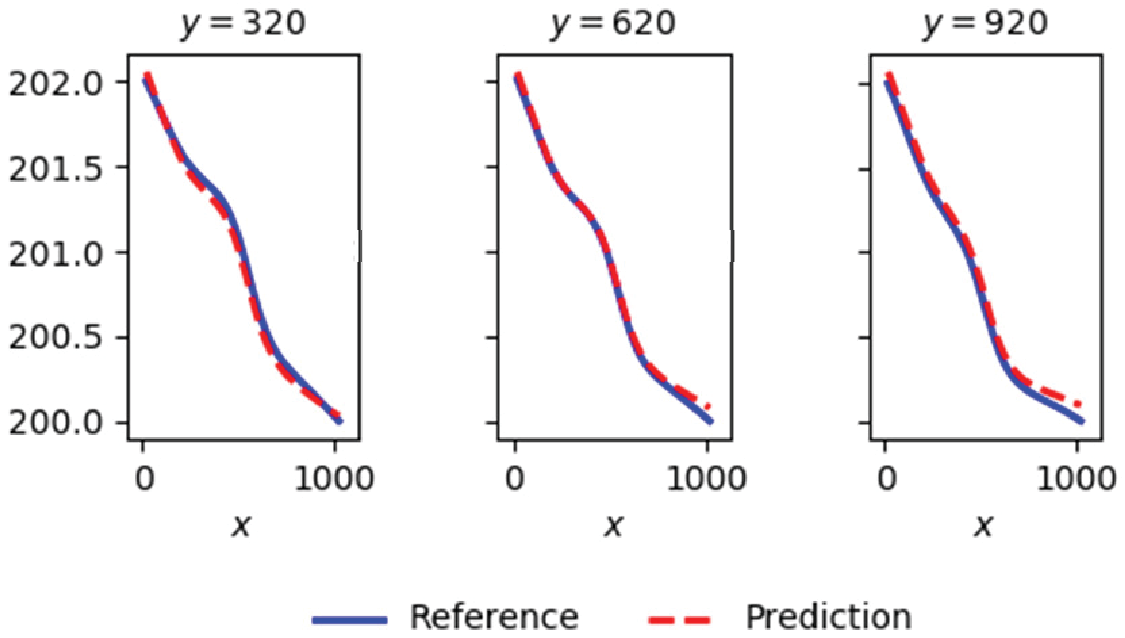}}\\
\caption{Predictive results obtained by TgNN-LD with fixed multiplier values.}
\label{fig5}
\end{center}
\end{figure*}

We then take one set of multipliers (8.1127E+00, 2.8652E-01, and 1.0940E+00) into Eq.\ref{neqf1} and keep them the same during the whole training stage to further verify the above conclusion. Related results are shown in Fig.\ref{fig5}. From Fig.\ref{fig5}, it can be seen that the predictive results obtained by fixed multipliers' values are not as good as the ones obtained by dynamic changing multipliers.
\subsection{Predicting the future response from noisy data}
To investigate the robustness of our proposed method, we add noise with the following formulation into the training data \cite{2aISI:000527390200029}\\
\begin{equation}
h^*\left( {t,x,y} \right) = h\left( {t,x,y} \right) + h_{diff} \left( {x,y} \right) \times \alpha \%  \times \varepsilon
\label{neqf10}
\end{equation}
where $h_{diff} \left( {x,y} \right)$, $\alpha \%$, and $\varepsilon$ denote the maximal difference obtained at location $\left( {x,y} \right)$ during the entire monitoring process, the noise level, and a uniform variable ranging from -1 to 1.

\begin{figure*}[htb]
\begin{center}
\subfigure[TgNN-LD: Correlation between reference and prediction]{\includegraphics[width=0.30\textwidth]{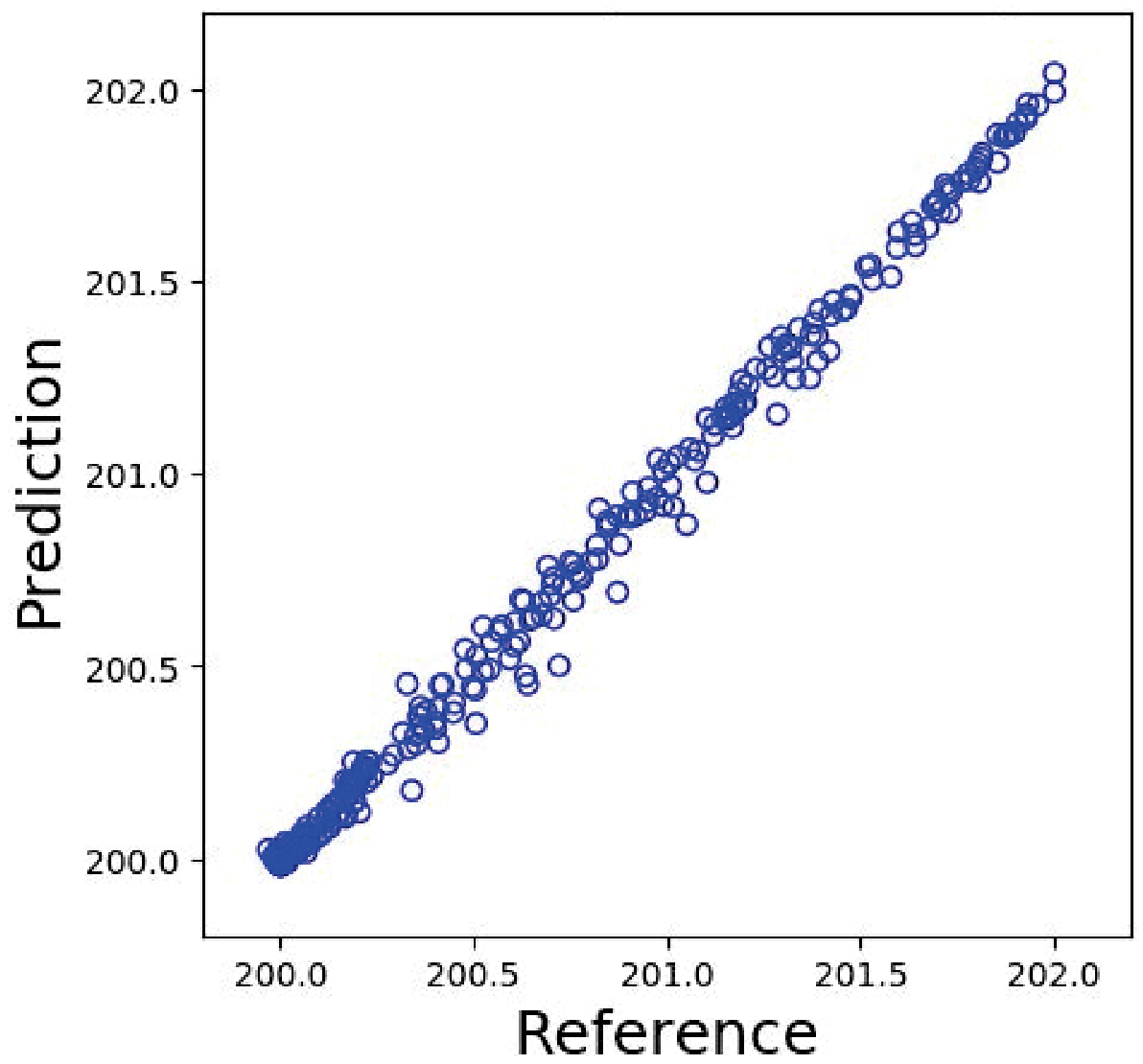}}
\subfigure[TgNN: Correlation between reference and prediction]{\includegraphics[width=0.30\textwidth]{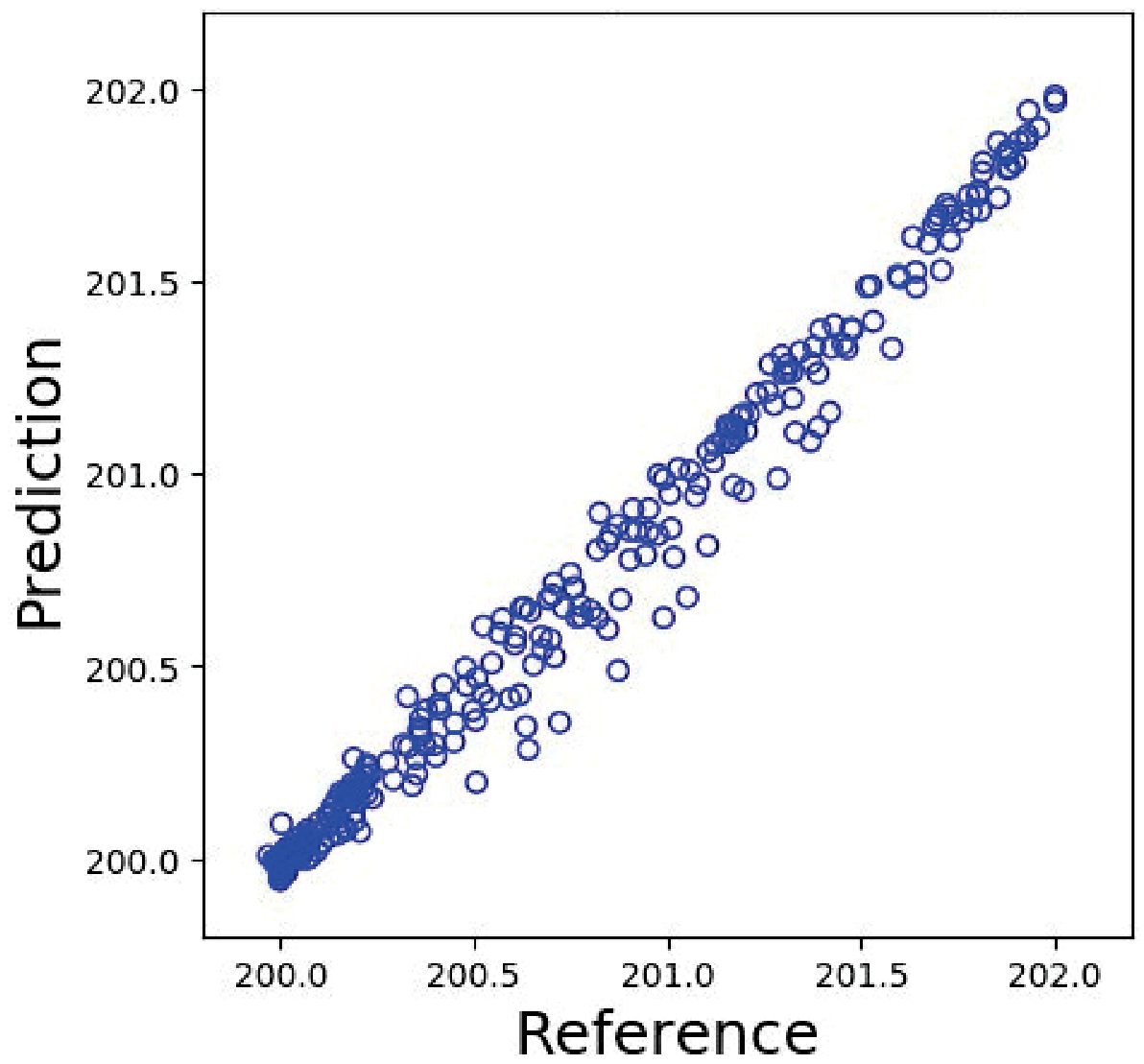}}
\subfigure[TgNN-1: Correlation between reference and prediction]{\includegraphics[width=0.30\textwidth]{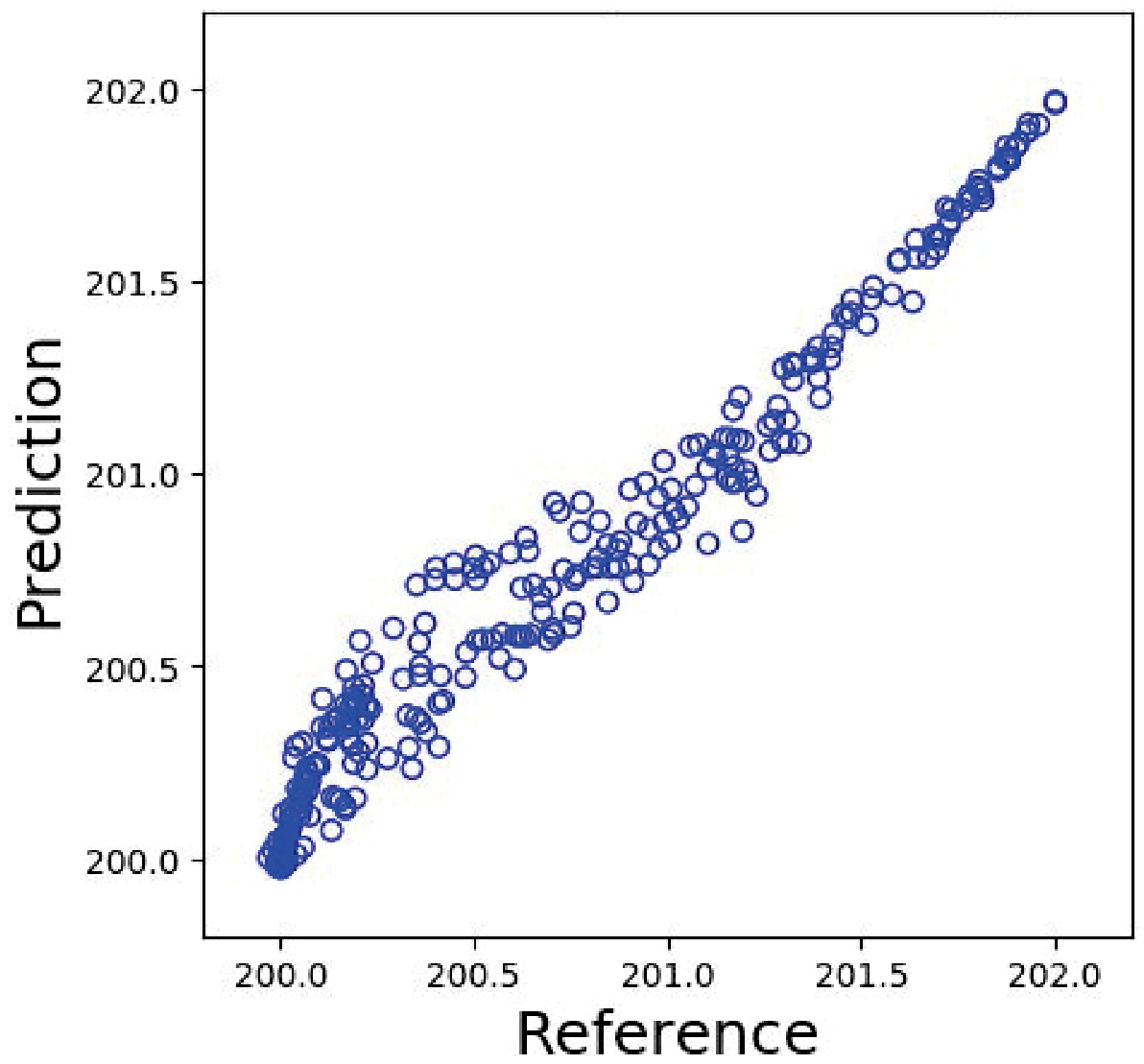}}\\
\subfigure[TgNN-LD: Prediction vs reference]{\includegraphics[width=0.30\textwidth]{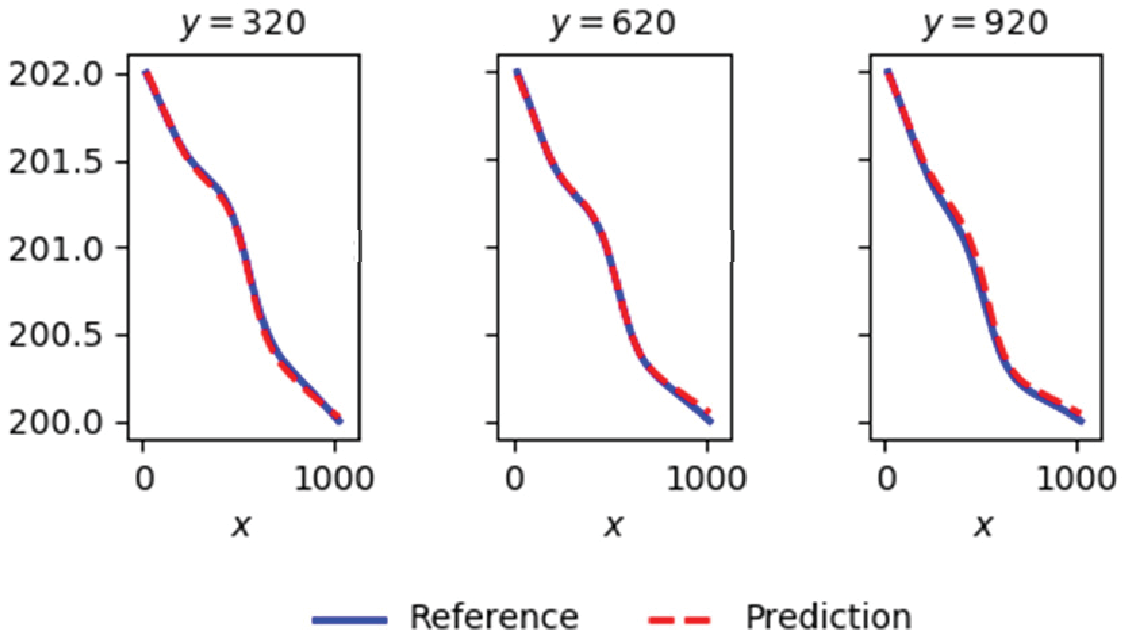}}
\subfigure[TgNN: Prediction vs reference]{\includegraphics[width=0.30\textwidth]{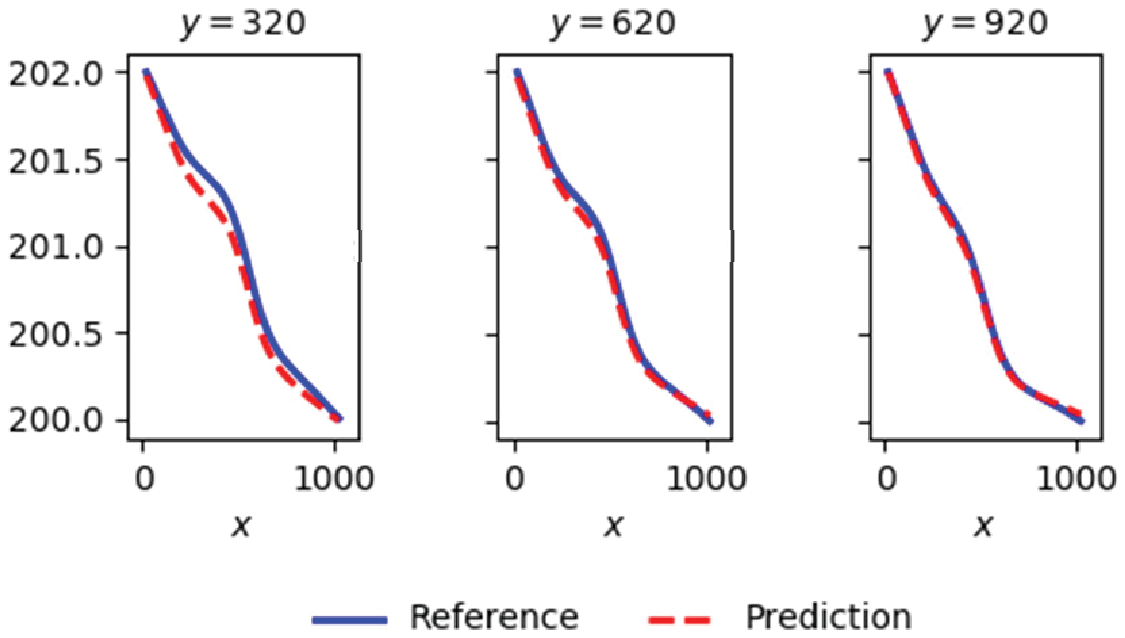}}
\subfigure[TgNN-1: Prediction vs reference]{\includegraphics[width=0.30\textwidth]{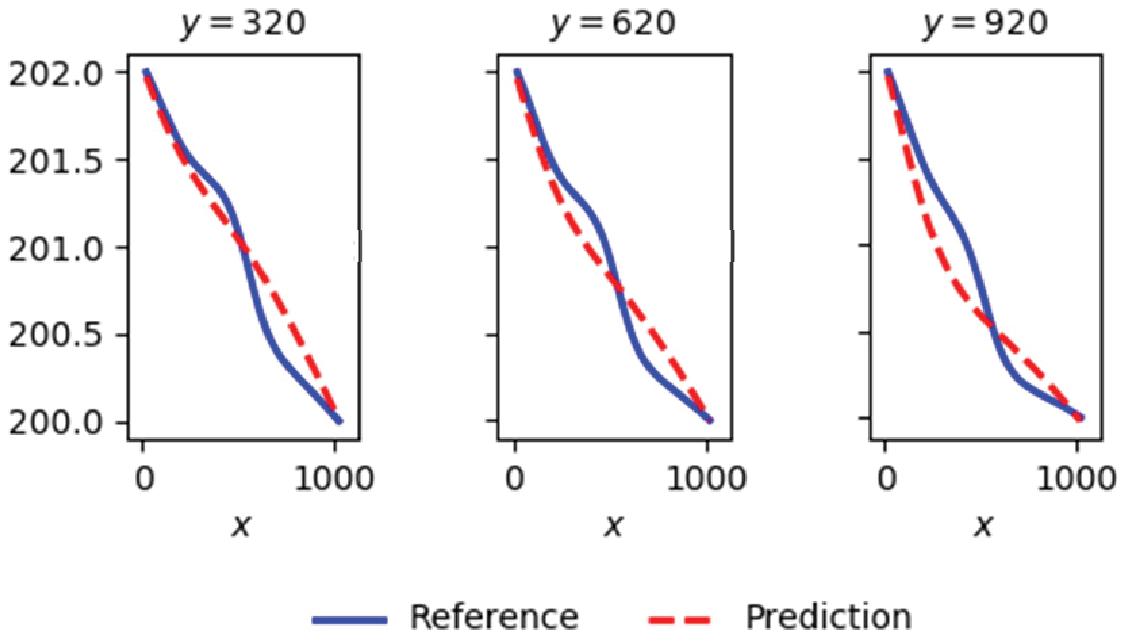}}\\
\caption{Predictive results obtained by TgNN-LD, TgNN, and TgNN-1 with $\alpha \%  = 5\%$.}
\label{fig6}
\end{center}
\end{figure*}

\begin{figure*}[htb]
\begin{center}
\subfigure[TgNN-LD: Correlation between reference and prediction]{\includegraphics[width=0.30\textwidth]{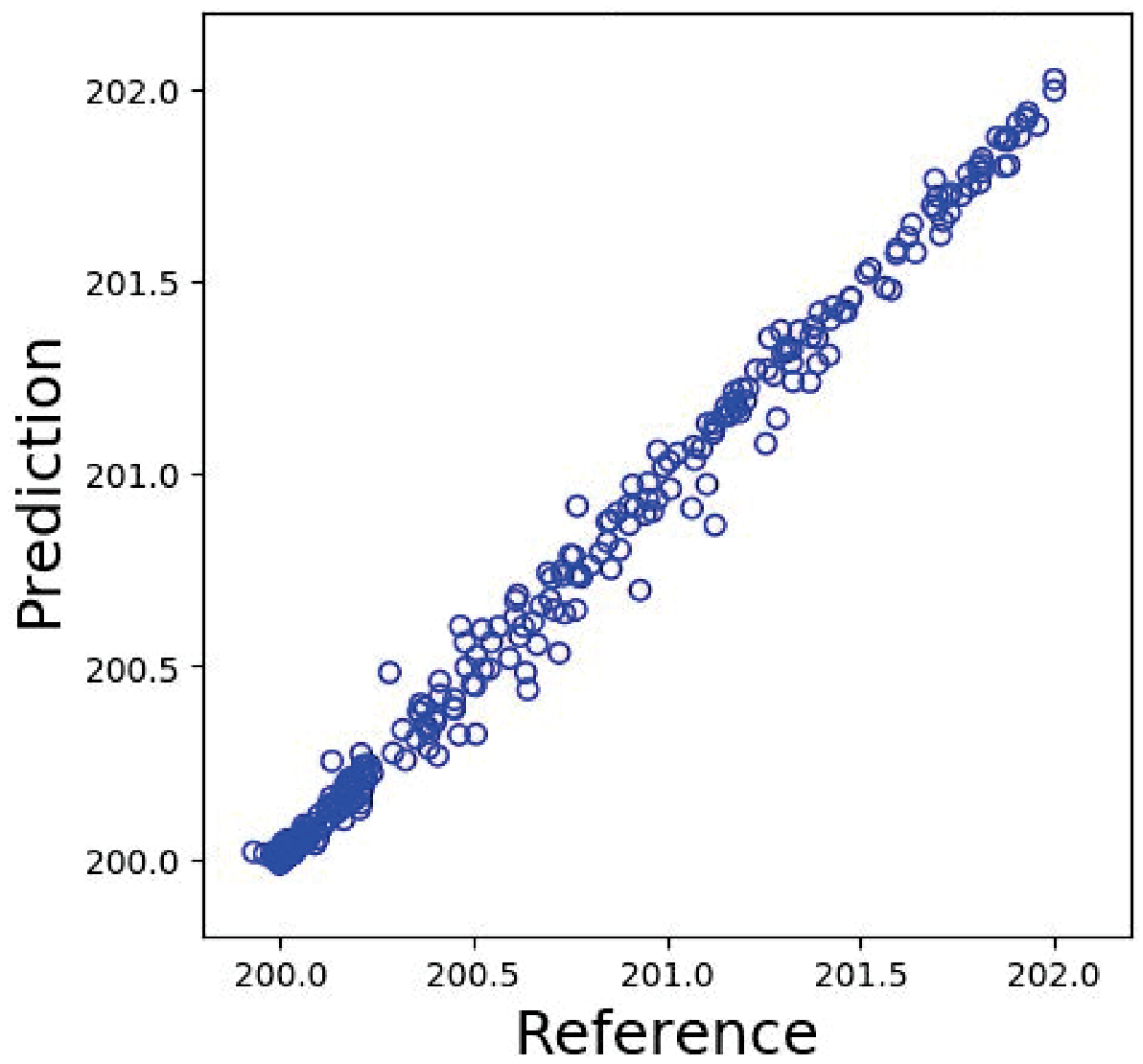}}
\subfigure[TgNN: Correlation between reference and prediction]{\includegraphics[width=0.30\textwidth]{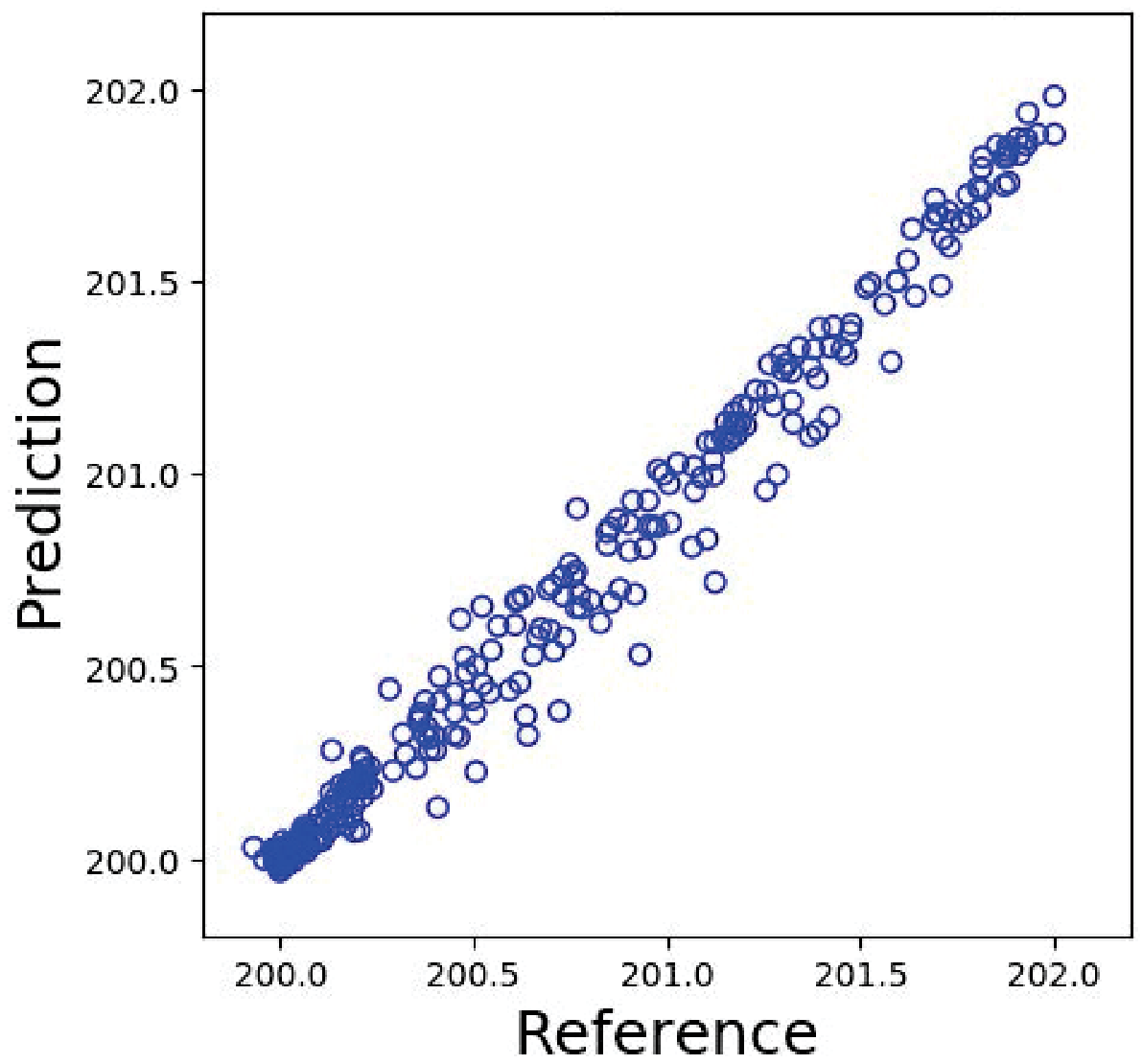}}
\subfigure[TgNN-1: Correlation between reference and prediction]{\includegraphics[width=0.30\textwidth]{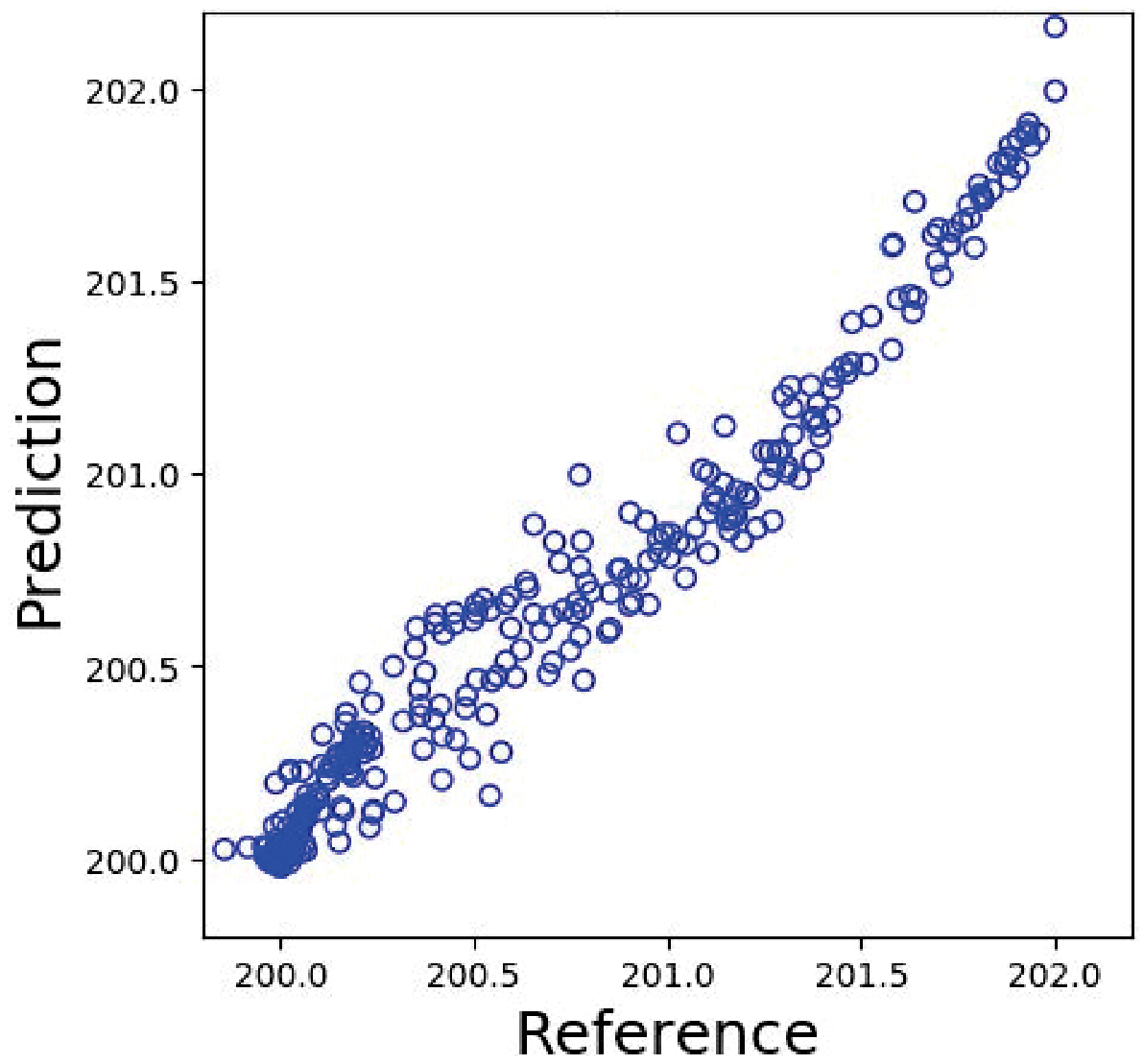}}\\
\subfigure[TgNN-LD: Prediction vs reference]{\includegraphics[width=0.30\textwidth]{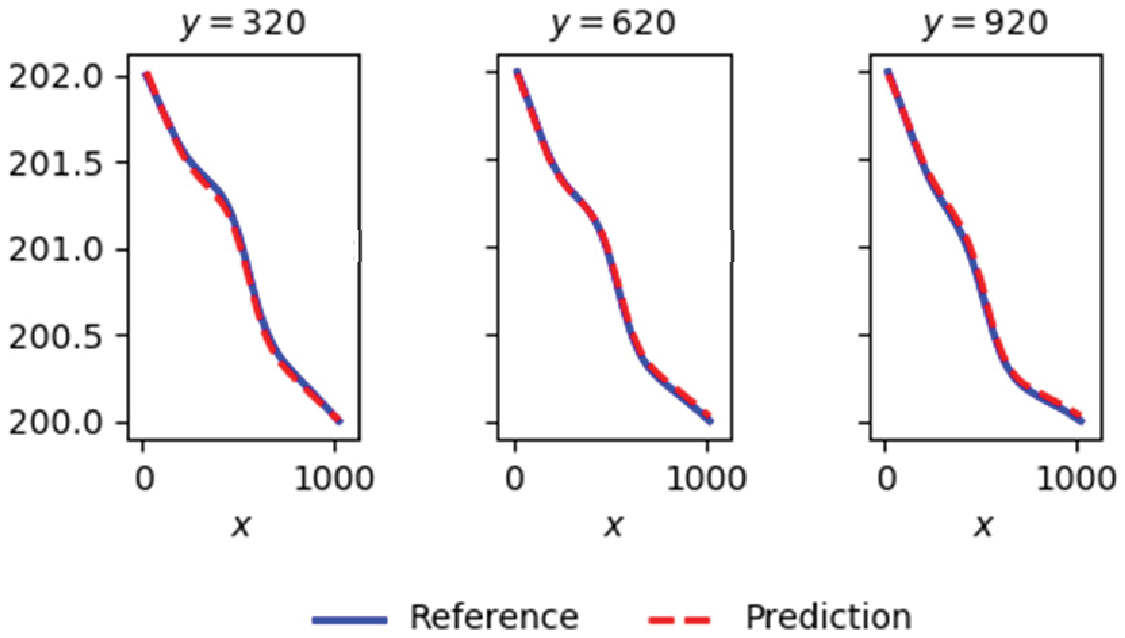}}
\subfigure[TgNN: Prediction vs reference]{\includegraphics[width=0.30\textwidth]{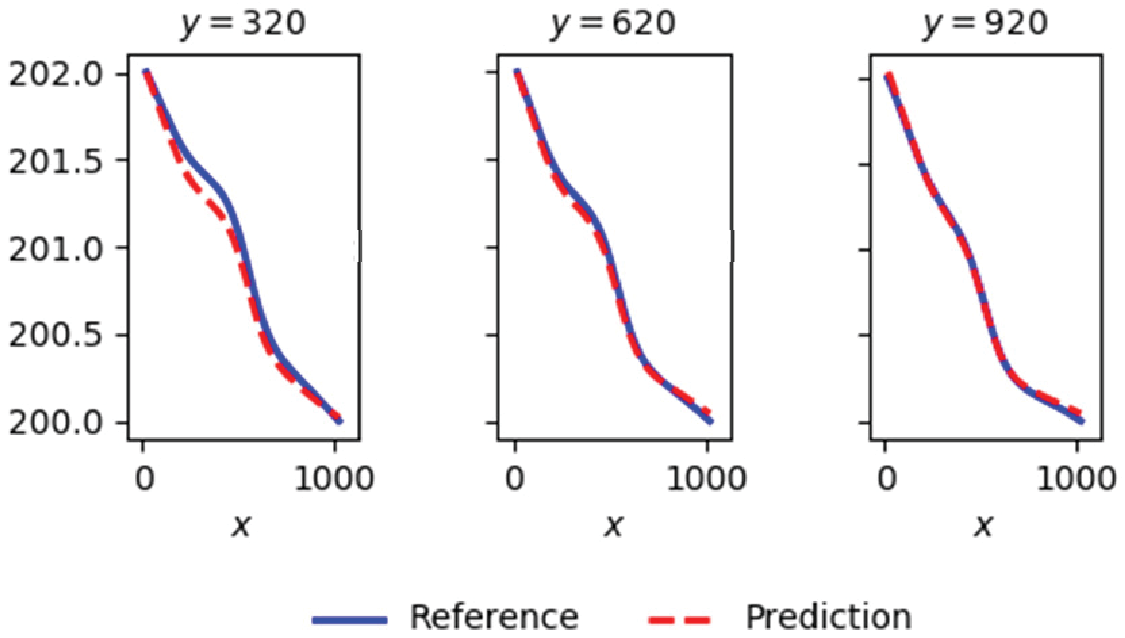}}
\subfigure[TgNN-1: Prediction vs reference]{\includegraphics[width=0.30\textwidth]{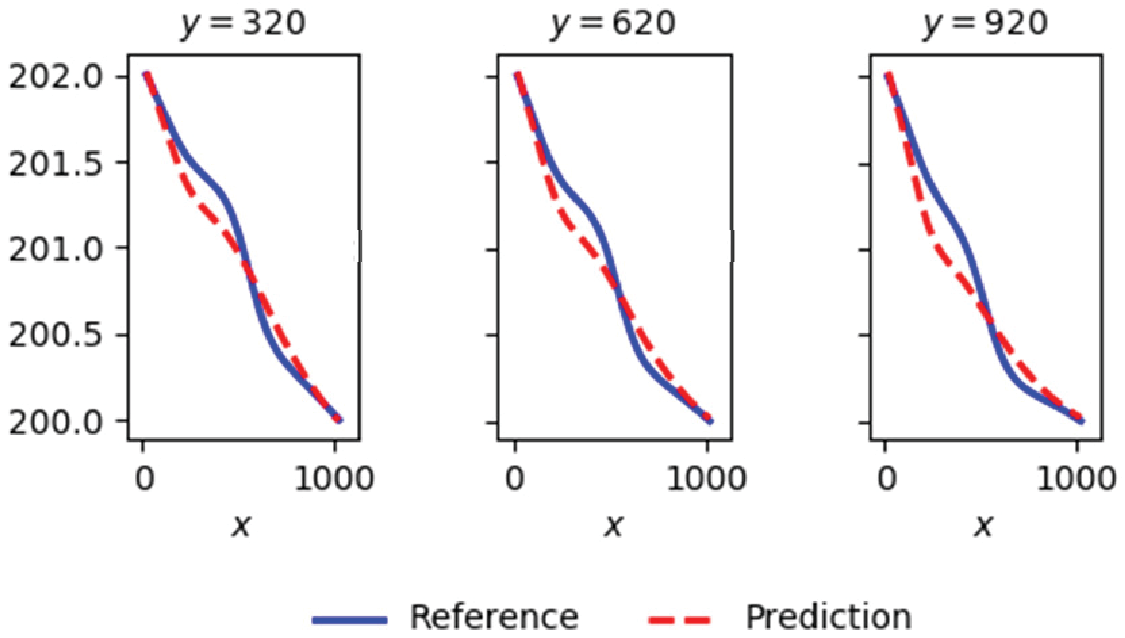}}\\
\caption{Predictive results obtained by TgNN-LD, TgNN, and TgNN-1 with $\alpha \%  = 10\%$.}
\label{fig7}
\end{center}
\end{figure*}

\begin{figure*}[htb]
\begin{center}
\subfigure[TgNN-LD: Correlation between reference and prediction]{\includegraphics[width=0.30\textwidth]{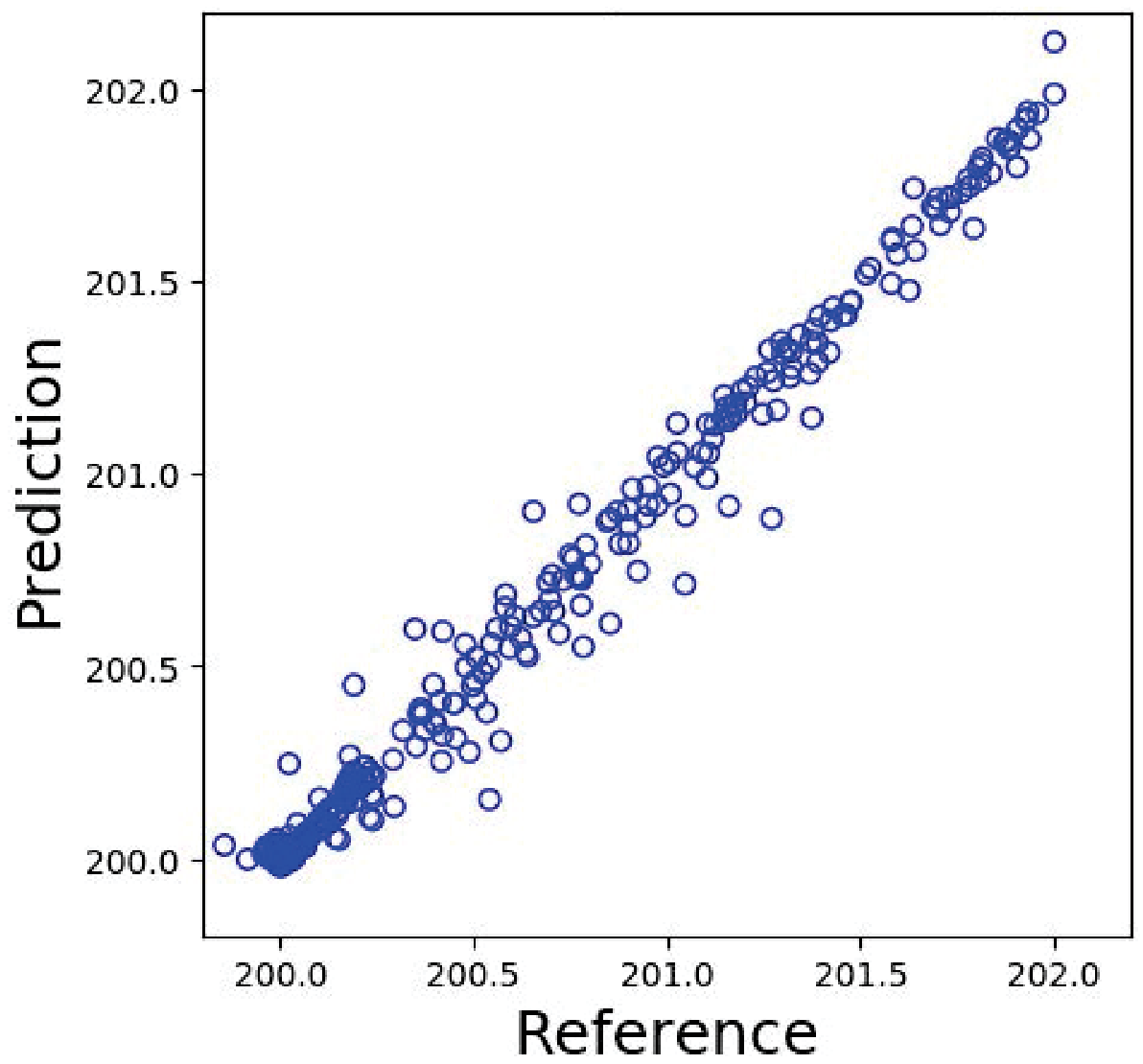}}
\subfigure[TgNN: Correlation between reference and prediction]{\includegraphics[width=0.30\textwidth]{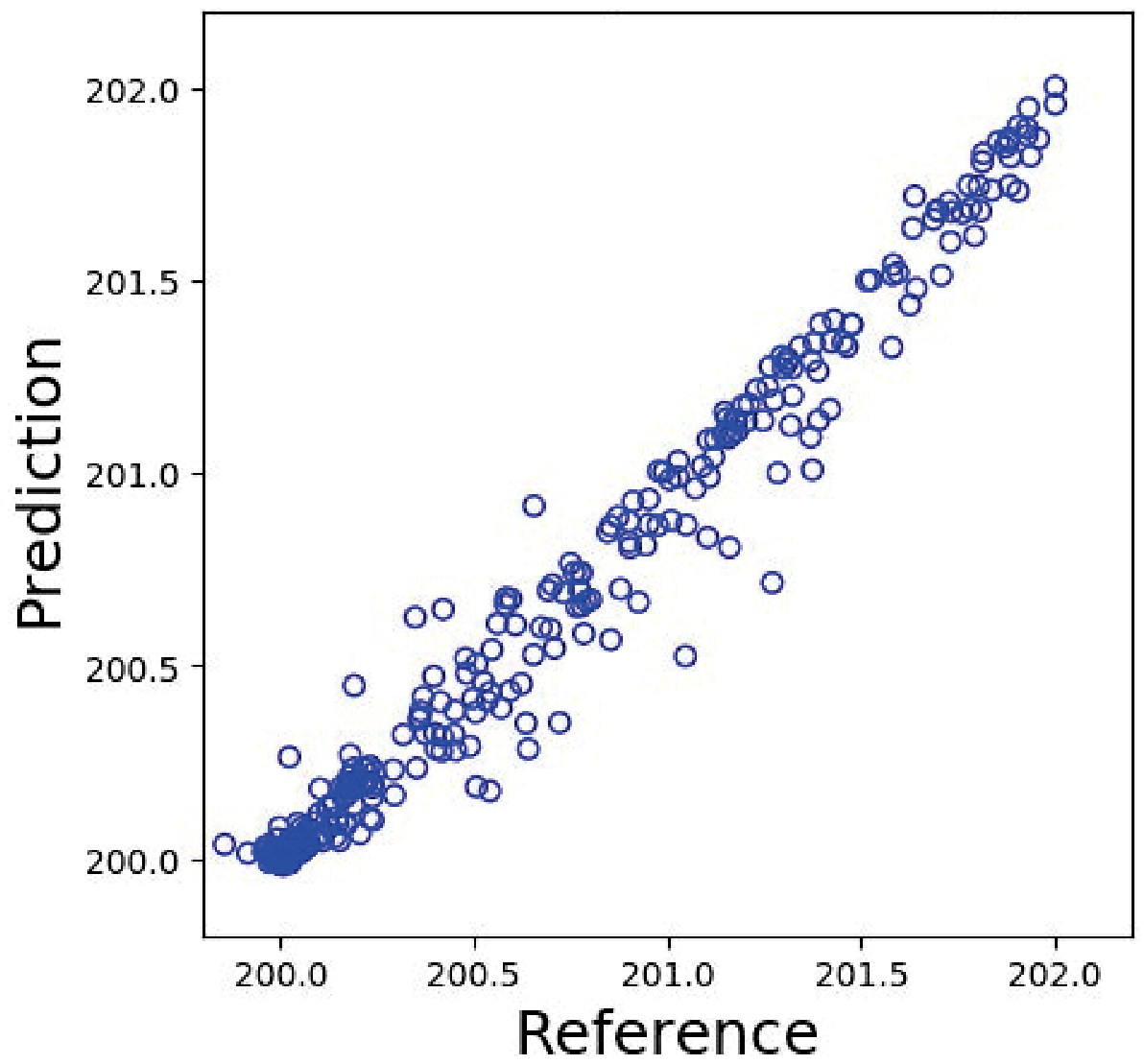}}
\subfigure[TgNN-1: Correlation between reference and prediction]{\includegraphics[width=0.30\textwidth]{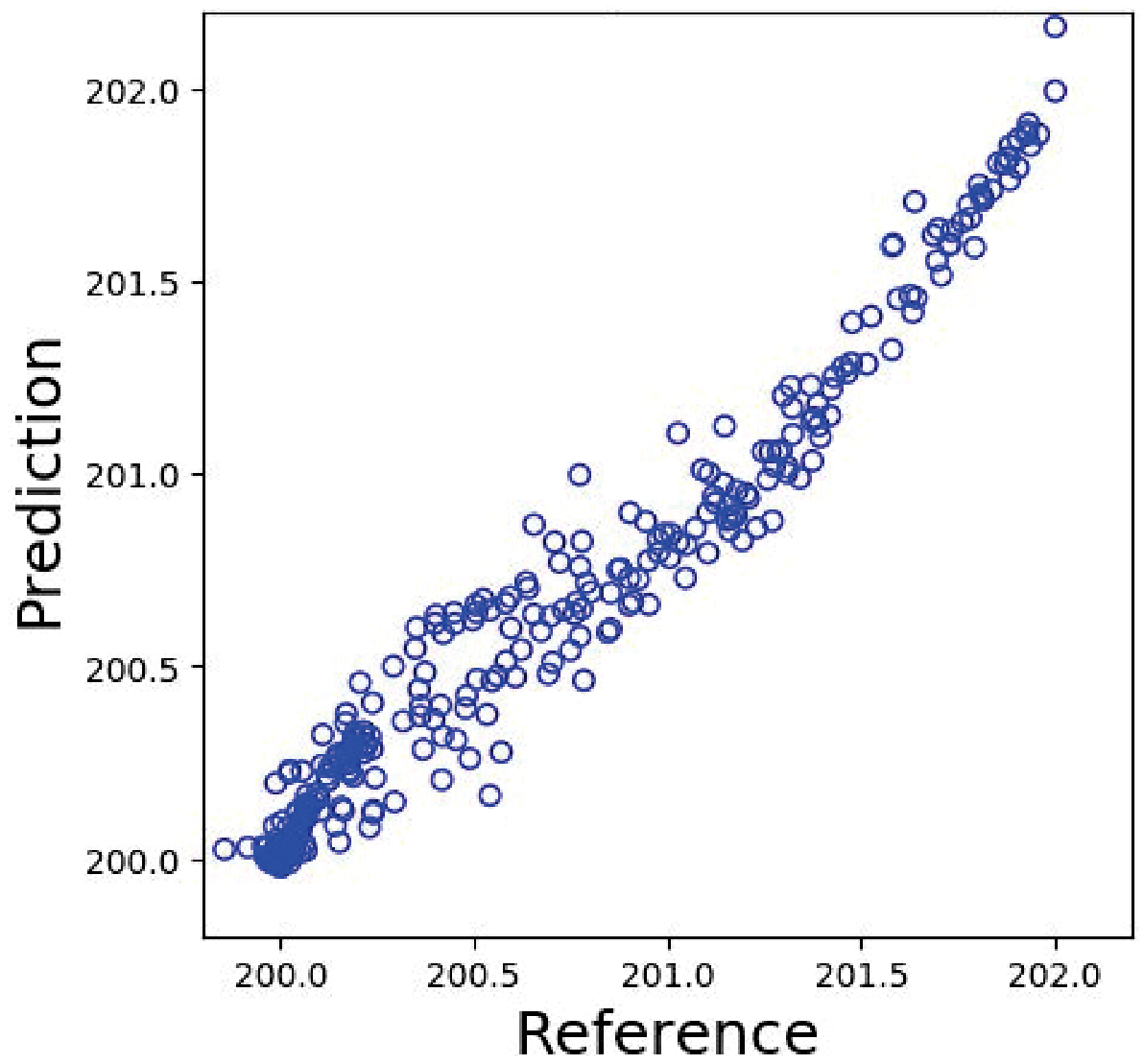}}\\
\subfigure[TgNN-LD: Predctition vs reference]{\includegraphics[width=0.30\textwidth]{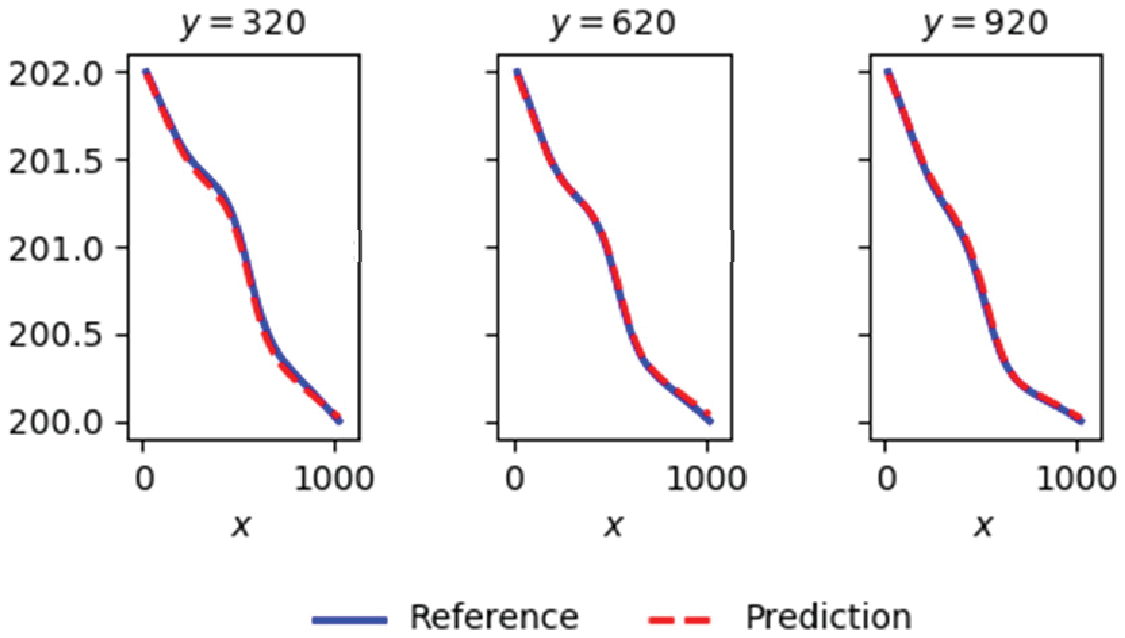}}
\subfigure[TgNN: Predctition vs reference]{\includegraphics[width=0.30\textwidth]{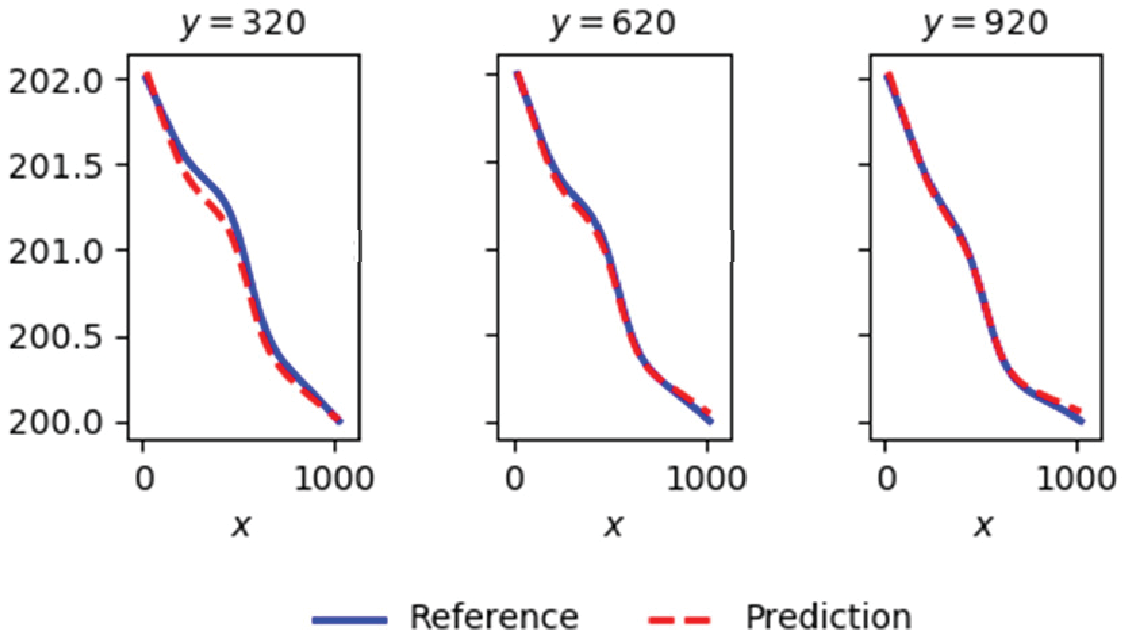}}
\subfigure[TgNN-1: Predctition vs reference]{\includegraphics[width=0.30\textwidth]{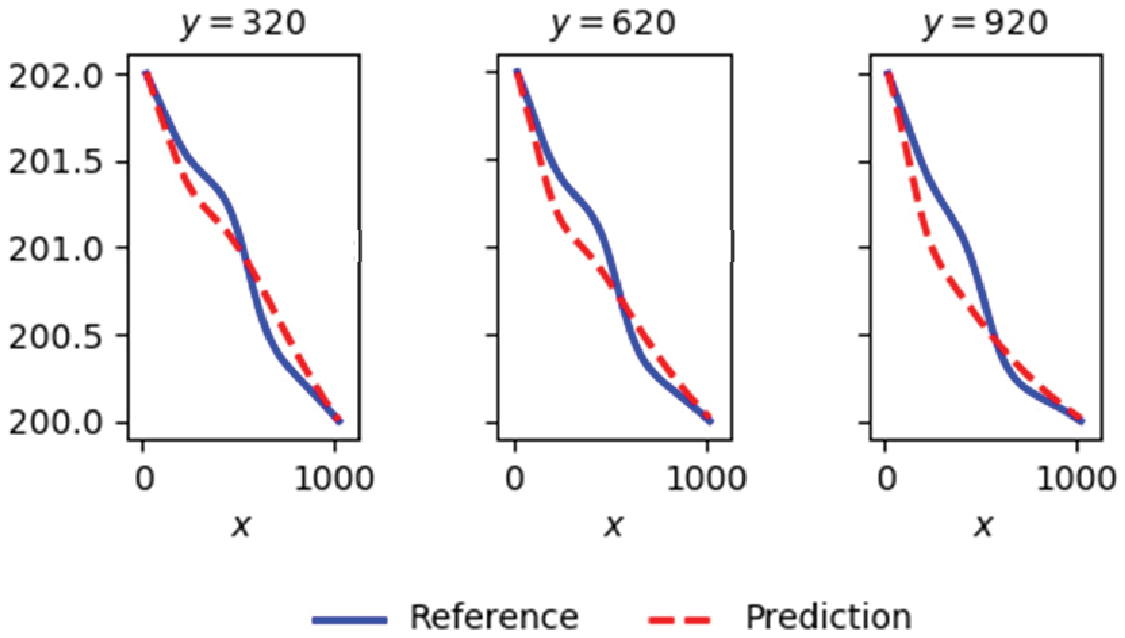}}\\
\caption{Predictive results obtained by TgNN-LD, TgNN, and TgNN-1 with $\alpha \%  = 20\%$.}
\label{fig8}
\end{center}
\end{figure*}

Figs.\ref{fig6}-\ref{fig8} show the predictive results obtained by TgNN-LD, TgNN, and TgNN-1 under noise levels of 5\%, 10\%, and 20\%, respectively, with their error L2 and R2 results listed in Tab.\ref{tab:table5Values_of_different_noise_percentage}. From Figs.\ref{fig6}-\ref{fig8} and Tab.\ref{tab:table5Values_of_different_noise_percentage}, it can be found that TgNN-LD can always obtain the best results among the three methods on correlation between the reference and predicted hydraulic head when noise exists. Moreover, it can also achieve the best results on the error L2 and R2 results. Although the training time obtained by TgNN-LD under different noise levels seems slightly longer than the other two, comparisons between the prediction and reference demonstrate the improvement of incorporating the Lagrangian dual approach.
\begin{table*}[htbp]
  \centering
  \caption{Prediction results with different noise percentages obtained by TgNN-LD, TgNN, and TgNN-1.}
    \begin{tabular}{cccccccccccccccccccccccccccc}
    \toprule
	& noise level	&L2 error	&R2	&Training time/s\\
\midrule
TgNN-LD	&$\alpha \%  = 5\%$ &\textbf{2.2848E-04}&	\textbf{9.9481E-01}&	206.9471\\
\midrule
TgNN	& $\alpha \%  = 5\%$ &	4.9887E-04&	9.7524E-01&	204.6485\\
\midrule
TgNN-1	&$\alpha \%  = 5\%$	&6.4963E-04&	9.5802E-01&	\textbf{201.8772}\\
\midrule
TgNN-LD	& $\alpha \%  = 10\%$ & \textbf{2.6835E-04}&	\textbf{9.9285E-01}&	208.6079\\
\midrule
TgNN	& $\alpha \%  = 10\%$ &	4.9018E-04&	9.7614E-01	&\textbf{202.5091}\\
\midrule
TgNN-1	& $\alpha \%  = 10\% $&	5.6168E-04&	9.6867E-01	&205.8537\\
\midrule
TgNN-LD	& $\alpha \%  = 20\% $& \textbf{3.6974E-04}&	\textbf{9.8652E-01}	&204.9997\\
\midrule
TgNN	& $\alpha \%  = 20\% $&	5.5094E-04	&9.7007E-01	&\textbf{204.1725}\\
\midrule
TgNN-1	& $\alpha \%  = 20\%$ &	7.2383E-04	&9.4833E-01	&204.9317\\
    \bottomrule
    \end{tabular}%
  \label{tab:table5Values_of_different_noise_percentage}%
\end{table*}%
\subsection{Training under the dynamic epoch}
To investigate the predictive performance more deeply, we substitute the stopping criterion, i.e., a fixed number of iterations, with a dynamic epoch, which has a close relationship with changes of loss values. From Fig.\ref{fig1}, it can be seen that the training process has usually already converged with the iteration number of 2000. Therefore, we set the total number of epoch, denoted as $n_{total}$, as 2000. The dynamic epoch is denoted as $n_D$.

\begin{figure*}
\centering
\subfigure[stopping epoch=1712]{\includegraphics[width=0.30\textwidth]{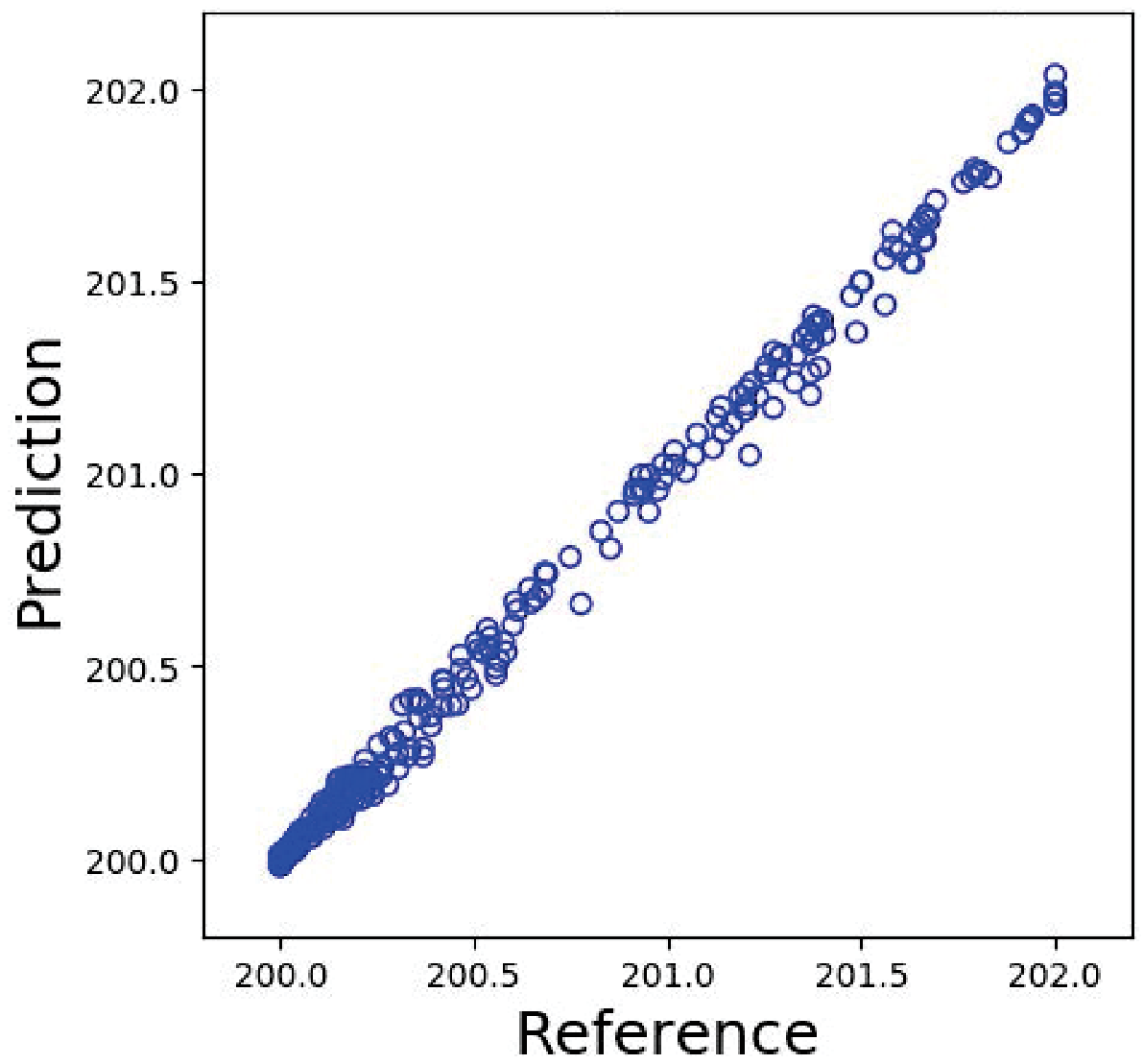}}
\subfigure[stopping epoch=1810]{\includegraphics[width=0.30\textwidth]{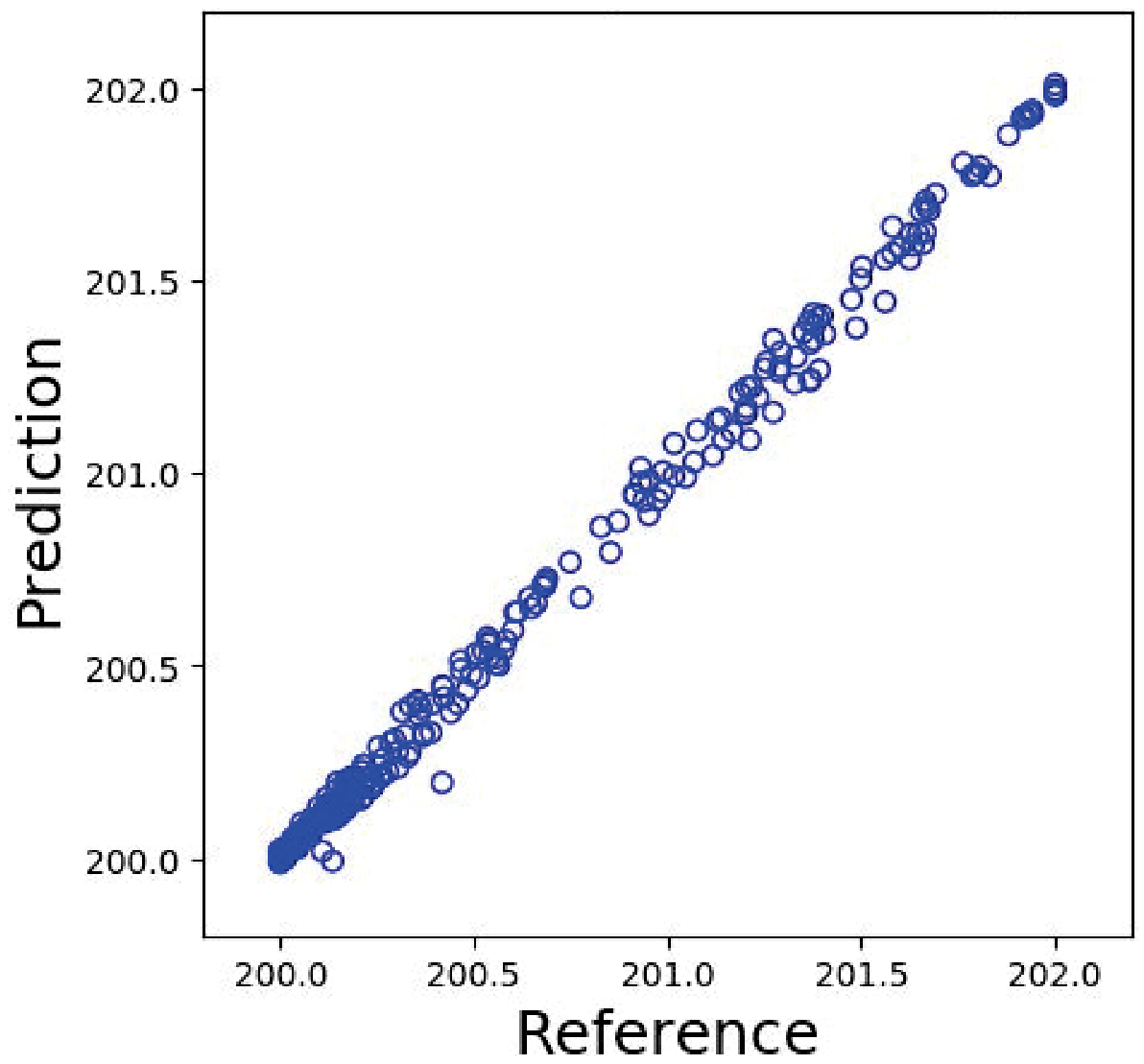}}
\subfigure[stopping epoch=1829]{\includegraphics[width=0.30\textwidth]{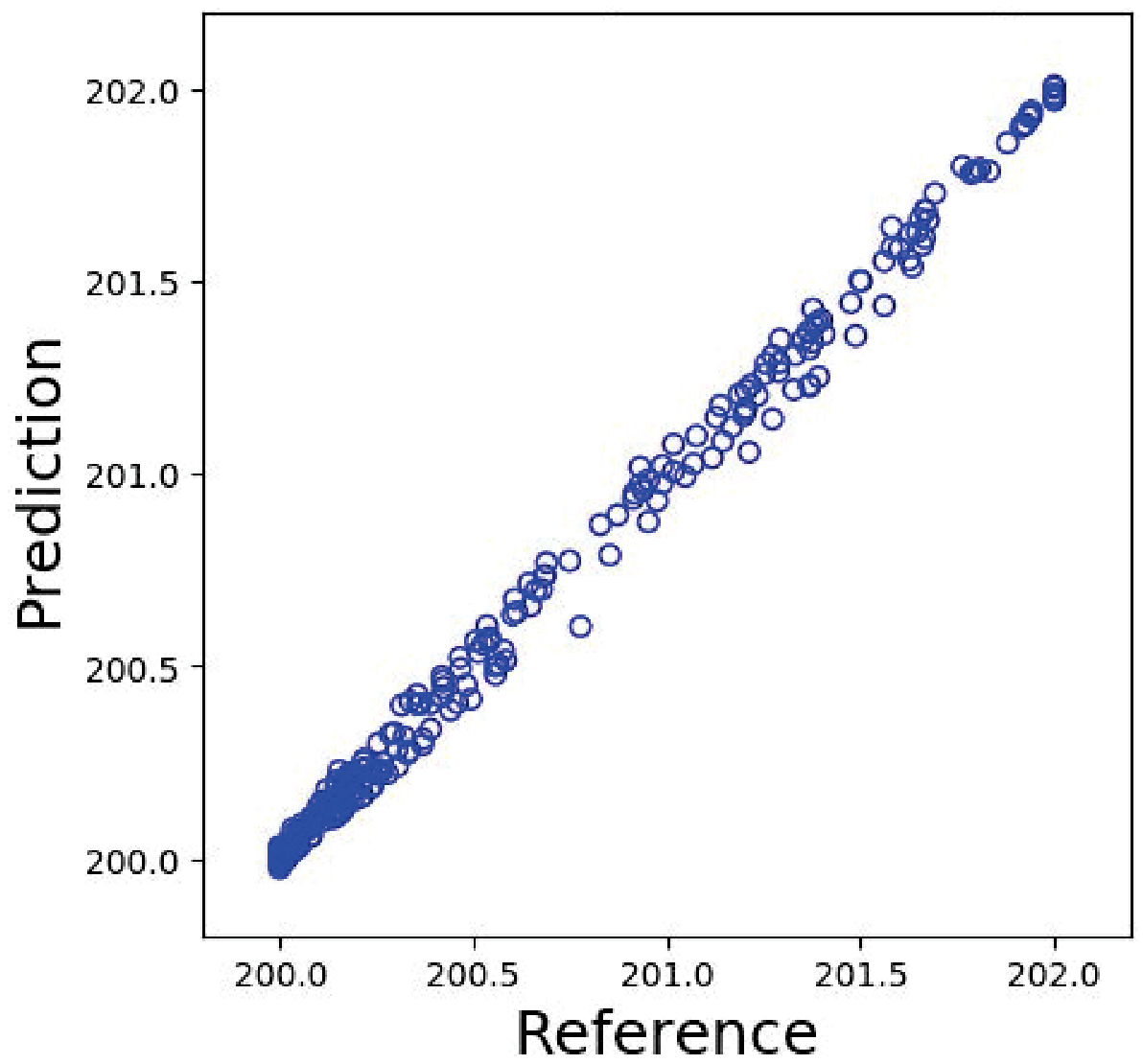}}\\
\subfigure[stopping epoch=1712]{\includegraphics[width=0.3\textwidth]{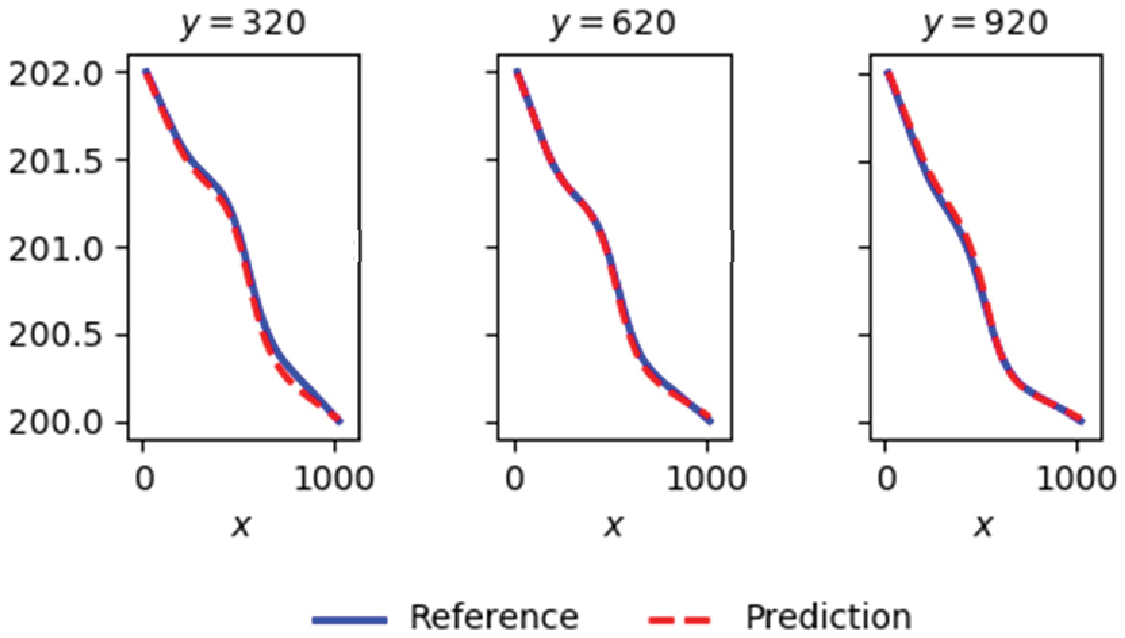}}
\subfigure[stopping epoch=1810]{\includegraphics[width=0.3\textwidth]{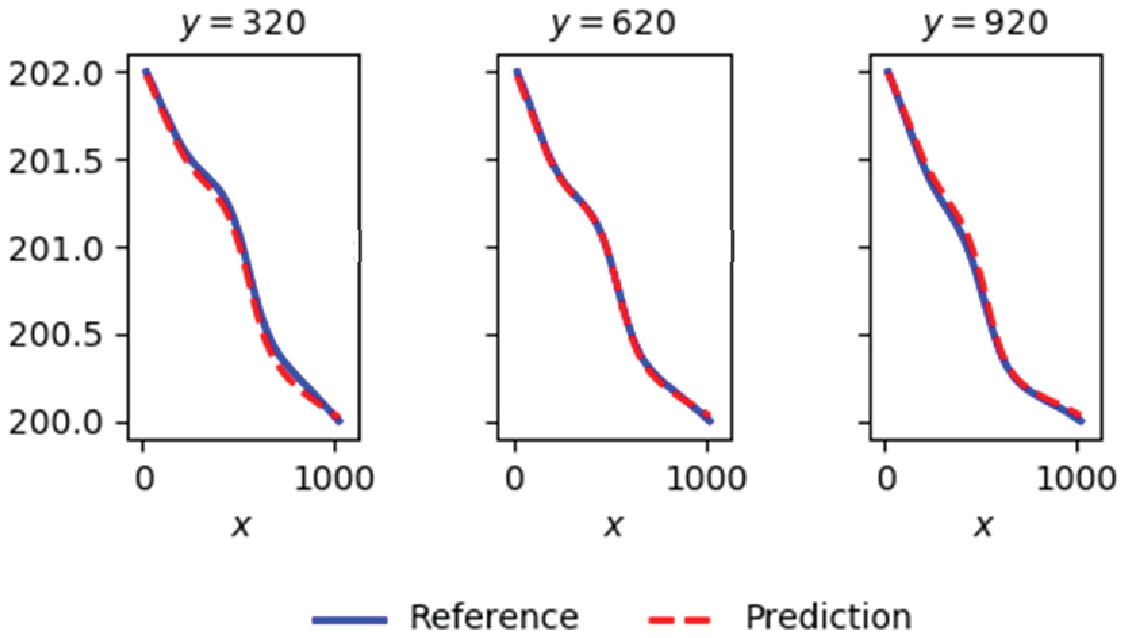}}
\subfigure[stopping epoch=1829]{\includegraphics[width=0.3\textwidth]{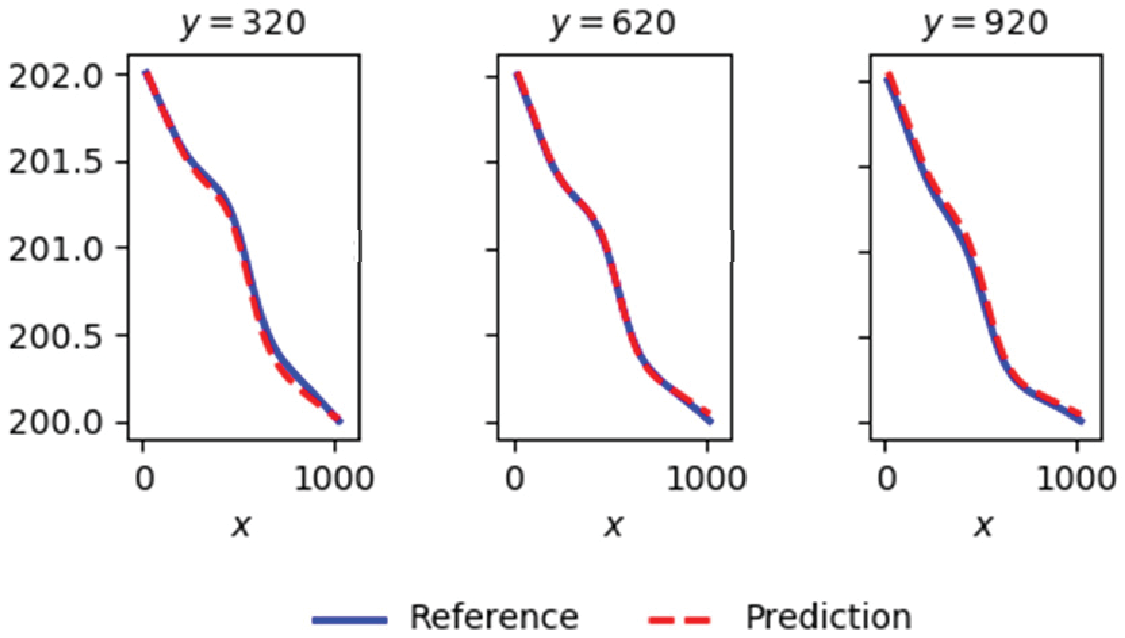}}\\
\caption{Predictive results obtained by TgNN-LD with a dynamic epoch.}
\label{fig11}
\end{figure*}

In our experiment, we maintain a time window with the length of $L_C$. For per obtained loss value, we compare whether loss values in the current time window are less than a threshold, $\beta$. If this criterion is satisfied, we stop the iteration and output the predictive results. Herein, we recommend $L_C=10$ and $\beta=0.006$. Tab.\ref{tab:compare_dynamic_epochs1} lists a number of good prediction results under the dynamic epoch obtained by TgNN-LD, with their results on correlation between reference and prediction and prediction versus reference being presented in Fig.\ref{fig11}. From the numerical results in Tab.\ref{tab:compare_dynamic_epochs1}, it seems that TgNN-LD can come to convergence by a dynamic stopping epoch.
\begin{table*}[htbp]
  \centering
  \caption{A number of good prediction results under the dynamic epoch obtained by T\MakeLowercase{g}NN-LD.}
    \begin{tabular}{cccccccccccccccccccccccccccc}
    \toprule
    stopping epoch&	error L2&	R2	&Training time/s\\
     \midrule
    1712&	2.0178E-04&	0.995948184&	164.2439\\
    \midrule
    1810&	2.0807E-04&	0.99569132&	    204.4629\\
    \midrule
    1829&	2.1342E-04&	0.995467097&	203.6778\\
    \bottomrule
    \end{tabular}%
  \label{tab:compare_dynamic_epochs1}
\end{table*}%
\section{Conclusion}\label{sec5}
In this paper, we propose a Lagrangian dual-based TgNN framework to assist in balancing training data and theory in the TgNN model. It provides theoretical guidance for the update of weights for the theory-guided neural network framework. Lagrangian duality is incorporated into TgNN to automatically determine the weight values for each term and maintain an excellent tradeoff between them. The subsurface flow problem is investigated as a test case. Experimental results demonstrate that the proposed method can increase the predictive accuracy and produce a superior training model compared to that obtained by an \emph{ad-hoc} procedure within limited computational time.

In the future, we would like to combine the proposed Lagrangian dual-based TgNN framework with more informed deep learning approaches, such as TgNN with weak-form constraints. It can also be utilized to solve more application problems, such as the two-phase flow problem in energy engineering, to enhance the training ability of TgNN and achieve accurate predictions.

\bibliographystyle{IEEEtran}
\bibliography{mybibfile}

\end{document}